\documentclass[final,3p,times]{elsarticle} 

\usepackage{lineno,hyperref}

\biboptions{sort&compress}

\usepackage[utf8]{inputenc}
\usepackage[T1]{fontenc}
\usepackage{floatpag}
\usepackage{float}
\usepackage{microtype}
\usepackage{graphicx}
\usepackage{epsfig}
\usepackage[export]{adjustbox}
\usepackage{soul}
\usepackage{subfig}
\usepackage{pifont}
\usepackage{multirow}
\usepackage{amsmath}
\usepackage{amssymb}
\usepackage{bm}
\usepackage[flushleft]{threeparttable}
\usepackage{mdframed}
\usepackage{array}
\usepackage{booktabs}
\usepackage{lscape}
\usepackage{rotating}
\usepackage{url}
\usepackage[ruled,vlined,titlenumbered]{algorithm2e}
\usepackage[x11names,table]{xcolor}
\usepackage{boldline}

\newtheorem{example}{Example}

\newdefinition{remark}{Remark}

\SetAlFnt{\scriptsize}
\SetAlCapFnt{\scriptsize}
\SetAlCapNameFnt{\scriptsize}

\usepackage{hyperref}

\begin{document}

\begin{frontmatter}

\title{Uncertainty in Automated Ontology Matching: Lessons Learned from an Empirical Experimentation}




\author[1]{Inès \textsc{Osman}\corref{mycorrespondingauthor}} 
\ead{ines.osman@fst.utm.tn}

\author[2]{Salvatore F. \textsc{Pileggi}}
\ead{SalvatoreFlavio.Pileggi@uts.edu.au}

\author[3,4]{Sadok \textsc{Ben Yahia}}
\ead{sadok.ben@taltech.ee}

\cortext[mycorrespondingauthor]{Corresponding author}

\address[1]{LIPAH - LR11ES14, Faculty of Sciences of Tunis, University of Tunis El Manar, Tunisia}

\address[2]{Faculty of Engineering and IT, University of Technology Sydney, Australia}

\address[3]{Department of Software Science,
Tallinn University of Technology, Estonia}

\address[4]{The Maersk Mc-Kinney Moller Institute, Center of Industrial Software,  University of Southern Denmark, Alsion 2, 6400- Sønderborg, Denmark}

%

\begin{abstract}
Data integration is considered a classic research field and a pressing need within the information science community. Ontologies play a critical role in such a process by providing well-consolidated support to link and semantically integrate datasets via interoperability. This paper approaches data integration from an application perspective, looking at techniques based on \textit{ontology matching}. An ontology-based process may only be considered adequate by assuming manual matching of different sources of information. However, since the approach becomes unrealistic once the system scales up, automation of the matching process becomes a compelling need. Therefore, we have conducted experiments on actual data with the support of existing tools for automatic ontology matching from the scientific community.
Even considering a relatively simple case study (\textit{i.e.}, the spatio-temporal alignment of global indicators), outcomes clearly show significant uncertainty resulting from errors and inaccuracies along the automated matching process. More concretely, this paper aims to test on real-world data a bottom-up knowledge-building approach, discuss the lessons learned from the experimental results of the case study, and draw conclusions about uncertainty and uncertainty management in an automated ontology matching process. While the most common evaluation metrics clearly demonstrate the unreliability of fully automated matching solutions, properly designed semi-supervised approaches seem to be mature for a more generalized application.



\end{abstract}



\begin{keyword}
\texttt{Ontology}\sep~\,\texttt{Ontology\,\,Matching} 
\sep~\,\texttt{Uncertainty} 
\sep~\,\texttt{Uncertainty\,\,Management}
\sep~\,\texttt{Semantic\,\,Web}
\MSC[2010] 00-01\sep  99-00
\end{keyword}

\end{frontmatter}


\section{Introduction}

Data integration, defined as \textit{"the problem of combining data residing at different sources and providing the user with a unified view of these data"}~\cite{lenzerini2002data}, can be considered a well-covered research field, as could witness the myriad of contributions in literature. Its relevance is determined by the practical implications in the different application domains, and it is well-recognized within the information science community.
Most modern systems work at a semantic level~\cite{GARCIASANCHEZ2020102153} where data integration may be understood at different levels (\textit{e.g.} concept~\cite{ZHANG2006121} or multi-media~\cite{LIN2007488}).
Semantic technology has been largely adopted in data integration~\cite{lenzerini2002data}. 
It is definitely central to a more holistic approach, where data integration is considered a part of a more complex knowledge-building process.
The proper adoption of semantic technology is extremely effective in supporting data integration and reuse via interoperability~\mbox{\cite{noy2004semantic}}.
Overall, associating formal semantics with data is a key step in the fields of artificial intelligence and database management. In addition, the analysis of semantic data can underpin sophisticated data mining techniques~\cite{delgado2001mining,dou2015semantic}.

In the context of this work, "knowledge building" is seen as the process of combining "raw data" in order to create "rich data spaces" in which "semantics" are defined in a formal way ~\cite{lenzerini2002data}. While data integration aims at establishing a common, unified view of data from different sources, the specification of formal semantics enables a further level of complexity as the "meaning" of data is also represented. 
Ontology is a classic philosophical concept that is related to the study of "the nature of being".
It has become an active part of the computer world~\cite{guarino1995formal}. Ontologies are rich data models aimed at the specification of semantics. They support the representation and processing of knowledge in a machine-readable format according to a model close to the human one. The adoption of ontologies allows effective solutions to knowledge building since semantic data is enabled in the Semantic Web~\cite{berners_Lee_2001,shadbolt2006semantic} and modeled according to an advanced interoperability model, which is commonly referred to as \textit{semantic interoperability}~\cite{decker2000semantic}.
We fully rely on an ontology-based approach to support the data integration process. The benefits of ontology in different application domains are well-known and have been extensively discussed from different perspectives in several contributions.


An ad-hoc approach to knowledge building is time-consuming, error-prone, and, in general, very expensive. Looking at the increasing complexity and scale of systems, automated and semi-automated data integration processes are becoming more and more relevant to assure effectiveness and performance on a large scale. The automation of the knowledge-building process becomes required once the target system scales up.

Ontology matching is a crucial step in the knowledge-building process. It reconciles the differences between ontologies and resolves their heterogeneity problem. A seamless and systematic knowledge-building process
may only be considered adequate by assuming a manual matching of concepts from the different sources of information. However, since manual matching is far from being scalable, automation becomes a compelling need. Automated ontology matching systems use a similarity computation algorithm to find similarities between ontologies to be integrated. However, as extensively discussed in the rest of this paper, such automation leads to a situation of uncertainty since correctness and accuracy cannot be guaranteed in general terms. In this paper, we focus on uncertainty and uncertainty management in the automated process of ontology matching.

We approach the knowledge-building process from a practical perspective. Indeed, we apply a real-world case study dealing with the spatio-temporal alignment of global indicators. We propose several experiments on real data by adopting existing tools for dataset conversion and ontology matching. We first adopt a previously developed conversion tool that enables the systematic translation from raw data (relational tables) to rich semantic datasets (ontologies)~\cite{pileggi2020ontological}. Then, we adopt one of the best ontology matching tools available in the ontology community to find similarities between the resulting ontologies (i.e., the converted datasets).
To test the current limitations of automatic ontology matching in practice, we have measured the uncertainty resulting from the ontology matching process according to common evaluation metrics.
Experimental results clearly show that the automated matching process inevitably introduces errors and inaccuracies, resulting in significant uncertainty, and demonstrate the fundamental unreliability of fully automated integration solutions. We believe that, in general terms, properly designed semi-supervised integration approaches could be effective even at an application level.

Overall, this paper reviews existing automatic ontology matching approaches focusing on the aspect of uncertainty, performs experiments on a practical case study of knowledge-building, extracts the lessons learned from the experimental results regarding uncertainty measures, uncertainty causes, uncertainty situations, and possible uncertainty solutions, and deduces open issues and challenges in this context.

\paragraph{\textbf{Structure of the paper}} 

The following section recalls the notion of \textit{Ontology} which is the main technology used in this paper. Section~\ref{section3} briefly describes an overview of the ontology-based approach to data integration (the knowledge-building process) and summarizes related work on ontology matching. Section~\ref{section4} reviews in detail the related work on uncertainty and uncertainty management in ontology matching. It explains the reasons for the situations of uncertainty in ontology matching and presents the different approaches to resolve (or at least reduce) them. Section~\ref{section5} gathers all the traditional evaluation metrics used in the literature to measure uncertainty in ontology alignments. The core part of the paper is composed of two sections, which respectively deal with (\textit{i}) the description of the experiments carried out in the case study and the analysis of the experimental results (Section~\ref{section6}), and (\textit{ii}) the discussion of issues and challenges faced in these experiments, the lessons learned derived from the case study, and some avenues for future research directions (Section~\ref{section7}). Finally, as usual, the paper includes a conclusion section that also briefly discusses our future work.

\section{Key Notion: Ontology\label{section2}}


Ontology is a logic-based, rich data model that formally defines and describes the terms of a particular vocabulary (and their relationships) in a machine-interpretable format to provide a common understanding of a given domain. 

An ontology can be interpreted as a tuple $o =\, <\mathcal{C}, \mathcal{I}, \mathcal{P}, \mathcal{V}, \mathcal{A}x>$, such that $\mathcal{C}$ is the set of concepts (or classes of individuals, or classes); $\mathcal{I}$ is the set of individuals (or instances/objects); $\mathcal{P}$ is the set of properties (or relations) which is divided into two sets: $\mathcal{OP}$ is the set of object properties (or relationships/associations), and $\mathcal{DP}$ is the set of datatype properties (or attributes); $\mathcal{V}$ is the set of datatype values (or data values, or data literals) specified by data types; and $\mathcal{A}x$ is the set of axioms, such as axioms of \textit{subsumption} (\textit{a.k.a.} \textit{inclusion}, "\textit{is-a}", child--parent, sub-entity--super-entity, hyponymy--hypernymy, specialization--generalization) between two concepts or two properties; axioms of \textit{instantiation} (or \textit{typing}) between concepts and individuals, properties and property instances, data types and data values; axioms of \textit{\textit{disjointness}} (or \textit{exclusion}) between two concepts or two properties; axioms of \textit{equivalence} (or \textit{assignment}) between two concepts or two properties; as well as other logical axioms such as restrictions on properties and complex relations.

Instantiation axioms between classes and their individuals are also called \textit{class assertions}. For example, the individual "\textit{Italy}" is an instance of the class "\textit{Country}" ($<Italy \longrightarrow type \longrightarrow Country>$); and the individual "\textit{Rome}" is an instance of the class "\textit{City}" ($<Rome \longrightarrow type \longrightarrow City>$).
Instantiation axioms between properties and their property instances are also called \textit{property assertions}. An \textit{object property assertion} means that an object property links an individual of a given class (called \textit{domain}) to an individual of a given class (called \textit{range}). For example, "\textit{Rome}" is the capital of "\textit{Italy}" ($<Rome \longrightarrow capitalOf \longrightarrow Italy>$);
While a \textit{datatype property assertion} means that a datatype property links an individual of a given class (called \textit{domain}) to a data value of a given data type (called \textit{range}), e.g., integer, string, boolean, real, etc. For example, "\textit{Rome}" has a population of "\textit{$4.400.000$}" which is a value of the data type \textit{Integer} ($<Rome \longrightarrow has\_population \longrightarrow ``4.400.000">$). The subject of a property is called its domain, and the object of a property is called its range ($<\textit{domain} \longrightarrow \textit{property} \longrightarrow \textit{range}>$).

Thus, an ontology is a set of triplets $<subject \longrightarrow predicate \longrightarrow object>$ (or $<entity_1 \longrightarrow relation \longrightarrow entity_2>$). And it can also be viewed as a directed labeled graph, such that entities are \textit{nodes}, and relations are \textit{edges}.

The set of \textit{classes}, \textit{object properties}, \textit{datatype properties}, \textit{individuals}, and \textit{data values} is called \textit{entities} (or \textit{resources} ---except for \textit{data values} because they do not have identities (\textit{i.e.}, unique identifiers)---).

\section{Ontology Matching in the Knowledge-Building Process\label{section3}}





The knowledge-building process, as understood in this paper, is not limited to data integration; it also includes semantic enrichment and annotations. Different ontology-based solutions have been proposed to integrate data within a range of scientific and business contexts~\cite{gardner2005ontologies,smith2007obo,zhang2018ontology},
as well as to support the integration among systems~\cite{mate2015ontology}.
The process can be centralized, meaning that a global schema can be adopted to provide integrated access to information~\cite{gardner2005ontologies}.

\subsection{Knowledge-Building Process}

\begin{figure}[!ht]
\centering
\includegraphics[scale=0.45, frame]{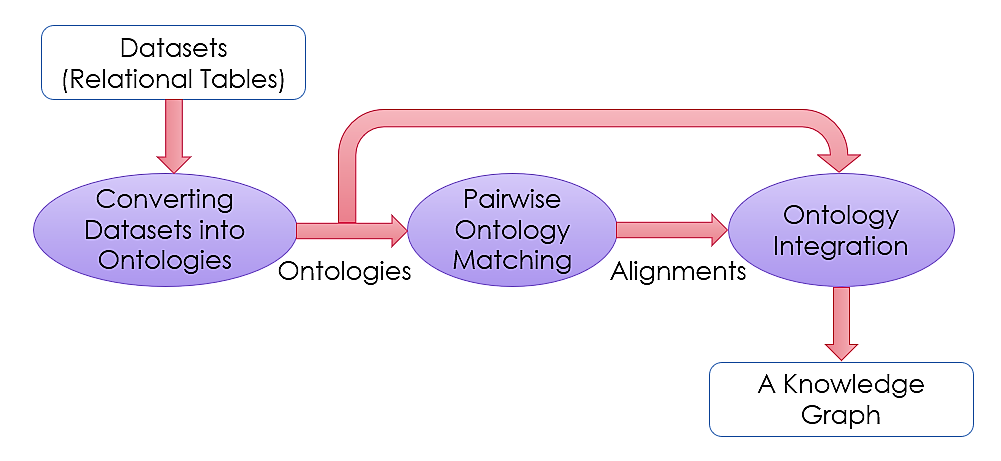}
\caption{Knowledge-Building Process Driven by Data Integration.}
\label{overview}
\end{figure}

Knowledge building through data integration is understood as the process of semantically integrating raw data~\cite{lenzerini2002data}. Ontologies can be used to define rich data spaces (knowledge) in which semantics are formally specified. In this work, the knowledge-building process takes as input heterogeneous raw datasets (assumed to be a relational database) and returns as output an integrated semantic data space represented by a knowledge graph. As shown in Figure~\ref{overview}, this process is composed of the three following main steps: \textit{i})~Conversion of datasets into ontologies, \textit{ii})~Ontology matching, and \textit{iii})~Ontology integration (or ontology merging).

In other words, the system takes datasets (one or more tables according to the classical relational model) as input. The considered datasets provide different information about a domain, with some overlapping. First, if datasets are not available in a semantic format (i.e., RDF or OWL), they are converted into ontologies. Second, semantic matching between each pair of ontologies is performed to obtain pairwise alignments containing correspondences between the equivalent entities. Third, ontologies are aggregated and then merged by adding equivalence links (i.e., equivalence \textit{bridging} axioms) reflecting the alignments’ correspondences. Finally, an integrated ontology (or a knowledge graph) composed of the input ontologies and the equivalence bridge axioms is generated.

A semi-supervised or even fully automated knowledge-building process can be established by using external tools that implement dataset conversion, ontology matching, and ontology integration.

\subsubsection{Dataset Conversion}

By adopting semantic web technology, physical integration is implicitly supported since data available in a semantic format can be systematically and automatically added to the data space. This assumes that datasets and alignments are already imported into the semantic space. However, the conversion of data into a semantic format is not always an obvious step, especially for non-technical users. To properly support this conversion process, the virtual table model can be adopted. The latter is a simple and intuitive approach to data integration that assumes the target dataset is described in relational tables and automatically translated in a semantic format. From a user perspective, an external dataset may be mapped into a virtual table and automatically converted to OWL. A relational table is converted to an ontology, as follows:

\begin{itemize}
 \item The \textit{ID} (or the \textit{primary key}) in the relational model is converted into a \textit{class}/\textit{concept} in the ontology model. In this case, the primary key should not be a composed field (i.e., composed of multiple fields).
 
  \item \textit{Associations} (or \textit{foreign keys}) in the relational model are converted into \textit{object properties} in the ontology model.
 
  \item \textit{Attributes} (or \textit{data fields}) in the relational model are converted into \textit{datatype properties} in the ontology model.

  \item Key values (\textit{Data}) in the relational model are converted into \textit{individuals}/\textit{instances} in the ontology model.
 
 \item Attribute values (\textit{Data}) in the relational model are converted into \textit{data values} (or \textit{literals}) in the ontology model.
\end{itemize}

\subsubsection{Ontology Matching}

\paragraph{Ontology \textit{matching}} Also known as \textit{ontology alignment}, ontology matching is the process of finding semantic correspondences (mainly similarities) among entities from different ontologies (commonly two ontologies). Each type of entity (\textit{classes}, \textit{object properties}, \textit{datatype properties}, and \textit{instances}) is matched in isolation, such that no \textit{class-to-property}, \textit{class-to-individual} or \textit{property-to-individual} correspondences are found. Entity pairs that have the same name and meaning or have different names but the same meaning should be matched. Ontology matching is an essential preceding step for ontology integration. Current ontology matching tools have become proficient in identifying \textit{equivalence} correspondences between two ontologies. Ontology matching systems use different matching algorithms called \textit{ontology matchers}.

\paragraph{Semantic correspondence} Given two ontologies $\mathcal{O}_1$ and $\mathcal{O}_2$, a correspondence is a triple $<e_{\mathcal{O}_1}, r,\, e_{\mathcal{O}_2}>$. Formally, a correspondence is a 4-tuple  $c = \, <e_{\mathcal{O}_1}, e_{\mathcal{O}_2}, r, \theta>$, such that $e_{\mathcal{O}_1}$ is an entity from $\mathcal{O}_1$ and $e_{\mathcal{O}_2}$ is an entity from $\mathcal{O}_2$; $r$ is a semantic relation between $e_{\mathcal{O}_1}$ and $e_{\mathcal{O}_2}$ such as \textit{equivalence} ($\equiv$), \textit{subsumption} ($\sqsubseteq$/$\sqsupseteq$), \textit{disjointness} ($\perp$), \textit{instantiation} ($\in$), or \textit{overlap} ($\between$); and $\theta$ is a confidence value (or a confidence score) that assigns a correctness degree on the identified relation and typically ranges in the interval $[0-1]$. The higher the confidence value, the more likely the correspondence holds~\cite{euzenat2013ontology}. The confidence value of a given correspondence reflects the matcher's "belief" in the correctness (or reliability) of that correspondence~\cite{2006}.
In the case of \textit{equivalence} relation, $c$ is denoted by $<e_{\mathcal{O}_1} \equiv\, e_{\mathcal{O}_2}>$, and $\theta$ reflects the \textit{similarity value} (\textit{a.k.a.} the \textit{similarity measure}, the \textit{similarity score}, or the \textit{similarity degree}). 

\paragraph{Ontology alignment} Denoted as $\mathcal{A} = \{c_1, c_2, \, \ldots, c_n\}$, an ontology alignment is a set of semantic correspondences relating entities from two ontologies. It is normally understood as the result of an ontology-matching process. Alignments are usually expressed in the RDF Alignment format\footnote{\url{https://moex.gitlabpages.inria.fr/alignapi/format.html}}~\cite{2007_2} which is the standard format for representing ontology alignments. In general, an alignment is considered a \textit{partial} \textit{many-to-many} alignment. Indeed, in a \textit{partial} alignment, there could be many entities in $\mathcal{O}_1$ or $\mathcal{O}_2$ that have no counterpart (equivalent) entities in the other ontology, whereas in a \textit{many-to-many} alignment, an entity from one ontology ($\mathcal{O}_1$ or $\mathcal{O}_2$) can be matched to one or many entities from the other ontology. A
\textit{many-to-many} alignment is both a \textit{one-to-many} alignment and a \textit{many-to-one} alignment.

\subsubsection{Ontology Integration}

The simplest way for integrating ontologies is the \textit{Simple Union}~\cite{raunich2012towards} or the \textit{simple merge}~\cite{osman2021ontology,ines} approach. It consists of aggregating the input ontologies or importing them into a new ontology and adding \textit{bridging} axioms that translate the alignment(s) between them. The semantic correspondences of each alignment (resulting from the matching step) are transformed into \textit{bridging} axioms in order to link the overlapping parts of the input ontologies~\cite{predoiu2005d4,de2006ontology}.

The integration of two ontologies can be formalized by the following \textit{merge} function: \textit{Merge($\mathcal{O}_1$, $\mathcal{O}_2$, $\mathcal{A}$) = $\mathcal{O}_3$}~\cite{euzenat2013ontology}, such that $\mathcal{O}_1$ and $\mathcal{O}_2$ are two ontologies to be integrated, and $\mathcal{A}$ is the alignment between them.

OWL (Ontology Web Language)~\cite{mcguinness2004owl,motik2009owl} is the most widely used language for representing and defining structured knowledge in the Semantic Web~\cite{berners_Lee_2001,shadbolt2006semantic}. It uses the rich formal semantics of Description Logics (DL)~\cite{baader2003description} to express ontologies and reason on them. The OWL language provides direct mechanisms to link equivalent entities. Indeed, \textit{equivalence} correspondences between entities can be expressed by OWL built-in statements or axioms, as follows:

\begin{itemize}
    \item Equivalence between \textit{classes} is expressed by the \texttt{<owl:equivalentClass>} axiom;
    
    \item Equivalence between \textit{object} or \textit{datatype properties} is expressed by the \texttt{<owl:equivalentProperty>} axiom;
    
    \item Equivalence between \textit{individuals} (rather called "identity") is expressed by the \texttt{<owl:sameAs>} axiom.
\end{itemize}
Therefore, when we integrate two ontologies, the integrated ontology $\mathcal{O}_3$ can be considered as the aggregation or the union of $\mathcal{O}_1$, $\mathcal{O}_2$ and $\mathcal{O}_\mathcal{A}$ (i.e., $\mathcal{O}_3 = \mathcal{O}_1 \cup \mathcal{O}_2 \cup \mathcal{O}_\mathcal{A}$)~\cite{jimenez2009ontology}. The correspondences of the alignment $\mathcal{A}$ can be viewed as an ontology $\mathcal{O}_\mathcal{A}$ (\textit{a.k.a.} an \textit{intermediate ontology}~\cite{kalfoglou2003ontology}, a \textit{bridge ontology}~\cite{kim2005moa}, or an \textit{articulation ontology}~\cite{mitra2000graph}).

The resulting ontology $\mathcal{O}_3$ can be called an \textit{integrated ontology} or a \textit{merged ontology}. It can also be called a \textit{knowledge graph} (KG) because a KG is composed of entities from many independent sources and can cover different domains at the same time~\cite{paulheim2017knowledge}.

\subsection{Ontology Matching: Related Work}

Ontology matching is a crucial step in the knowledge-building process. In this subsection, we will briefly report different ontology matching techniques used in the literature.

Real-world ontologies often introduce interoperability issues, as they are designed by different developers and communities with distinct requirements and tools and for different purposes and applications.
Independently developed ontologies that describe similar domains or the same domain often have different conceptual models, different domain perspectives (points of view), different levels of expression detail (granularity), and different naming conventions. In other words, ontologies describing the same piece of knowledge (i.e., the same entities) will often be heterogeneous. To integrate their knowledge, ontologies need to reconcile their heterogeneity by using an ontology matching process.
Because of the compelling need to minimize human intervention and speed up the matching process~\cite{2007}, automatic matching is becoming an essential requirement in many contexts and applications. Due to its unquestionable relevance in data integration within modern systems, ontology matching has been extensively discussed in the literature in terms of approaches, issues, solutions, challenges, and assessment (\textit{e.g.}~\cite{choi2006survey,euzenat2013ontology,granitzer2010ontology,kalfoglou2003ontology,ochieng2018large,otero2015ontology,rahm2011towards,shvaiko2013ontology,mohammadi2020evaluating}).

The most common classification for ontology matching solutions distinguishes between \textit{content-based techniques} and \textit{context-based techniques}~\cite{euzenat2007ontology,euzenat2013ontology}. While the former class adopts only internal knowledge contained in the ontologies to be matched, the latter relies on external (or contextual) resources different from the target ontologies (e.g., linguistic resources and external vocabularies).
Another possible classification distinguishes between \textit{element-level} techniques (where entities are considered in isolation) and \textit{structure-level} techniques (where entities are considered along with their relationships to other entities in the ontology).


\paragraph{Content-Based Ontology Matching}

Different types of solutions adopt a content-based approach as follows: 

\begin{itemize}
 
\item \textit{Terminological techniques} consider entities as strings or words.

\begin{itemize}
 
\item \textit{String-based} techniques adopt the similarity among strings, using string measures (or string distance functions) such as Jaccard, Euclidean, n-grams, Levenshtein, Jaro-Winkler, TF-IDF, \textit{etc}.

\item While \textit{language-based} or \textit{lexical} techniques adopt the similarity among words, using Natural Language Processing techniques (such as tokenization, lemmatization, and stopword removal, \textit{etc.}), or using external resources (such as domain-specific thesauri, lexicons, and dictionaries (\textit{e.g.} \href{https://wordnet.princeton.edu/}{WordNet}~\cite{miller1995wordnet}, \href{https://www.nlm.nih.gov/research/umls/index.html}{UMLS}~\cite{bodenreider2004unified})). External resources are used in order to find similarities between words based on linguistic relations between them, \textit{e.g.}, \textit{synonymy}, \textit{hypernymy}, and \textit{hyponymy} relations.
These techniques are usually applied before running the \textit{string-based} techniques.
\end{itemize}

\item \textit{Structural techniques} rely on the (tree-like hierarchy) structure of the ontology and are based on the \textit{principle of locality\label{locality}}~\cite{jimenez2011logmap}. The latter assumes that entities adjacent or neighbors to entities of correct correspondence are likely to be matched to each other. The hierarchy neighbors of a given entity are its (direct) parents, (direct) children, siblings, and leaves. For example, if entities $e_1$ from $\mathcal{O}_1$ and $e_2$ from $\mathcal{O}_2$ are correctly matched, then the neighbors of $e_1$ are likely to be matched to the neighbors of $e_2$.

\item \textit{Semantic-based techniques} compare the semantic interpretation of the entities to be matched (e.g., using a DL ontology reasoner (by inference or deduction) or using external resources). They are based on intuition, assuming that if two entities share the same interpretation, then they are the same.
It should be noted that \textit{language-based} techniques using external resources (such as WordNet~\cite{miller1995wordnet}) can also be considered \textit{semantic} techniques because linguistic relations (\textit{synonyms}, \textit{hypernyms}, and \textit{hyponyms}, etc.) are also semantic relations.

\item \textit{Instance-based techniques} compute the correspondences by comparing the sets of individuals. They are based on intuition, assuming that if the instances are alike, then the classes to which they belong are also alike.

\item \textit{Constraint-based techniques} are based on the similarity of the internal constraints/structure of the entities (\textit{e.g.}, the domains and ranges of properties, and the data types of datatype properties, \textit{etc.}). They are less popular than the previous ones and are commonly used in combination with others.

\end{itemize}

\paragraph{Context-Based Ontology Matching}

Context-based techniques are normally classified depending on the external resources they adopt:

\begin{itemize}
\item "Formal resource-based techniques" use formally represented resources to do the matching. These resources are usually external ontologies (e.g., upper-level ontologies or domain-specific ontologies), alignments from previously matched ontologies, linked data, or instances that are not part of the ontology (e.g., knowledge graphs), etc.

It should be noted that \textit{language-based} techniques using external resources (such as WordNet~\cite{miller1995wordnet}) can also be considered \textit{formal resource-based} techniques because linguistic or semantic relations (\textit{synonyms}, \textit{hypernyms}, and \textit{hyponyms}, \textit{etc}.) can be found in an external ontology or in any external formal resource.

\item \textit{Informal resource-based techniques} use more informal resources to perform the matching process, such as linguistic resources (\textit{e.g.}, dictionaries) or annotated resources (e.g., encyclopedia pages and pictures). Similarities among entities are based on how these entities are related to such resources. For example, if two classes annotate the same set of pictures, these classes can be considered equivalent. In this case, to deduce reliable correspondences, informal resources are typically large corpora of related entities.

\end{itemize}

Within an automatic process of ontology matching, it has been shown that a combination of complementary matchers (\textit{string-based}, \textit{lexical}, \textit{structural}, \textit{semantic}, and \textit{instance-based} matchers) improves the alignment quality~\cite{2011}. Nowadays, many ontology matching tools use the technique of \textit{matcher ensembles} (a.k.a. \textit{matcher combination}, or \textit{matcher aggregation}) to produce optimal results.

\section{Ontology Matching: Uncertainty \& Uncertainty Management\label{section4}}

\subsection{Ontology Matching: Uncertainty}

The ontology matching process is the main cause of uncertainty in the knowledge-building process.
Once the systems scale up, the knowledge-building process using existing consolidated automatic tools is required.
This automation generates uncertainty. Uncertain information is generally understood as information whose veracity cannot be determined or assessed~\cite{2008,2007_2}. However, uncertainty is an intrinsic and unavoidable circumstance in automatic data integration and a critical issue for the reliability of systems and underlying processes.
In the context of this work, the focus is on uncertainty in schema matching (and in particular in ontology matching).

We notice two types of uncertainty in ontology matching:~(\textit{i}) uncertainty caused by the domain of discourse, and (\textit{ii}) uncertainty caused by the matcher combination strategy.

\subsubsection*{Uncertainty Caused by the Ontology Domain}

While fully automatic ontology matching tools intrinsically lead to situations of uncertainty (as human experts do not verify them), even semi-supervised or manual approaches may lead to errors. Indeed, the users may not sufficiently understand the ontology domain (such as \textit{bioinformatics} or \textit{biomedical} domains) and therefore may provide imprecise and incorrect correspondences.
Additionally, some domains can be intrinsically error-prone, as the aimed correspondences could not be clear~\cite{2009}.
For example, concepts can have ambiguous semantics when they are closely related but neither completely synonymous (i.e., equivalent) nor completely dissimilar (i.e., disjoint)~\cite{2011,2011_2,2010}. In this case, matching systems become uncertain as it is not accurate to state whether two entities are equivalent or not~\cite{2014}.

Besides, in an ontology, entities are not supposed to be semantically disjoint by default. Entities follow the "Open World Assumption" where they are supposed to overlap and share a certain amount of common information (even if it is not yet specified)~\cite{2014}. Ontologies belonging to the same domain or to similar domains often introduce \textit{linguistic}, \textit{structural}, and \textit{semantic} ambiguities, resulting from their different domain representations~\cite{2012}. These ambiguities will lead to the creation of an ambiguous alignment after performing the ontology matching process.

\label{ambiguité}
Let us introduce the notion of "\textit{alignment ambiguity}". A \textit{many-to-many} alignment is actually an ambiguous alignment since it contains ambiguous correspondences~\cite{euzenat2013ontology}. Ambiguous correspondences are also called \textit{competing} correspondences~\cite{faria2013agreementmakerlight}, or \textit{higher-multiplicity} correspondences (or correspondences of \textit{higher multiplicity})~\cite{stoilos2018novel}. The latter are sets of correspondences that match one entity from the first ontology to multiple entities from the second ontology, or vice versa~\cite{euzenat2013ontology}. These correspondences have in common the same \textit{source} entity (coming from $\mathcal{O}_1$) or the same \textit{target} entity (coming from $\mathcal{O}_2$). In other words, an ambiguous correspondence contains at least an entity (coming from $\mathcal{O}_1$ or $\mathcal{O}_2$) that appears in other correspondences. Here is an example of a set of three ambiguous \textit{equivalence} correspondences written in the DL syntax:
    \begin{center}
      {\ttfamily $\mathcal{O}_1$:Student $\equiv$ $\mathcal{O}_2$:Student
      
      $\mathcal{O}_1$:Scholar $\equiv$ $\mathcal{O}_2$:Student
      
      $\mathcal{O}_1$:PhD\_Student $\equiv$ $\mathcal{O}_2$:Student
      }
    \end{center}

Klir and Yuan~\cite{klir} distinguish two main types of uncertainty:~\textit{Fuzziness} (lack of sharp or definite distinctions) and \textit{Ambiguity} (co-existence of \textit{many-to-one} or \textit{one-to-many} possible correspondences, leading to a choice disagreement). Ontologies carry this ambiguity in the process of ontology matching as well as the process of ontology integration~\cite{2012}. In addition, such an error rate is likely to increase with the scale of the target system. An uncertain matching affects the performance of the whole system~\cite{2014}.


\subsubsection*{Uncertainty Caused by the Matcher Combination}

\textit{Lexical} matchers assume that similar entities are likely to have similar names, while \textit{structural} matchers assume that similar entities are likely to have similar hierarchy neighbors (i.e., sub-entities, super-entities, etc.); \textit{Semantic} matchers assume that similar entities are likely to have the same meaning; In contrast, \textit{instance-based} matchers assume that similar concepts are likely to have similar individuals~\cite{2007}. To aggregate these assumptions by which different matchers assess the similarity between entities, a "combined matcher" automatically combines the alignments produced by all matchers and returns a single alignment. The latter contains correspondences with overall confidence values resulting from the combination of multiple confidence values generated by individual matchers~\cite{2003}.
Confidence values can be combined by using one of the following strategies: \textit{min}, \textit{max}, \textit{average}, or \textit{weighted sum}. Combination algorithms often assign a weight to each matcher. The confidence values of the combined correspondences strongly depend on the choice of the matcher weights~\cite {2014}.

Nevertheless, when the number of individual matchers increases, it becomes trickier to combine their respective alignments. The actual challenge in any system that manages uncertainty is to check the reliability of the generated similarity scores (or confidence values). Indeed, unreliable values reduce the quality of the alignment, as erroneous correspondences could falsely have high similarity scores. Getting reliable and meaningful confidence values is always sought after, and it is one of the most challenging issues~\cite{2009}. The performance of automated matching systems heavily depends on the validity of the combination strategy, which is hard to design. The \textit{average} combination is often considered to be the most effective strategy. Overall, neglecting the importance of correspondence similarity values does generate uncertainty in ontology alignments and deteriorates the quality of the ontology integration process~\cite{2010}.

\subsection{Ontology Matching: Uncertainty Management}\label{sec:uncertainty}

Uncertainty management plays a crucial role in real-world applications. It has been recognized as the next issue in data integration~\cite{2006}. Managing uncertainty in schema matching is the problem of dealing with imprecise or inaccurate correspondences~\cite{2007_2,2007}.
Non-supervised matching is often imperfect~\cite{2011_2} as no automatic matching tool can be expected to produce the exact expected alignment (by finding all the correct correspondences between entities)~\cite{2011}. This automatic process always brings with it a degree of uncertainty (imperfection or imprecision)~\cite{2003}. Therefore, it has become evident that we should manage partially incorrect alignments in order to improve their quality~\cite{2007_2}.

Related work on data integration manipulates uncertainty until a choice is made to keep only exact information. There are two options for reducing (or removing) uncertainty in ontology alignments:~(\textit{i}) either manually, through \textit{user feedback}, or (\textit{ii}) automatically, using \textit{alignment filtering} techniques.

In the (\textit{user-driven}) manual approach, correct and incorrect correspondences can be manually selected from the alignment. For example, the user can manually validate all the ambiguous correspondences or all the low-confidence correspondences under a certain confidence threshold (assuming that low-confidence correspondences are less reliable than high-confidence ones) by keeping the correct correspondences and deleting the false ones. In addition, the user can alter correspondences by changing their relation type (e.g., from \textit{equivalence} to \textit{subsumption}) or by changing their confidence value (e.g., from a low value to a high value if they are falsely assigned to very low confidence values). The user can also add missing correspondences if possible. Otherwise, users can be incorporated into a semi-supervised matching process whenever the system asks for help~\cite{2003}.

In the (\textit{alignment filtering}) automatic approach, two different levels of uncertainty management can be highlighted:~(\textit{i}) the \textit{alignment trimming}, and (\textit{ii}) the \textit{alignment disambiguation}.

\subsubsection*{Alignment Trimming\label{threshold}}

The \textit{alignment trimming}~\cite{faria2013agreementmakerlight} approach is also called \textit{thresholding}~\cite{shvaiko2013ontology} or \textit{confidence cut}~\cite{david2011alignment}. It aims to trim the alignment with a threshold in order to ensure that only the best correspondences are maintained in the alignment. The \textit{trimming} process removes from the alignment all correspondences whose confidence values are below a given threshold. Formally, this approach discards all correspondences with a confidence value $c < \alpha$ by applying an $\alpha$-cut to the alignment, such that $\alpha$ is the confidence threshold and $\alpha \in [0, 1]$. This technique increases the alignment \textit{precision} at the expense of the alignment \textit{recall} (see  Subsection~\ref{advanced}).

Semi-supervised user feedback cycles or automatic learning-based approaches both choose the threshold. In the former way (which is the most common), many trials should be performed by adjusting or varying the threshold value until the best or optimal threshold is found~\cite{2011}.

Assigning a given confidence threshold depends on the application in question. For instance, we can assign relatively low thresholds to a recommendation application since incorrect correspondences are tolerated; however, we should assign high thresholds to a scientific application since incorrect correspondences are not acceptable~\cite{2014}. Reducing the alignment uncertainty by \textit{alignment trimming} generates a loss of information~\cite{2011}, as some of the deleted correspondences can be correct. Indeed, it is impossible to separate correct correspondences from incorrect correspondences using a trimming threshold because correspondences under any threshold can include correct and incorrect correspondences. As a result, any selection of a threshold will still generate false correspondences and/or missing correspondences. As a result, choosing a threshold and performing a trimming process alone will not yield the ideal alignment~\cite{2006}.

As a rule of thumb, the uncertainty resulting from the ontology matching process is either lost or transformed into exact information~\cite{2007_2}. Hence, to avoid losing relevant information from the alignment, we can preserve these uncertain correspondences and manage uncertainty manually~\cite{2008}.

\subsubsection*{Alignment Disambiguation}\label{ambiguity}

Ambiguous correspondences are often a source of uncertainty in ontology alignments because of their ambiguous interpretation (see Subsection~\ref{ambiguité}). There are two possible approaches for disambiguating a set of ambiguous \textit{equivalence} correspondences:

\bigskip

\textbf{(\textit{i})}~In the first approach, an alignment disambiguation process converts a \textit{many-to-many} alignment into a \textit{one-to-one} alignment. It is considered a \textit{bipartite matching} problem (or a \textit{bipartite filtering} problem). This approach consists of selecting the most similar pair of entities (i.e., the correspondence that has the highest confidence value) and removing the remaining correspondences that involve one of these entities. It assumes that only one correspondence reflects a correct semantic equivalence, while the other ones (i.e., those with lower confidence values) are incorrect correspondences that do not reflect a strict synonym but rather an overlap relation~\cite{stoilos2018novel}. Therefore, only one \textit{equivalence} correspondence is considered among the set of ambiguous correspondences. To do so, we can apply the \textit{Stable Matching} algorithm (or the \textit{Stable Marriage} algorithm)~\cite{gale1962college}. The latter assigns only one object $o_1$ from a set $\mathcal{S}_1$ to only one object $o_2$ from a set $\mathcal{S}_2$, such that there is no other correspondence involving one of the two objects and having a higher similarity. This algorithm favors stronger (high-valued)
individual correspondences—in terms of similarity values. After applying the \textit{Stable Matching} algorithm, there is no object (entity) involved in multiple correspondences.

\medskip

There are other algorithms, such as \textit{Maximum Weight Matching} (or \textit{Maximum Weight Bipartite Filtering})~\cite{munkres1957algorithms,cruz2009efficient}. It aims to get a maximum weighted sub-alignment that maximizes the sum of the confidence values of all correspondences constituting the alignment. That is, it tries to maximize the
total similarity values of the selected correspondences. However, we believe that the \textit{Stable Marriage} works better than the \textit{Max Weight Matching} in this case. Indeed, \textit{Stable Marriage} always chooses the best correspondence for each entity in isolation (in a local manner) and thus guarantees that all high-confidence correspondences are selected. While the \textit{Max Weight Matching} chooses the best correspondence for each entity in a global manner, which does not guarantee that all high-confidence correspondences are selected. Assuming the confidence values provided by the matching tool are reliable and truly reflect the probability of correspondences being correct, it is always better to have one high-confidence correspondence than two medium-confidence correspondences in an alignment. Since the goal of \textit{alignment filtering} is to minimize the number of incorrect correspondences and maximize the number of correct ones, we believe that the \textit{Stable Marriage} is the best choice.

\medskip

There is another disambiguation idea that is based on the \textit{principle of locality} (see Subsection~\ref{locality}). It assumes that low confidence values in the neighborhood of a correspondence $c$ can reveal the incorrectness of $c$. For instance, given a correspondence $c$ composed of the entity $e_1$ from $\mathcal{O}_1$ and the entity $e_2$ from $\mathcal{O}_2$ ($c =\, <e_1 \equiv e_2>$), if the correspondences relating the neighbors of $e_1$ and $e_2$ have low confidence values, then the correspondence $c$ is likely to be incorrect. Recall that the neighbors of an entity are the more general entities (ancestors) and/or the more specific entities (descendants) of that entity. For each ambiguous correspondence, this algorithm~\cite{tordai2012combining} counts the confidence proportion of correspondences reachable by the neighbors of the correspondence in question. Then, it selects the correspondence with the highest confidence ratio. This algorithm is not always appropriate, especially when there are no correspondences in the neighborhood of a correspondence and when the hierarchy of ontologies is not deep enough.

\bigskip

\label{approche2}\textbf{(\textit{ii})}~As for the second approach, since the terms of one ontology are more general (or more detailed) than the terms of the other ontology (i.e., one ontology is more general or more granular than the other), ~\cite{stoilos2018novel}, all \textit{equivalence} relations in the ambiguous correspondences are considered \textit{subsumption} relations. In this case, entities of the first ontology are decomposed into several more specific entities in the second ontology, or vice versa~\cite{2006}. Thus, the sets of ambiguous \textit{equivalence} correspondences are actually incorrect. This alignment disambiguation approach transforms every ambiguous \textit{equivalence} correspondence into a \textit{subsumption} correspondence by altering its relation type from the \textit{equivalence} relation ("$\equiv$") to the \textit{inclusion} relation ("$\sqsubseteq$" or "$\sqsupseteq$")~\cite{stoilos2018novel,solimando2017minimizing,arnold2013semantic}.

\bigskip

As a result, we deduce that a generic alignment disambiguation approach is difficult to define.

\section{Ontology Alignment Evaluation Metrics\label{section5}}

Several evaluation metrics have been defined and are commonly adopted within the research community to quantitatively assess the accuracy of alignments resulting from an automatic ontology matching system.

To do so, a \textit{reference} alignment (understood as the expected or intended alignment—the "gold standard") should be available~\cite{do2002comparison}. In practice, an alignment is a set of correspondences between entities from different ontologies. Therefore, an alignment $\mathcal{A}$, returned by a given ontology matching tool, can be compared to the \textit{reference} alignment $\mathcal{R}$ by checking for the overlap of the two sets of correspondences~\cite{ehrig2005relaxed}. In general terms, the most common ontology alignment evaluation metrics are adaptations of classical metrics from the Information Retrieval (IR) field.

\subsection{Basic Evaluation Metrics}

The basic evaluation metrics assume the following definitions~\cite{euzenat2013ontology}:

\begin{itemize}
\item\textbf{True positives} are correspondences that have been correctly found by an automatic ontology matching tool.
\begin{center}
\framebox[0.31\textwidth]{\textbf{True Positives (TP) $= \mathcal{A} \cap \mathcal{R}$}}
\end{center}

\item\textbf{False positives} are correspondences that have been falsely found by an automatic ontology matching tool.

\begin{center}
\framebox[0.31\textwidth]{\textbf{False Positives (FP) $= \mathcal{A} - \mathcal{R}$}}
\end{center}

\item\textbf{False negatives} are correct correspondences that have not been found by an automatic ontology matching tool.

\begin{center}
\framebox[0.32\textwidth]{\textbf{False Negatives (FN) $= \mathcal{R} - \mathcal{A}$}}
\end{center}

\item \textbf{True negatives} are false correspondences that have been correctly discarded by an automatic matching tool.

\begin{center}
\framebox[0.61\textwidth]{
\begin{minipage}{\linewidth}
\centering
\textbf{True Negatives (TN) $= (E \times E^{'}) - (\mathcal{A} \cup \mathcal{R})$}

$E$ and $E^{'}$ = number of entities in the input ontologies $\mathcal{O}_1$ and $\mathcal{O}_2$. 

$(E \times E^{'})$ = set of all possible correspondences between $\mathcal{O}_1$ and $\mathcal{O}_2$.
\end{minipage}
}
\end{center}
\end{itemize}

Based on such definitions, the set of automatically identified correspondences comprises true positives and false positives ($\mathcal{A} = TP + FP$), and the set of expected correspondences is composed of true positives and true negatives ($\mathcal{R} = TP + TN$). Evidently, false positives and false negatives reduce the matching accuracy. Therefore, an efficient ontology matching system aims to minimize both of them.

\subsection{Advanced Evaluation Metrics\label{advanced}}
The advanced evaluation metrics assume the following definitions ~\cite{ehrig2005relaxed,euzenat2007semantic,euzenat2013ontology,mohammadi2020evaluating}:

\begin{itemize}
\item \textbf{Precision} measures the "correctness" of an alignment. It reflects the share of correct correspondences among all the ones found. It is defined as the ratio of the number of correctly found correspondences (TP) over the total number of found correspondences (TP + FP). Given a \textit{reference} alignment $\mathcal{R}$, the \textit{Precision} of an alignment $\mathcal{A}$ is a function $P : \land \times \land \rightarrow [0, 1]$, such that:
\begin{equation*}
Precision = P(\mathcal{A}, \mathcal{R}) = \frac{|\mathcal{A} \cap \mathcal{R}|}{|\mathcal{A}|} = \frac{|TP|}{|TP| + |FP|}
\end{equation*}

\item \textbf{Recall} measures the "completeness" of an alignment. It reflects the share of correct correspondences among all the expected correspondences. It is the ratio of the number of correctly found correspondences (TP) over the total number of expected correspondences to be found (TP + TN). Given a \textit{reference} alignment $\mathcal{R}$, the \textit{Recall} of an alignment $\mathcal{A}$ is a function $R : \land \times \land \rightarrow [0, 1]$, such that:
\begin{equation*}
Recall = R(\mathcal{A}, \mathcal{R})= \frac{|\mathcal{A} \cap \mathcal{R}|}{|\mathcal{R}|} = \frac{|TP|}{|TP| + |TN|}
\end{equation*}
In the ideal case, \textit{Precision} and \textit{Recall} reach their highest value of $1.0$.

\item \textbf{Noise \& Silence} are the complement measures of \textit{Precision} and \textit{Recall}. Given a \textit{reference} alignment $\mathcal{R}$, the \textit{Noise} and the \textit{Silence} of an alignment $\mathcal{A}$ are functions $N$ and $S : \land \times \land \rightarrow [0, 1]$, such that:
\begin{equation*}
Noise = N(\mathcal{A}, \mathcal{R})= 1 - Precision
\end{equation*}
\begin{equation*}
Silence = S(\mathcal{A}, \mathcal{R})= 1 - Recall
\end{equation*}

\item \textbf{F-measure} combines \textit{precision} and \textit{recall}, as they cannot accurately assess the matching quality alone. Indeed, the ontology matching tool can have a high \textit{precision} and a low \textit{recall}, and vice versa. \textit{F-measure} is a combined metric that attaches different importance to \textit{precision} and \textit{recall}. Given a \textit{reference} alignment $\mathcal{R}$ and a number $\alpha$ between $0$ and $1$ ($0 \leqslant \alpha \leqslant 1$), the \textit{F-measure} of an alignment $\mathcal{A}$ is a function $F_\alpha: \land \times \land \rightarrow [0, 1]$, such that:
\begin{equation*}
F\text{-}measure\big(\alpha\big) = F_\alpha(\mathcal{A}, \mathcal{R}) = \frac{Precision \times Recall}{(1 - \alpha) \times Precision + \alpha \times Recall}
\end{equation*}

The $\alpha$ parameter defines the relative balance between \textit{precision} and \textit{recall}, as higher $\alpha$ values give more importance to \textit{precision} with respect to \textit{recall}. When $\alpha = 1$, \textit{F-measure} is equal to \textit{Precision}; and when $\alpha = 0$, \textit{F-measure} is equal to \textit{Recall}. A value of $\alpha = 0.5$ defines the equal importance of both core metrics. Therefore, when $\alpha = 0.5$, \textit{F-measure} becomes the harmonic mean of \textit{precision} and \textit{recall}, as follows:
\begin{equation*}
    F1 = F\text{-}measure\big(0.5\big) =  F_{0.5}(\mathcal{A}, \mathcal{R}) = 2 \times \frac{Precision \times Recall}{Precision + Recall}
\end{equation*}

$F\text{-}measure(0.5)$, also called $F1$, is the most commonly used variant of $F\text{-}measure(\alpha)$ in IR since it balances the importance of \textit{precision} and \textit{recall} so that neither is compensated by the other.

Matching tools may need parameter tuning. In this case, \textit{F-measure} is adopted as a driving factor to perform parameter tuning because values that maximize \textit{F-measure} are considered the optimal ones. Hence, this metric is not only helpful in evaluating the quality of alignments but also in selecting input parameters for matching systems, such as the confidence threshold that returns the best \textit{F-measure} value (see Subsection~\ref{threshold}).

\item \textbf{Overall} (or \textit{matching accuracy}~\cite{melnik2002similarity}) is explicitly developed for schema matching purposes, unlike the previously mentioned metrics. It measures the manual error correction effort. That is, it reflects the post-matching effort needed to fix the alignment by adding missing correspondences (FN) and removing false correspondences (FP).
It is the ratio of the number of errors (FP + FN) over the total number of expected correspondences (TP + TN).
Given a \textit{reference} alignment $\mathcal{R}$, the \textit{Overall} of an alignment $\mathcal{A}$ is a function $O : \land \times \land \rightarrow [-\infty, 1]$, such that:
\begin{equation*}
    Overall = O(\mathcal{A}, \mathcal{R})= Recall \times \Bigg( 2 - \frac{1}{Precision}\Bigg)
\end{equation*}
\begin{equation*}
    = 1 - \frac{ |\mathcal{A}-\mathcal{R}| + |\mathcal{R}-\mathcal{A}|}{|\mathcal{R}|} = 1 - \frac{|FP| + |FN|}{|TP| + |TN|}
\end{equation*}
    
An \textit{overall} value ranges between $[-\infty, 1]$, where negative values are associated with "bad" matching performances. If an alignment $\mathcal{A}$ has a number of false positives higher than the number of true positives ($Precision < 0.5$), its \textit{Overall} will have a negative value, which means that the alignment $\mathcal{A}$ is not worth the repair effort. Indeed, if more than half of the correspondences of $\mathcal{A}$ are false, the user would make less effort to manually match the ontologies from scratch than to manually modify the alignment of $\mathcal{A}$.

In the ideal case, when $Precision = Recall = 1$, \textit{F-measure} and \textit{Overall} reach their highest value of $1.0$. It should be noted that \textit{Overall} values are always lower than \textit{F-measure} values~\cite{euzenat2013ontology}.

\end{itemize}

\subsection{Ambiguity Evaluation Metric}

All the metrics mentioned above reflect the \textit{correctness} and the \textit{completeness} of the generated alignment. However, the following metric reflects the \textit{ambiguity} of the generated alignment.

\begin{itemize}
   \item \textbf{Ambiguity Degree}~\cite{euzenat2013ontology} measures the "ambiguity" of an alignment. It is the proportion of ambiguous correspondences (i.e., entities that are matched to several entities). In other words, it is the proportion of entities from $\mathcal{O}_1$ that are matched to at least two entities from $\mathcal{O}_2$, and vice versa. The number of ambiguous correspondences in an alignment ($\#Amb$) is considered an absolute metric that varies according to the size of the alignment. Therefore, it is preferred to use a relative metric (\%) reflecting the percentage of ambiguous correspondences in an alignment (regardless of the size of the alignment). The \textit{Ambiguity Degree} of an alignment $\mathcal{A}$ is a function $Ambiguity : \land \times \land \rightarrow [0\%, 100\%]$, such that:
    \begin{equation*}
    Ambiguity\,\,degree = Ambiguity(\mathcal{A})= \frac{\Big(\big|\#Amb(\mathcal{A})\big| \times 100\Big)}{\big|\mathcal{A}\big|} = \frac{\Big(\big|\#Amb(\mathcal{A})\big| \times 100\Big)}{\big|TP\big| + \big|FP\big|}
    \end{equation*}
   
If the alignment has an \textit{Ambiguity degree} of $0\%$, this means that it is not an ambiguous alignment. Otherwise, it is an ambiguous alignment.

\end{itemize}

\section{Case Study:~Spatio-Temporal Alignment of Global Indicators\label{section6}}

We propose a classic case study that integrates several independent global indicators into a unique semantic data space. Successful integration is expected to provide a consistent view of the different indicators along the time and spatial dimensions, and, in general, all concepts should be matched.

\subsection{Experiments Description}

The case study, as proposed in this paper, has been inspired by the famous portal \textit{Our World in Data}~\cite{OurWorldInData}, which publishes and discusses heterogeneous indicators for countries from all over the world. In this portal, the different datasets are published in independent \texttt{csv} files. Still, they are considered integrated as the terminology used and the meaning of the fields in the different files are uniform. Our experiments consider raw data as originally provided by the respective sources, namely the datasets downloaded from their original links, as provided on the Our World in Data website. As such, the datasets in our experiments are actually heterogeneous. We aim for an automatic integration that emulates the integrated version of data published in the Our World in Data portal.



\setcounter{table}{-1}
\begin{table}[!htb]
\centering
\caption{Input Dataset Pairs.}
\label{tab:experimentData}
\begin{adjustbox}{scale=0.9,center}
\begin{threeparttable}
\begin{tabular}{c>{\raggedright}p{0.35\linewidth}l}
\toprule
\textbf{$\,\,\,\,\,$Experiment ID$\,\,\,\,\,$} & \multicolumn{1}{c}{\textbf{Input Dataset 1}}                                                                                 & \multicolumn{1}{c}{\textbf{Input Dataset 2}}                                            \\ \midrule[1.35pt]
Exp. 1\label{exp1}          & Countries of the World~\cite{ref1}                                                      & Country Profile Variables~\cite{ref2}           \\\midrule 
Exp. 2\label{exp2}          & Food Security Indicators (V\_2.6)~\cite{ref3} & Prevalence of Undernourishment~\cite{ref4}                      \\\midrule 
Exp. 3\label{exp3}          & Food Security Indicators (I\_2.3)~\cite{ref5}                                           & Prevalence of Undernourishment ~\cite{ref4}                      \\ \midrule
Exp. 4\label{exp4}          & Food Security Indicators (I\_2.4)~\cite{ref5}                                           & Prevalence of Undernourishment~\cite{ref4}                      \\\midrule 
Exp. 5\label{exp5}          & Macroeconomic Data (GDP)~\cite{ref6}                                 & GDP (current US$\,$\$)~\cite{ref7}                      \\ \midrule
Exp. 6\label{exp6}          & WDI Country~\cite{ref8}                                        & WUP2018 Largest Cities~\cite{ref9}                                 \\ \midrule 
Exp. 7\label{exp7}          & Total Life Expectancy-Historical~\cite{ref10}
& Life Expectancy at Birth~\cite{ref11}                      \\ \midrule
Exp. 8\label{exp8}          & Historical Gas Emissions~\cite{ref12}                                           & List of Electoral Democracies~\cite{ref13}                            \\ \midrule
Exp. 9\label{exp9}          & List of Electoral Democracies~\cite{ref13}                                                                 & Sexual Violence in Childhood~\cite{ref14} \\ \midrule
Exp. 10\label{exp10}         & Historical Gas Emissions~\cite{ref12}                                           & Sexual Violence in Childhood~\cite{ref14} \\ \bottomrule
\end{tabular}
\end{threeparttable}
\end{adjustbox}
\end{table}

In table~\ref{tab:experimentData}, we report the input datasets for the carried-out experiments.
 
Each experiment involves a pair of input datasets, since the matching tools adopted do not directly support the matching of more than two ontologies at a time. Each input dataset represents a relational table (provided in the \texttt{csv} format). As per previous explanations, the systematic integration process is composed of two seamless phases:~(\textit{i}) First, we convert original raw data (\texttt{csv} files) into ontologies (\texttt{owl} files) by using the dataset conversion tool described in Subsection~\ref{conversionTool}; (\textit{ii}) Then, we automatically match each pair of the resulting (converted) ontologies by using the most sophisticated available ontology matching tools (described in Subsection~\ref{sec:tools}), to finally obtain an alignment (i.e., an \texttt{rdf} file) for each matched ontology pair.

The considered pairs can involve datasets belonging to the same domain or datasets from different domains. The latter case is prevalent in the context of overlapping, complementary, or interdisciplinary domains. Datasets in Experiment~1 describe a list of countries and several associated indicators (e.g., region, population, population density, surface area, GDP (Gross Domestic Product), birth/fertility rate, net migration, literacy, etc.).
In Experiments~2, 3 and~4, datasets provide indicators of \textit{food security} and \textit{undernourishment} for different countries in different years.
Experiments~5 and~6 describe \textit{economic indicators} for different countries in different years, while Experiment~7 targets \textit{life expectancy} indicators. Finally, Experiments~8, 9, and~10 perform cross-domain matching (between the domains of \textit{democracy}, \textit{violence against children}, and \textit{CO$_2$ \& greenhouse gas emissions}).

Regardless of the meaning of the data, the original tables report the values of given indicators for different countries in different years. The actual structure may vary from case to case, but, in most cases, it proposes typical patterns used to organize spatio-temporal data. For tables describing a particular indicator, rows represent the names of countries and columns represent years (or intervals of years), or vice versa. In some other cases, tables report the values of more than one indicator for different countries in a single year (or in a single interval of years): Here, rows represent the names of countries, and columns represent indicators. More rarely, tables report the values of multiple indicators for different countries in different years: In this case, rows represent ID numbers (enumerated numbers/indexes), and columns contain a year or interval column, a country column, and indicators' columns.


Overall, the test bed proposed cannot be considered critical, neither in terms of size nor complexity. It is instead characterized by its small scale and low complexity. Input datasets are characterized by several columns that vary from $2$ to $63$ and several rows in the range of $[46-540]$.

After converting and matching the input dataset pairs, we will get the output alignments. First, we will evaluate the quality of the output alignments. Then, we will trim and/or disambiguate these alignments and evaluate them again. We aim to see how the alignment \textit{trimming} and \textit{disambiguation} processes can affect the quality and uncertainty of the output alignments. In the alignment \textit{trimming} process, we will proceed as described in Section~\ref{threshold}. And in the alignment \textit{disambiguation} process, we will apply a personalized simplified version of the \textit{Stable Marriage} approach.

For the alignment \textit{trimming}, we will use the Alignment API\footnote{\url{https://moex.gitlabpages.inria.fr/alignapi/}}~\cite{euzenat2004api,david2011alignment}. The latter is a Java programming interface that facilitates the manipulation and evaluation of ontology alignments written in the RDF Alignment format. Given an alignment and a threshold value as input, the Alignment API automatically trims the input alignment using the predefined method \texttt{cut()} and returns a new trimmed alignment. The trimmed output alignment will only contain correspondences that have a confidence measure greater than or equal to the chosen threshold value.

\begin{figure}[!htbp]
\captionsetup{justification=centering}
\centering
\sbox0{\subfloat[][\textbf{Step 1} --- left to right\label{mapppp1}]{\includegraphics[width=0.2\linewidth]{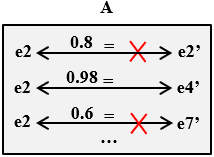}}\quad \vrule \quad
\subfloat[][\textbf{Step 2} --- right to left\label{mapppp2}]{\includegraphics[width=0.2\linewidth]{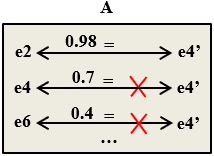}}}
\centering
\begin{minipage}{\wd0}
  \usebox0
  \caption{Alignment Disambiguation (A Simplified Approach for the Stable Marriage~\cite{gale1962college})\label{mapppp}}
\end{minipage}
\end{figure}

For the alignment \textit{disambiguation}, we will apply an approach that is composed of two steps (see Figure~\ref{mapppp}): First, we go through all the correspondences in the alignment, and we disambiguate each set of ambiguous correspondences having a \textit{source} entity in common (coming from $\mathcal{O}_1$), as shown in Figure~\ref{mapppp1}; Second, we go through all the correspondences again, and we disambiguate each set of ambiguous correspondences having a \textit{target} entity in common (coming from $\mathcal{O}_2$), as shown in Figure~\ref{mapppp2}. In each step, disambiguation consists of keeping the strongest correspondence (i.e., the one with the highest similarity score) from the set of ambiguous correspondences and deleting the rest.
If, by chance, two correspondences (from the set of ambiguous correspondences) have exactly the same similarity score, we keep both of them because we cannot randomly choose one over the other. The same thing applies if more than two correspondences happen to have the same similarity score, which is very rare. This alignment disambiguation approach produces the same results as the \textit{Stable Marriage} approach. The disambiguated output alignment does not contain any ambiguous correspondences (i.e., it is composed of only 1-to-1 correspondences).

Finally, a \textit{reference} alignment is manually provided by authors for each target experiment (as in table~\ref{tab:experimentData}) for assessment purposes. Given an alignment (to be evaluated) and a \textit{reference} alignment as input, the Alignment API~\cite{david2011alignment} automatically returns the scores of the basic evaluation metrics (TP, TN, FP, FN) as well as the scores of the advanced evaluation metrics (\textit{Precision}, \textit{Recall}, \textit{F-measure}, \textit{Overall}, \textit{Noise}, and \textit{Silence}) which reflect the quality of the input alignment. As for the ambiguity evaluation metric, we create a simple algorithm in Java that takes as input an alignment and returns the score of the \textit{Ambiguity degree} metric, reflecting the ambiguity degree of that input alignment.

Most (if not all) applications that require the combination of multiple indicators are unlikely to be error-tolerant, meaning that the resulting alignment is expected to be entirely correct. Despite its relative simplicity, the case study proposed is very relevant in different contexts and application domains. More concretely, the target case study is characterized by the need to compose indicators dynamically, and the resulting integration is expected to be precise and accurate. Practical examples may be identified, among others, in the areas of urban planning (\textit{e.g.},~\cite{PILEGGI20172059}) and global sustainable development (\textit{e.g.},~\cite{su12010088}). Further requirements may characterize specific systems, such as real-time environments (in disaster management~\cite{liu2013dynamic}).

\subsection{External Tools Used in the Case Study}

\subsubsection*{Tool for Dataset Conversion\label{conversionTool}}

The dataset conversion tool~\cite{pileggi2020ontological} supports the conversion of a given relational table (\textit{i.e.}, a \texttt{csv} or \texttt{excel} file) into an ontology (i.e., an \texttt{owl} file). It applies the virtual table approach to facilitate such a process, assuming a supervised environment. The user interface allows users to straightforwardly import a relational table through the copy and paste functions. Automatic retrieval from a relational database is also possible. Users are asked to characterize each column of the table in the tool interface (i.e., \textit{ID} column, \textit{association} column, or \textit{attribute} column). The tool requires relatively simple user inputs, as raw data can be just copied and pasted from external sources by adopting the current GUI.

\subsubsection*{Tools for Automatic Ontology Matching\label{sec:tools}}

Ontology matching tools take as input a pair of ontologies (in the \texttt{owl} format) and return as output an ontology alignment (in the \texttt{rdf} format). The \textit{Ontology Alignment Evaluation Initiative} \footnote{\url{http://oaei.ontologymatching.org/}} (OAEI) is the most known international campaign for the enhancement and evaluation of ontology matching tools. Outstanding results from the OAEI community are presented yearly in the \textit{Ontology Matching Workshop} (OM), co-located with the prestigious \textit{International Semantic Web Conference} (ISWC).
We briefly describe below the two most popular tools within the OAEI community. We adopted both tools in our experiments to highlight the possible impact of the tool adopted on the final outcome.
\begin{itemize}
    \item{\textbf{LogMap}\footnote{\url{https://github.com/ernestojimenezruiz/logmap-matcher}}~\cite{jimenez2011logmap}} is a highly scalable ontology matching tool. It performs an iterative process that starts with an initial set of \textit{equivalence} correspondences obtained from a \textit{lexical} matching process, then computes new correspondences by applying a \textit{structural} matching process to the hierarchy neighbors of (the entities composing) the initial correspondences---based on the \textit{principle of locality}. To achieve scalability, LogMap relies on lexical and structural indexes of the input ontologies. 
LogMap has taken part in the annual OAEI competition several times and has always been among the top-ranked solutions. It is considered by many to be the best ontology matching tool currently available.

\item{\textbf{AML}\footnote{\url{https://github.com/AgreementMakerLight/AML-Project}}~\cite{faria2013agreementmakerlight}} is a scalable ontology matching system characterized by a comprehensive user interface.
 
The lexical and structural information of both input ontologies is stored in internal structures (\textit{i.e.}, HashMaps and MultiMaps).
%
%
AML combines different individual matchers (i.e., the \textit{lexical} matcher, the \textit{word-based string} matcher, and alternatively the \textit{mediating} and \textit{parametric string} matchers) and keeps the highest confidence values in case of repeated correspondences.
%
%
It should be noted that the \textit{mediating} matcher uses a third external ontology as background knowledge.
To achieve scalability, AML adopts efficient indexation by making HashMap cross-search in the matching process.
AML has taken part in the annual OAEI competition on several occasions and has achieved good results. It is considered one of the best ontology matching tools currently available.
\end{itemize}

We have used the online Web Interface version\footnote{\url{http://krrwebtools.cs.ox.ac.uk/logmap/}} of LogMap. As for AML, we have used the \href{https://github.com/AgreementMakerLight/AML-Project/releases}{version $3.1$} of AML, which includes a Graphical User Interface (GUI). It should be noted that we have not used the latest AML version (version $3.2$) because it does not include instance matching (i.e., it only performs schema matching).

\subsection{Experimental Results}

Initial results are summarized in Table~\ref{tab1}. The latter shows the quality of ontology alignments resulting from each experiment using LogMap and AML. It reports the scores of the evaluation metrics, as previously described in Section~\ref{section5}.

\begin{table}[!htb]
\centering
\caption{Results Summary.\label{tab1}}
\begin{adjustbox}{scale=0.9,center}
\begin{threeparttable}
\begin{tabular}{l|lV{3.5}rrr|rrrr|r}
\toprule
\multicolumn{1}{c|}{\textbf{Exp.}}                     & \textbf{Tool}   & \multicolumn{1}{c}{\begin{tabular}[c]{@{}c@{}}\boldmath{$\mathcal{R}$}$\,$\tnote{\boldmath{$\dagger$}}\end{tabular}} & \multicolumn{1}{c}{\begin{tabular}[c]{@{}c@{}}\boldmath{$\mathcal{A}$}$\,$\tnote{\boldmath{$\ddagger$}}\end{tabular}} & \multicolumn{1}{c|}{\begin{tabular}[c]{@{}c@{}}\boldmath{$Amb$}$\,$\tnote{$\S$}\end{tabular}} & \multicolumn{1}{c}{{\cellcolor{Yellow1}} \textbf{\begin{tabular}[c]{@{}c@{}}\boldmath{$Precision$}\end{tabular}}} & \multicolumn{1}{c}{{\cellcolor{Orchid1}} \textbf{\begin{tabular}[c]{@{}c@{}}\boldmath{$Recall$}\end{tabular}}} & \multicolumn{1}{c}{{\cellcolor{CadetBlue1}} \textbf{\begin{tabular}[c]{@{}c@{}}\boldmath{$F\textnormal{-}measure$}\end{tabular}}} & \multicolumn{1}{c|}{ {\cellcolor{OliveDrab1}} \textbf{\begin{tabular}[c]{@{}c@{}}\boldmath{$Overall$}\end{tabular}}} & \multicolumn{1}{c}{{\cellcolor{Salmon1}} \textbf{\begin{tabular}[c]{@{}c@{}}\boldmath{$Ambiguity$}\end{tabular}}}\\ \midrule[1.35pt]
\multirow{2}{*}{\textbf{Exp. 1}}  & LogMap                                                      & \multirow{2}{*}{$267$}                           & $212$ & $2$ &  $0.995$                                  &  $0.79$                               & $0.881$ & $0.786$ & $0.94\%$                                              \\ 
                         & AML                                                      &                            & $270$ & $186$ &                                   $0.255$                                  & $0.258$  & $0.256$ & $-0.49$ & $68.88\%$                                                                \\ \midrule
\multirow{2}{*}{\textbf{Exp. 2}}  & LogMap                                                       & \multirow{2}{*}{$446$}                           & $200$ & $0$ &  $1.0$                                & $0.448$                                & $0.619$ & $0.448$  & $0\%$                                              \\ 
                         & AML                                                       &                            & $227$  & $15$  & $0.929$                                  & $0.473$                                  & $0.627$  & $0.437$ & $6.6\%$                                                                    \\ \midrule
\multirow{2}{*}{\textbf{Exp. 3}}  & LogMap                                                       & \multirow{2}{*}{$434$}                           & $194$ & $0$ &  $0.984$                                 & $0.44$                                & $0.6$ & $0.433$ & $0\%$                                               \\ 
                         & AML                                                       &                            & $227$ & $25$ & $0.903$                                 & $0.472$                                & $0.62$ & $0.421$  & $11\%$                                                                   \\ \midrule
\multirow{2}{*}{\textbf{Exp. 4}}  & LogMap                                                       & \multirow{2}{*}{$55$}                            & $61$ & $0$ & $0.426$                                  &  $0.472$                                & $0.448$ & $-0.163$ & $0\%$                                              \\ 
                         & AML                                                      & & $331$                             & $317$ & $0.033$ & $0.2$                                   & $0.056$                                   & $-5.61$  & $95.77\%$                                                                   \\ \midrule
\multirow{2}{*}{\textbf{Exp. 5}}  & LogMap                                                       & \multirow{2}{*}{$137$}                             & $3$ & $0$ & $1.0$                                  & $0.021$                                  & $0.042$  & $0.021$ & $0\%$                                               \\ 
                         & AML                                                       &                              & $3$ & $0$&   $1.0$                                & $0.021$                                  & $0.042$ & $0.021$  & $0\%$                                                                    \\ \midrule
\multirow{2}{*}{\textbf{Exp. 6}}  & LogMap                                                       & \multirow{2}{*}{$102$}                            & $31$ & $0$ & $1.0$                                  &  $0.3$                                 & $0.466$ & $0.3$ & $0\%$                                              \\ 
                         & AML                                                      &                             & $158$ & $129$ & $0.177$                                  & $0.274$                                  & $0.215$  & $-1.0$ & $81.64\%$                                                                   \\ \midrule
\multirow{2}{*}{\textbf{Exp. 7}}  & LogMap                                                       & \multirow{2}{*}{$521$}                           & $290$ & $0$ &  $0.996$                                 &  $0.554$                               & $0.712$ & $0.552$ & $0\%$                                              \\ 
                         & AML                                                      &                             & $3680$ & $3628$ & $0.013$                                 & $0.095$                                  & $0.023$  & $-6.871$ & $98.58\%$                                                                    \\ \midrule
\multirow{2}{*}{\textbf{Exp. 8}}  & LogMap                                                       & \multirow{2}{*}{$195$}                           & $190$ & $0$ & $1.0$                                   & $0.974$   & $0.987$                                & $0.974$ & $0\%$                                              \\ 
                         & AML                                                       &                            & $190$ & $2$ & $0.989$                                   & $0.964$                                 & $0.976$  & $0.953$  & $1\%$                                                                   \\ \midrule
\multirow{2}{*}{\textbf{Exp. 9}}                   & LogMap                                                       & \multirow{2}{*}{$196$}                           & $177$ & $0$ & $1.0$                                  & $0.9$                                 & $0.949$  & $0.9$ & $0\%$                                               \\ 
                         & AML                                                       &                            & $192$ & $2$ & $0.973$                                    &  $0.954$                                & $0.963$  & $0.928$  & $1.04\%$                                                                   \\ \midrule
\multirow{2}{*}{\textbf{Exp. 10}} & LogMap                                                       & \multirow{2}{*}{$197$}                           & $179$ & $0$ & $1.0$                                  & $0.91$                                  & $0.952$  & $0.91$ & $0\%$                                               \\ 
                         & AML                                                       &                            & $187$ & $0$ & $0.994$                                   & $0.944$                                 & $0.968$   & $0.939$  & $0\%$                                                                   \\ \bottomrule
\end{tabular}
\begin{tablenotes}[para,flushleft]
        {\small
            \item[\boldmath{$\dagger$}] $\mathcal{R}:$ Number of correspondences in the reference alignment $\mathcal{R}$ = Number of expected correspondences $= TP + TN$.
            
            \item[\boldmath{$\ddagger$}] $\mathcal{A}:$ Number of correspondences in the output alignment $\mathcal{A}$ = Number of detected correspondences $= TP + FP$.
           
            \item[\boldmath{$\S$}] $Amb:$ Number of ambiguous correspondences in the output alignment $\mathcal{A}$.
         }
    \end{tablenotes}
\end{threeparttable}
\end{adjustbox}
\end{table}

\begin{table}[p]
\centering
\caption{Results Summary (after Alignment \textit{Disambiguation})\label{tab2}}
\begin{adjustbox}{scale=0.9,center}
\begin{threeparttable}
\begin{tabular}{l|lV{3.5}rr>{\columncolor[gray]{0.83}}r|rrrr|>{\columncolor[gray]{0.83}}r}
\toprule
\multicolumn{1}{c}{\textbf{Exp.}}                     & \textbf{Tool}   & \multicolumn{1}{c}{\begin{tabular}[c]{@{}c@{}}\boldmath{$\mathcal{R}$}$\,$\tnote{\boldmath{$\dagger$}}\end{tabular}} & \multicolumn{1}{c}{\begin{tabular}[c]{@{}c@{}}\boldmath{$\mathcal{A}$}$\,$\tnote{\boldmath{$\ddagger$}}\end{tabular}} & \multicolumn{1}{c|}{\begin{tabular}[c]{@{}c@{}}\boldmath{$Amb$}$\,$\tnote{$\S$}\end{tabular}} & \multicolumn{1}{c}{{\cellcolor{Yellow1}} \textbf{\begin{tabular}[c]{@{}c@{}}\boldmath{$Precision$}\end{tabular}}} & \multicolumn{1}{c}{ {\cellcolor{Orchid1}} \textbf{\begin{tabular}[c]{@{}c@{}}\boldmath{$Recall$}\end{tabular}}} & \multicolumn{1}{c}{ {\cellcolor{CadetBlue1}} \textbf{\begin{tabular}[c]{@{}c@{}}\boldmath{$F\textnormal{-}measure$}\end{tabular}}} & \multicolumn{1}{c|}{ {\cellcolor{OliveDrab1}} \textbf{\begin{tabular}[c]{@{}c@{}}\boldmath{$Overall$}\end{tabular}}} & \multicolumn{1}{c}{ {\cellcolor{Salmon1}} \textbf{\begin{tabular}[c]{@{}c@{}}\boldmath{$Ambiguity$}\end{tabular}}}\\ \midrule[1.35pt]
\multirow{2}{*}{\textbf{Exp. 1}}  & LogMap                                                      & \multirow{2}{*}{$267$}                           & $211$ & $0$ &  $1.0$                                  & $0.79$                                 & $0.882$ & $0.79$ & $0\%$                                              \\ 
                         & AML                                                          &                            & $118$ & $0$ & $0.559$                                 & $0.247$                                 & $0.342$ & $0.052$ & $0\%$                                                                 \\ \midrule
\multirow{2}{*}{\textbf{Exp. 2}}  & LogMap                                                       & \multirow{2}{*}{$446$}                           & $200$ & $0$ & $1.0$                                 & $0.448$                                & $0.619$ & $0.448$ & $0\%$                                              \\ 
                         & AML                                                           &                            & $218$  & $0$ & $0.958$                                   & $0.468$                                 & $0.629$ & $0.448$ & $0\%$                                                                    \\ \midrule
\multirow{2}{*}{\textbf{Exp. 3}}  & LogMap                                                        & \multirow{2}{*}{$434$}                           & $194$ & $0$ & $0.984$                                  & $0.44$                                & $0.6$ & $0.433$ & $0\%$                                               \\ 
                         & AML                                                           &                            & $210$ & $0$ & $0.952$                                 & $0.46$                                & $0.621$ & $0.437$ & $0\%$                                                                   \\ \midrule
\multirow{2}{*}{\textbf{Exp. 4}}  & LogMap                                                        & \multirow{2}{*}{$55$}                            & $61$ & $0$ & $0.426$                                  & $0.472$                                 & $0.448$  & $-0.163$ & $0\%$                                              \\ 
                         & AML                                                          &                             & $26$ & $0$ & $0.346$                                   & $0.163$                                  & $0.222$ & $-0.145$ & $0\%$                                                                   \\ \midrule
\multirow{2}{*}{\textbf{Exp. 5}}  & LogMap                                                        & \multirow{2}{*}{$137$}                             & $3$ & $0$ & $1.0$                                  & $0.021$                                  & $0.042$  & $0.021$ & $0\%$                                               \\ 
                         & AML                                                           &                              & $3$ & $0$ & $1.0$                                  & $0.021$                                   & $0.042$ & $0.021$ & $0\%$                                                                    \\ \midrule
\multirow{2}{*}{\textbf{Exp. 6}}  & LogMap                                                        & \multirow{2}{*}{$102$}                            & $31$ & $0$ & $1.0$                                  & $0.3$                                  & $0.466$  & $0.3$ & $0\%$                                              \\ 
                         & AML                                                          &                             & $40$ & $0$ & $0.7$                                  & $0.274$                                 & $0.394$ & $0.156$ & $0\%$                                                                   \\ \midrule
\multirow{2}{*}{\textbf{Exp. 7}}  & LogMap                                                        & \multirow{2}{*}{$521$}                           & $290$ & $0$ & $0.996$                                  & $0.554$                                & $0.712$  & $0.552$ & $0\%$                                              \\ 
                         & AML                                                          &                             & $171$ & $0$ & $0.292$                                 & $0.095$                                 & $0.144$ & $-0.136$ & $0\%$                                                                    \\ \midrule
\multirow{2}{*}{\textbf{Exp. 8}}  & LogMap                                                        & \multirow{2}{*}{$195$}                           & $190$ & $0$  & $1.0$                                   & $0.974$ & $0.987$                                & $0.974$ & $0\%$                                              \\ 
                         & AML                                                           &                            & $189$ & $0$ & $0.994$                                   & $0.964$                                 & $0.979$ & $0.958$ & $0\%$                                                                   \\ \midrule
\multirow{2}{*}{\textbf{Exp. 9}}                   & LogMap                                                        & \multirow{2}{*}{$196$}                           & $177$ & $0$ & $1.0$                                  & $0.9$                                & $0.949$ & $0.9$ & $0\%$                                               \\ 
                         & AML                                                           &                            & $191$ & $0$ & $0.979$                                    & $0.954$                                  & $0.966$ & $0.933$ & $0\%$                                                                   \\ \midrule
\multirow{2}{*}{\textbf{Exp. 10}} & LogMap                                                        & \multirow{2}{*}{$197$}                           & $179$ & $0$ & $1.0$                                  & $0.91$                                 & $0.952$ & $0.91$ & $0\%$                                               \\ 
                         & AML                                                           &                            & $187$ & $0$ & $0.994$                                   & $0.944$                                 & $0.968$ & $0.939$ & $0\%$                                                                   \\ \bottomrule
\end{tabular}
\begin{tablenotes}[para,flushleft]
        {\small
            \item[\boldmath{$\dagger$}] $\mathcal{R}:$ expected correspondences. \item[\boldmath{$\ddagger$}] $\mathcal{A}:$ detected correspondences.
           \item[\boldmath{$\S$}] $Amb:$ ambiguous correspondences in $\mathcal{A}$.
         }
    \end{tablenotes}
\end{threeparttable}
\end{adjustbox}

\bigskip
\bigskip

\centering
\caption{Results Summary (after Alignment \textit{Trimming}).\label{tab3}}
\begin{adjustbox}{scale=0.9,center}
\begin{threeparttable}
\begin{tabular}{l|lV{3.5}>{\columncolor{LightCyan2}}cV{3.5}rrr|rrrr|r}
\toprule
\multicolumn{1}{c|}{\textbf{Exp.}}                     & \textbf{Tool}   & {\cellcolor{LightCyan2}} \textbf{Threshold} & \multicolumn{1}{c}{\begin{tabular}[c]{@{}c@{}}\boldmath{$\mathcal{R}$}$\,$\tnote{\boldmath{$\dagger$}}\end{tabular}} & \multicolumn{1}{c}{\begin{tabular}[c]{@{}c@{}}\boldmath{$\mathcal{A}$}$\,$\tnote{\boldmath{$\ddagger$}}\end{tabular}} & \multicolumn{1}{c|}{\begin{tabular}[c]{@{}c@{}}\boldmath{$Amb$}$\,$\tnote{$\S$}\end{tabular}} & \multicolumn{1}{c}{ {\cellcolor{Yellow1}} \textbf{\begin{tabular}[c]{@{}c@{}}\boldmath{$Precision$}\end{tabular}}} & \multicolumn{1}{c}{ {\cellcolor{Orchid1}} \textbf{\begin{tabular}[c]{@{}c@{}}\boldmath{$Recall$}\end{tabular}}} & \multicolumn{1}{c}{ {\cellcolor{CadetBlue1}} \textbf{\begin{tabular}[c]{@{}c@{}}\boldmath{$F\textnormal{-}measure$}\end{tabular}}} & \multicolumn{1}{c|}{ {\cellcolor{OliveDrab1}} \textbf{\begin{tabular}[c]{@{}c@{}}\boldmath{$Overall$}\end{tabular}}} & \multicolumn{1}{c}{ {\cellcolor{Salmon1}} \textbf{\begin{tabular}[c]{@{}c@{}}\boldmath{$Ambiguity$}\end{tabular}}}\\ \midrule[1.35pt]
\multirow{2}{*}{\textbf{Exp. 1}}  & LogMap & --                                                      & \multirow{2}{*}{$267$}                           & $212$ & $2$ & $0.995$                                   & $0.79$                                & $0.881$  & $0.786$ & $0.94\%$                                              \\ 
                         & AML    & $0.77$                                                      &                            & $209$ & $138$ & $0.32$                                 & $0.25$                                 & $0.281$ & $-0.28$ & $66\%$                                                                 \\ \midrule
\multirow{2}{*}{\textbf{Exp. 2}}  & LogMap & --                                                       & \multirow{2}{*}{$446$}                           & $200$ & $0$ & $1.0$                                 & $0.448$                                & $0.619$ & $0.448$ & $0\%$                                              \\ 
                         & AML    & --                                                       &                            & $227$ & $15$ & $0.929$                                  & $0.473$                                 & $0.627$ & $0.437$ & $6.6\%$                                                                    \\ \midrule
\multirow{2}{*}{\textbf{Exp. 3}}  & LogMap & --                                                       & \multirow{2}{*}{$434$}                           & $194$ & $0$                                   & $0.984$                                & $0.44$ & $0.6$ & $0.433$  & $0\%$                                            \\ 
                         & AML    & --                                                       &                            & $227$ & $25$ & $0.903$                                 & $0.472$                                & $0.62$ & $0.421$ & $11\%$                                                                   \\ \midrule
\multirow{2}{*}{\textbf{Exp. 4}}  & LogMap & --                                                       & \multirow{2}{*}{$55$}                            & $61$ & $0$ & $0.426$                                  & $0.472$                                 & $0.448$ & $-0.163$ & $0\%$                                              \\ 
                         & AML    & $0.49$                                                      &                             & $36$ & $23$ & $0.3$                                   & $0.2$                                  & $0.241$ & $-0.254$ & $63.88\%$                                                                   \\ \midrule
\multirow{2}{*}{\textbf{Exp. 5}}  & LogMap & --                                                       & \multirow{2}{*}{$137$}                             & $3$ & $0$ & $1.0$                                  & $0.021$                                  & $0.042$ & $0.021$ & $0\%$                                               \\ 
                         & AML    & --                                                       &                              & $3$ & $0$ & $1.0$                                  & $0.021$                                  & $0.042$ & $0.021$ & $0\%$                                                                    \\ \midrule
\multirow{2}{*}{\textbf{Exp. 6}}  & LogMap & --                                                       & \multirow{2}{*}{$102$}                            & $31$ & $0$ & $1.0$                                  & $0.3$                                  & $0.466$ & $0.3$ & $0\%$                                              \\ 
                         & AML    & $0.46$                                                      &                             & $36$ & $8$ &    $0.777$                               & $0.274$                                 & $0.4$ & $0.196$ & $22.22\%$                                                                   \\ \midrule
\multirow{2}{*}{\textbf{Exp. 7}}  & LogMap & --                                                       & \multirow{2}{*}{$521$}                           & $290$ & $0$ & $0.996$                                  & $0.554$                                & $0.712$ & $0.552$ & $0\%$                                              \\ 
                         & AML    & $0.51$                                                      &                             & $50$ & $0$ & $0.98$                                 & $0.094$                                 & $0.171$ & $0.092$ & $0\%$                                                                    \\ \midrule
\multirow{2}{*}{\textbf{Exp. 8}}  & LogMap & --                                                       & \multirow{2}{*}{$195$}                           & $190$ & $0$ & $1.0$                                   & $0.974$ & $0.987$                                & $0.974$ & $0\%$                                              \\ 
                         & AML    & --                                                       &                            & $190$ & $2$ & $0.989$                                   & $0.964$                                 & $0.976$ & $0.953$ & $1\%$                                                                   \\ \midrule
\multirow{2}{*}{\textbf{Exp. 9}}                   & LogMap & --                                                       & \multirow{2}{*}{$196$}                           & $177$ & $0$ & $1.0$                                  & $0.9$                                & $0.949$ & $0.9$ & $0\%$                                               \\ 
                         & AML    & --                                                       &                            & $192$ & $2$ & $0.973$                                    & $0.954$                                 & $0.963$ & $0.928$ & $1\%$                                                                   \\ \midrule
\multirow{2}{*}{\textbf{Exp. 10}} & LogMap & --                                                       & \multirow{2}{*}{$197$}                           & $179$ & $0$ & $1.0$                                  & $0.91$                                 & $0.952$ & $0.91$ & $0\%$                                               \\ 
                         & AML    & --                                                       &                            & $187$ & $0$ & $0.994$                                   & $0.944$                                 & $0.968$ & $0.939$ & $0\%$                                                                   \\ \bottomrule
\end{tabular}
\begin{tablenotes}[para,flushleft]
        {\small
            \item[\boldmath{$\dagger$}] $\mathcal{R}:$ expected correspondences. \item[\boldmath{$\ddagger$}] $\mathcal{A}:$ detected correspondences.
           \item[\boldmath{$\S$}] $Amb:$ ambiguous correspondences in $\mathcal{A}$.
         }
    \end{tablenotes}
\end{threeparttable}
\end{adjustbox}
\end{table}

By looking at the number of correspondences in the \textit{reference} alignment $\mathcal{R}$ and in the resulting alignment $\mathcal{A}$ and comparing them, we can initially approximate the quality of the resulting alignments at first glance. To do so, we propose a further intuitive metric called $\Delta$, which expresses the difference between the number of expected correspondences and the number of detected correspondences ($\Delta = \mathcal{R}-\mathcal{A}$). By definition, $\Delta=0$ is not synonymous with a correct matching because of potential false positives. Yet, such a metric is a valuable indicator to have a simple preliminary assessment by identifying potential under-matching ($\Delta>0$) and over-matching ($\Delta<0$).

In Table~\ref{tab1}, advanced metrics (\textit{precision}, \textit{recall}, \textit{F-measure}, and \textit{overall}) show similar results for the two adopted tools in most experiments. More concretely, for the \textit{precision} metric, both tools provide a significantly different result in Experiments~1, 4, 6, and~7, and minor differences in the remaining experiments. The \textit{recall} metric presents remarkable differences in Experiments~1, 4, and~7. \textit{F-measure} values are quite different in Experiments~1, 4, and~7, while a pointless difference also shows up in Experiment~6. Finally, \textit{overall} returns strongly different values in Experiments~1 and~7 and a more limited divergence in Experiments~4 and~6. In terms of performance, the ontology matching process averagely results in high values for \textit{precision} with some evident exceptions (Experiment~4 especially), while \textit{recall}, \textit{F-measure} and \textit{overall} values associated with the different experiments present important differences.

It is also worth mentioning that the three last experiments (in cross-domain matching) display a higher performance in terms of both \textit{precision} and \textit{recall}. Intuitively, integrating ontologies from different domains is relatively easier than integrating ontologies within the same domain because the number of potentially ambiguous correspondences is supposed to be averagely lower (as shown in Table~\ref{tab1}). Additionally, ontologies belonging to the same domain or contiguous domains are often characterized by fine-grained terminology and heterogeneity.
Therefore, ontology matching tools generate more false correspondences when matching ontologies from related or close domains.

All results are summarized in Tables~\ref{tab1}, \ref{tab2}, \ref{tab3}, and~\ref{tab4}. In Table~\ref{tab1}, we evaluate the initially resulting alignments (i.e., without performing any alignment \textit{disambiguation} or \textit{trimming} processes to these alignments). In Table~\ref{tab2}, we evaluate the resulting disambiguated alignments (i.e., we disambiguate the original resulting alignments by transforming them from \texttt{N-to-N} alignments to \texttt{1-to-1} alignments, then we evaluate them). In Table~\ref{tab3}, we evaluate the resulting trimmed alignments (i.e., we trim the original resulting alignments by choosing the optimal threshold for each case and performing a confidence cut, then we evaluate them). In Table~\ref{tab4}, we evaluate the resulting trimmed and disambiguated alignments (i.e., we perform a confidence cut to the original resulting alignments—after choosing the optimal threshold for each alignment—then we transform them into \texttt{1-to-1} alignments, and we finally evaluate them.)

In Tables~\ref{tab2} and ~\ref{tab4}, all output alignments no longer contain any sets of ambiguous correspondences. Therefore, the scores of the \textit{ambiguity degree} metric are null (\textit{See} the columns in gray). In Tables~\ref{tab3} and ~\ref{tab4}, we report the trimming \textit{threshold} values considered in the different experiments. Recall that alignment \textit{trimming} applies a confidence $\alpha$-cut to the produced alignment, where $\alpha \in [0, 1]$ is the confidence threshold value. Threshold tuning is performed in order to choose the optimal threshold value for each case.
Indeed, in some experiments in Table~\ref{tab1}, AML alignments contain many false correspondences with very low confidence values. These low-confidence correspondences make the \textit{precision} of AML alignments decrease and therefore worsen the \textit{F-measure}. To improve AML performance in these experiments, we have used a trimming threshold that generates the best \textit{F-measure} results: We have made many manual trials to finally find the optimal threshold value that maximizes the \textit{F-measure} of these AML alignments. We did not introduce any thresholds to LogMap experiments since the LogMap results are optimal by default.

Tables~\ref{tab11}, \ref{tab22}, \ref{tab33}, \ref{tab44}, and \ref{tab55} compare the results of Tables~\ref{tab1}, \ref{tab2}, \ref{tab3}, and \ref{tab4} by taking one evaluation metric at a time. They separately show the four tables' \textit{precision}, \textit{recall}, \textit{F-measure}, \textit{overall}, and \textit{ambiguity degree} scores, respectively. In all these tables, the highest score is highlighted in bold for each experiment (and for each tool). The colored (up or down) arrows show an increase or a decrease compared with the scores of the initial results (from Table~\ref{tab1}). It should be noted that sometimes the \textit{trimming} process or the \textit{disambiguation} process has no effect on the results of Table~\ref{tab4}. In other words, some results after alignment \textit{disambiguation} (shown in Table~\ref{tab2}) do not change even if we add a \textit{trimming} process (as shown in Table~\ref{tab4}); and some results after alignment \textit{trimming} (shown in Table~\ref{tab3}) do not change even if we add a \textit{disambiguation} process (as shown in Table~\ref{tab4}).
That is why the arrows in the last column of Tables~\ref{tab11}, \ref{tab22}, \ref{tab33}, \ref{tab44}, and \ref{tab55} can have different colors according to the scores of Table~\ref{tab4}.

\begin{table}[p]
\centering
\caption{Results Summary (after Alignment \textit{Trimming} and \textit{Disambiguation}).\label{tab4}}
\begin{adjustbox}{scale=0.9,center}
\begin{threeparttable}
\begin{tabular}{l|lV{3.5}>{\columncolor{LightCyan2}}cV{3.5}rr>{\columncolor[gray]{0.83}}r|rrrr|>{\columncolor[gray]{0.83}}r}
\toprule
\multicolumn{1}{c|}{\textbf{Exp.}}                     & \textbf{Tool}   & {\cellcolor{LightCyan2}}\textbf{Threshold} & \multicolumn{1}{c}{\begin{tabular}[c]{@{}c@{}}\boldmath{$\mathcal{R}$}$\,$\tnote{\boldmath{$\dagger$}}\end{tabular}} & \multicolumn{1}{c}{\begin{tabular}[c]{@{}c@{}}\boldmath{$\mathcal{A}$}$\,$\tnote{\boldmath{$\ddagger$}}\end{tabular}} & \multicolumn{1}{c|}{\begin{tabular}[c]{@{}c@{}}\boldmath{$Amb$}$\,$\tnote{$\S$}\end{tabular}} & \multicolumn{1}{c}{ {\cellcolor{Yellow1}} \textbf{\begin{tabular}[c]{@{}c@{}}\boldmath{$Precision$}\end{tabular}}} & \multicolumn{1}{c}{ {\cellcolor{Orchid1}} \textbf{\begin{tabular}[c]{@{}c@{}}\boldmath{$Recall$}\end{tabular}}} & \multicolumn{1}{c}{ {\cellcolor{CadetBlue1}} \textbf{\begin{tabular}[c]{@{}c@{}}\boldmath{$F\textnormal{-}measure$}\end{tabular}}} & \multicolumn{1}{c|}{ {\cellcolor{OliveDrab1}} \textbf{\begin{tabular}[c]{@{}c@{}}\boldmath{$Overall$}\end{tabular}}} & \multicolumn{1}{c}{ {\cellcolor{Salmon1}} \textbf{\begin{tabular}[c]{@{}c@{}}\boldmath{$Ambiguity$}\end{tabular}}}\\ \midrule[1.35pt]
\multirow{2}{*}{\textbf{Exp. 1}}  & LogMap & --                                                      & \multirow{2}{*}{$267$}                           & $211$ & $0$ & $1.0$                                   & $0.79$                                & $0.882$ & $0.79$ & $0\%$                                              \\ 
                         & AML    & $0.77$                                                      &                            & $90$ & $0$ & $0.722$                                 & $0.243$                                 & $0.364$ & $0.149$ & $0\%$                                                                 \\ \midrule
\multirow{2}{*}{\textbf{Exp. 2}}  & LogMap & --                                                       & \multirow{2}{*}{$446$}                           & $200$ & $0$ & $1.0$                                 & $0.448$                                & $0.619$ & $0.448$ & $0\%$                                              \\ 
                         & AML    & --                                                       &                            & $218$ & $0$ & $0.958$                                  &    $0.468$                              & $0.629$ & $0.448$ & $0\%$                                                                    \\ \midrule
\multirow{2}{*}{\textbf{Exp. 3}}  & LogMap & --                                                       & \multirow{2}{*}{$434$}                           & $194$ & $0$ & $0.984$                                  & $0.44$                                & $0.6$ & $0.433$ & $0\%$                                               \\ 
                         & AML    & --                                                       &                            & $210$ & $0$ & $0.952$                                 & $0.46$                                & $0.621$ & $0.437$ & $0\%$                                                                   \\ \midrule
\multirow{2}{*}{\textbf{Exp. 4}}  & LogMap & --                                                       & \multirow{2}{*}{$55$}                            & $61$ & $0$ & $0.426$                                  & $0.472$                                 & $0.448$ & $-0.163$ & $0\%$                                              \\ 
                         & AML    & $0.49$                                                      &                             & $20$ & $0$ & $0.45$                                   & $0.163$                                  & $0.24$ & $-0.03$ & $0\%$                                                                   \\ \midrule
\multirow{2}{*}{\textbf{Exp. 5}}  & LogMap & --                                                       & \multirow{2}{*}{$137$}                             & $3$ & $0$ & $1.0$                                  & $0.021$                                  & $0.042$ & $0.021$ & $0\%$                                               \\ 
                         & AML    & --                                                       &                              & $3$ & $0$ & $1.0$                                  & $0.021$                                  & $0.042$ & $0.021$ & $0\%$                                                                    \\ \midrule
\multirow{2}{*}{\textbf{Exp. 6}}  & LogMap & --                                                       & \multirow{2}{*}{$102$}                            & $31$ & $0$ & $1.0$                                  & $0.3$                                  & $0.466$ & $0.3$ & $0\%$                                              \\ 
                         & AML    & $0.46$                                                      &                             & $29$ & $0$ & $0.965$                                  & $0.274$                                 & $0.427$ & $0.264$ & $0\%$                                                                   \\ \midrule
\multirow{2}{*}{\textbf{Exp. 7}}  & LogMap & --                                                       & \multirow{2}{*}{$521$}                           & $290$ & $0$ & $0.996$                                  & $0.554$                                & $0.712$  & $0.552$ & $0\%$                                              \\ 
                         & AML    & $0.51$                                                      &                             & $50$ & $0$ & $0.98$                                 & $0.094$                                 & $0.171$ & $0.092$ & $0\%$                                                                    \\ \midrule
\multirow{2}{*}{\textbf{Exp. 8}}  & LogMap & --                                                       & \multirow{2}{*}{$195$}                           & $190$ & $0$ & $1.0$                                   & $0.974$ & $0.987$                                & $0.974$ & $0\%$                                              \\ 
                         & AML    & --                                                       &                            & $189$ & $0$ & $0.994$                                   & $0.964$                                 & $0.979$ & $0.958$ & $0\%$                                                                   \\ \midrule
\multirow{2}{*}{\textbf{Exp. 9}}                   & LogMap & --                                                       & \multirow{2}{*}{$196$}                           & $177$ & $0$ & $1.0$                                  & $0.9$                                & $0.949$ & $0.9$ & $0\%$                                               \\ 
                         & AML    & --                                                       &                            & $191$ & $0$ & $0.979$                                    & $0.954$                                 & $0.966$ & $0.933$ & $0\%$                                                                   \\ \midrule
\multirow{2}{*}{\textbf{Exp. 10}} & LogMap & --                                                       & \multirow{2}{*}{$197$}                           & $179$ & $0$ & $1.0$                                  & $0.91$                                 & $0.952$ & $0.91$ & $0\%$                                               \\ 
                         & AML    & --                                                       &                            & $187$ & $0$ & $0.994$                                   & $0.944$                                 & $0.968$ & $0.939$ & $0\%$                                                                   \\ \bottomrule
\end{tabular}
\begin{tablenotes}[para,flushleft]
        {\small
            \item[\boldmath{$\dagger$}] $\mathcal{R}:$ expected correspondences. \item[\boldmath{$\ddagger$}] $\mathcal{A}:$ detected correspondences.
           \item[\boldmath{$\S$}] $Amb:$ ambiguous correspondences in $\mathcal{A}$.
         }
    \end{tablenotes}
\end{threeparttable}
\end{adjustbox}

\bigskip
\bigskip

\centering
\caption{Comparison between \textit{Precision} Results in Different Tables.\label{tab11}}
\begin{adjustbox}{scale=0.9,center}
\begin{threeparttable}
\begin{tabular}{l|lV{3.5}>{\raggedleft}p{0.13\linewidth}||r|r|r}
\toprule
\multicolumn{1}{c|}{\textbf{Exp.}}                     & \textbf{Tool}   &  \multicolumn{1}{c||}{ {\cellcolor{Yellow1}}\textbf{\begin{tabular}[c]{@{}c@{}}\boldmath{$Precision$}$\,$\tnote{$1$}\end{tabular}}} & \multicolumn{1}{c|}{ {\cellcolor{Yellow1}}\textbf{\begin{tabular}[c]{@{}c@{}}\boldmath{$Precision$}$\,$\tnote{$2$}$\,\,\,$\textcolor{red}{$\downarrow \uparrow$}\end{tabular}}} & \multicolumn{1}{c|}{  {\cellcolor{Yellow1}} \textbf{\begin{tabular}[c]{@{}c@{}}\boldmath{$Precision$}$\,$\tnote{$3$}$\,\,\,$\textcolor{SeaGreen3}{$\downarrow \uparrow$}\end{tabular}}} & \multicolumn{1}{c}{ {\cellcolor{Yellow1}} \textbf{\begin{tabular}[c]{@{}c@{}}\boldmath{$Precision$}$\,$\tnote{$4$}$\,\,\,$\textcolor{blue}{$\downarrow \uparrow$}\end{tabular}}} \\
\midrule[1.35pt]
\multirow{2}{*}{\textbf{Exp. 1}}  & LogMap                                                   &  $0.995$                             &  \textcolor{red}{\bm{$\uparrow$}}  \hfill\hfill\hfill\hfill\hfill\hfill\hfill\hfill\hfill\hfill\hfill\hfill\hfill\hfill\hfill\hfill\hfill\hfill\hfill\hfill\hfill\hfill\hfill\hfill\hfill\hfill\hfill\hfill\hfill\hfill\hfill $\textbf{1.0}$ & $0.995$ & \textcolor{red}{\bm{$\uparrow$}} \hfill\hfill\hfill\hfill\hfill\hfill\hfill\hfill\hfill\hfill\hfill\hfill\hfill\hfill\hfill\hfill\hfill\hfill\hfill\hfill\hfill\hfill\hfill\hfill\hfill\hfill\hfill\hfill\hfill\hfill $\textbf{1.0}$                                              \\ 
                         & AML & $0.255$                        & \textcolor{red}{\bm{$\uparrow$}} \hfill\hfill\hfill\hfill\hfill\hfill\hfill\hfill\hfill\hfill\hfill\hfill\hfill\hfill\hfill\hfill\hfill\hfill\hfill\hfill\hfill\hfill\hfill\hfill\hfill\hfill\hfill\hfill\hfill\hfill $0.559$ & \textcolor{SeaGreen3}{\bm{$\uparrow$}} \hfill\hfill\hfill\hfill\hfill\hfill\hfill\hfill\hfill\hfill\hfill\hfill\hfill\hfill\hfill\hfill\hfill\hfill\hfill\hfill\hfill\hfill\hfill\hfill\hfill\hfill\hfill\hfill\hfill\hfill\hfill $0.32$ & \textcolor{blue}{\bm{$\uparrow$}} \hfill\hfill\hfill\hfill\hfill\hfill\hfill\hfill\hfill\hfill\hfill\hfill\hfill\hfill\hfill\hfill\hfill\hfill\hfill\hfill\hfill\hfill\hfill\hfill\hfill\hfill\hfill\hfill\hfill\hfill\hfill $\textbf{0.722}$                                                              \\ \midrule
\multirow{2}{*}{\textbf{Exp. 2}}  & LogMap                                                       & $1.0$ & $1.0$ & $1.0$ & $1.0$                                              \\ 
                         & AML                & $0.929$                           & \textcolor{red}{\bm{$\uparrow$}} \hfill\hfill\hfill\hfill\hfill\hfill\hfill\hfill\hfill\hfill\hfill\hfill\hfill\hfill\hfill\hfill\hfill\hfill\hfill\hfill\hfill\hfill\hfill\hfill\hfill\hfill\hfill\hfill\hfill\hfill\hfill $\textbf{0.958}$ & $0.929$ & \textcolor{red}{\bm{$\uparrow$}} \hfill\hfill\hfill\hfill\hfill\hfill\hfill\hfill\hfill\hfill\hfill\hfill\hfill\hfill\hfill\hfill\hfill\hfill\hfill\hfill\hfill\hfill\hfill\hfill\hfill\hfill\hfill\hfill\hfill\hfill\hfill $\textbf{0.958}$                                                                   \\ \midrule
\multirow{2}{*}{\textbf{Exp. 3}}  & LogMap                                                       & $0.984$ & $0.984$ & $0.984$ & $0.984$                                               \\ 
                         & AML                                                       & $0.903$                           & \textcolor{red}{\bm{$\uparrow$}}  \hfill\hfill\hfill\hfill\hfill\hfill\hfill\hfill\hfill\hfill\hfill\hfill\hfill\hfill\hfill\hfill\hfill\hfill\hfill\hfill\hfill\hfill\hfill\hfill\hfill\hfill\hfill\hfill\hfill\hfill\hfill $\textbf{0.952}$ & $0.903$ & \textcolor{red}{\bm{$\uparrow$}}  \hfill\hfill\hfill\hfill\hfill\hfill\hfill\hfill\hfill\hfill\hfill\hfill\hfill\hfill\hfill\hfill\hfill\hfill\hfill\hfill\hfill\hfill\hfill\hfill\hfill\hfill\hfill\hfill\hfill\hfill\hfill $\textbf{0.952}$                                                                  \\ \midrule
\multirow{2}{*}{\textbf{Exp. 4}}  & LogMap                                                       & $0.426$ & $0.426$ & $0.426$ & $0.426$                                             \\ 
                         & AML                                                      & $0.033$ & \textcolor{red}{\bm{$\uparrow$}} \hfill\hfill\hfill\hfill\hfill\hfill\hfill\hfill\hfill\hfill\hfill\hfill\hfill\hfill\hfill\hfill\hfill\hfill\hfill\hfill\hfill\hfill\hfill\hfill\hfill\hfill\hfill\hfill\hfill\hfill\hfill $0.346$ & \textcolor{SeaGreen3}{\bm{$\uparrow$}} \hfill\hfill\hfill\hfill\hfill\hfill\hfill\hfill\hfill\hfill\hfill\hfill\hfill\hfill\hfill\hfill\hfill\hfill\hfill\hfill\hfill\hfill\hfill\hfill\hfill\hfill\hfill\hfill\hfill\hfill\hfill $0.3$ & \textcolor{blue}{\bm{$\uparrow$}} \hfill\hfill\hfill\hfill\hfill\hfill\hfill\hfill\hfill\hfill\hfill\hfill\hfill\hfill\hfill\hfill\hfill\hfill\hfill\hfill\hfill\hfill\hfill\hfill\hfill\hfill\hfill\hfill\hfill\hfill\hfill $\textbf{0.45}$                                                                   \\ \midrule
\multirow{2}{*}{\textbf{Exp. 5}}  & LogMap                                                       & $1.0$ & $1.0$ & $1.0$ & $1.0$                                              \\ 
                         & AML                                    & $1.0$ & $1.0$ & $1.0$ & $1.0$                                                                   \\ \midrule
\multirow{2}{*}{\textbf{Exp. 6}}  & LogMap                                                       & $1.0$ & $1.0$ & $1.0$ & $1.0$                                             \\ 
                         & AML                                                      & $0.177$ & \textcolor{red}{\bm{$\uparrow$}} \hfill\hfill\hfill\hfill\hfill\hfill\hfill\hfill\hfill\hfill\hfill\hfill\hfill\hfill\hfill\hfill\hfill\hfill\hfill\hfill\hfill\hfill\hfill\hfill\hfill\hfill\hfill\hfill\hfill\hfill\hfill $0.7$ & \textcolor{SeaGreen3}{\bm{$\uparrow$}} \hfill\hfill\hfill\hfill\hfill\hfill\hfill\hfill\hfill\hfill\hfill\hfill\hfill\hfill\hfill\hfill\hfill\hfill\hfill\hfill\hfill\hfill\hfill\hfill\hfill\hfill\hfill\hfill\hfill\hfill\hfill $0.777$ & \textcolor{blue}{\bm{$\uparrow$}} \hfill\hfill\hfill\hfill\hfill\hfill\hfill\hfill\hfill\hfill\hfill\hfill\hfill\hfill\hfill\hfill\hfill\hfill\hfill\hfill\hfill\hfill\hfill\hfill\hfill\hfill\hfill\hfill\hfill\hfill\hfill $\textbf{0.965}$                                                                   \\ \midrule
\multirow{2}{*}{\textbf{Exp. 7}}  & LogMap                                                       & $0.996$ & $0.996$ & $0.996$ & $0.996$                                             \\ 
                         & AML                                                      & $0.013$ & \textcolor{red}{\bm{$\uparrow$}} \hfill\hfill\hfill\hfill\hfill\hfill\hfill\hfill\hfill\hfill\hfill\hfill\hfill\hfill\hfill\hfill\hfill\hfill\hfill\hfill\hfill\hfill\hfill\hfill\hfill\hfill\hfill\hfill\hfill\hfill\hfill $0.292$ & \textcolor{SeaGreen3}{\bm{$\uparrow$}} \hfill\hfill\hfill\hfill\hfill\hfill\hfill\hfill\hfill\hfill\hfill\hfill\hfill\hfill\hfill\hfill\hfill\hfill\hfill\hfill\hfill\hfill\hfill\hfill\hfill\hfill\hfill\hfill\hfill\hfill\hfill $\textbf{0.98}$ & \textcolor{SeaGreen3}{\bm{$\uparrow$}} \hfill\hfill\hfill\hfill\hfill\hfill\hfill\hfill\hfill\hfill\hfill\hfill\hfill\hfill\hfill\hfill\hfill\hfill\hfill\hfill\hfill\hfill\hfill\hfill\hfill\hfill\hfill\hfill\hfill\hfill\hfill $\textbf{0.98}$                                                                   \\ \midrule
\multirow{2}{*}{\textbf{Exp. 8}}  & LogMap                                                       & $1.0$ & $1.0$ & $1.0$ & $1.0$                                             \\ 
                         & AML                                                       & $0.989$ & \textcolor{red}{\bm{$\uparrow$}} \hfill\hfill\hfill\hfill\hfill\hfill\hfill\hfill\hfill\hfill\hfill\hfill\hfill\hfill\hfill\hfill\hfill\hfill\hfill\hfill\hfill\hfill\hfill\hfill\hfill\hfill\hfill\hfill\hfill\hfill\hfill $\textbf{0.994}$ & $0.989$ & \textcolor{red}{\bm{$\uparrow$}} \hfill\hfill\hfill\hfill\hfill\hfill\hfill\hfill\hfill\hfill\hfill\hfill\hfill\hfill\hfill\hfill\hfill\hfill\hfill\hfill\hfill\hfill\hfill\hfill\hfill\hfill\hfill\hfill\hfill\hfill\hfill $\textbf{0.994}$                                                                   \\ \midrule
\multirow{2}{*}{\textbf{Exp. 9}}                   & LogMap                                                       & $1.0$ & $1.0$ & $1.0$ & $1.0$                                              \\ 
                         & AML                                                       & $0.973$                          & \textcolor{red}{\bm{$\uparrow$}} \hfill\hfill\hfill\hfill\hfill\hfill\hfill\hfill\hfill\hfill\hfill\hfill\hfill\hfill\hfill\hfill\hfill\hfill\hfill\hfill\hfill\hfill\hfill\hfill\hfill\hfill\hfill\hfill\hfill\hfill\hfill $\textbf{0.979}$ & $0.973$ & \textcolor{red}{\bm{$\uparrow$}} \hfill\hfill\hfill\hfill\hfill\hfill\hfill\hfill\hfill\hfill\hfill\hfill\hfill\hfill\hfill\hfill\hfill\hfill\hfill\hfill\hfill\hfill\hfill\hfill\hfill\hfill\hfill\hfill\hfill\hfill\hfill $\textbf{0.979}$                                                                  \\ \midrule
\multirow{2}{*}{\textbf{Exp. 10}} & LogMap                                                       & $1.0$ & $1.0$ & $1.0$ & $1.0$                                               \\ 
                         & AML                                                       & $0.994$ & $0.994$ & $0.994$ & $0.994$                                                                  \\ \bottomrule
\end{tabular}
\begin{tablenotes}[para,flushleft]
        {\small
            \item[\boldmath{$1$}] in Table~\ref{tab1}. \item[\boldmath{$2$}] in Table~\ref{tab2}. \item[\boldmath{$3$}] in Table~\ref{tab3}. \item[\boldmath{$4$}] in Table~\ref{tab4}.
         }
    \end{tablenotes}
\end{threeparttable}
\end{adjustbox}
\end{table}


\begin{table}[p]
\centering
\caption{Comparison between \textit{Recall} Results in Different Tables.\label{tab22}}
\begin{adjustbox}{scale=0.9,center}
\begin{threeparttable}
\begin{tabular}{l|lV{3.5}>{\raggedleft}p{0.107\linewidth}||r|r|r}
\toprule
\multicolumn{1}{c|}{\textbf{Exp.}}                     & \textbf{Tool}   & \multicolumn{1}{c||}{ {\cellcolor{Orchid1}} \textbf{\begin{tabular}[c]{@{}c@{}}\boldmath{$Recall$}$\,$\tnote{$1$}\end{tabular}}} & \multicolumn{1}{c|}{ {\cellcolor{Orchid1}} \textbf{\begin{tabular}[c]{@{}c@{}}\boldmath{$Recall$}$\,$\tnote{$2$}$\,\,\,$\textcolor{red}{$\downarrow \uparrow$}\end{tabular}}} & \multicolumn{1}{c|}{ {\cellcolor{Orchid1}} \textbf{\begin{tabular}[c]{@{}c@{}}\boldmath{$Recall$}$\,$\tnote{$3$}$\,\,\,$\textcolor{SeaGreen3}{$\downarrow \uparrow$}\end{tabular}}} & \multicolumn{1}{c}{ {\cellcolor{Orchid1}} \textbf{\begin{tabular}[c]{@{}c@{}}\boldmath{$Recall$}$\,$\tnote{$4$}$\,\,\,$\textcolor{blue}{$\downarrow \uparrow$}\end{tabular}}} \\ \midrule[1.35pt]
\multirow{2}{*}{\textbf{Exp. 1}}  & LogMap                                                   & $0.79$                             & $0.79$ & $0.79$ & $0.79$                                              \\ 
                         & AML & $\textbf{0.258}$                        & \textcolor{red}{\bm{$\downarrow$}}  \hfill\hfill\hfill\hfill\hfill\hfill\hfill\hfill\hfill\hfill\hfill\hfill\hfill\hfill\hfill\hfill\hfill\hfill\hfill\hfill\hfill\hfill\hfill\hfill\hfill\hfill\hfill\hfill\hfill\hfill\hfill $0.247$ & \textcolor{SeaGreen3}{\bm{$\downarrow$}}  \hfill\hfill\hfill\hfill\hfill\hfill\hfill\hfill\hfill\hfill\hfill\hfill\hfill\hfill\hfill\hfill $0.25$ & \textcolor{blue}{\bm{$\downarrow$}}  \hfill\hfill\hfill\hfill\hfill\hfill\hfill\hfill\hfill\hfill\hfill\hfill\hfill\hfill\hfill\hfill\hfill\hfill\hfill\hfill\hfill\hfill\hfill\hfill\hfill\hfill\hfill\hfill\hfill\hfill\hfill $0.243$                                                              \\ \midrule
\multirow{2}{*}{\textbf{Exp. 2}}  & LogMap                                                       & $0.448$ & $0.448$ & $0.448$ & $0.448$                                              \\ 
                         & AML                & $\textbf{0.473}$                           & \textcolor{red}{\bm{$\downarrow$}}  \hfill\hfill\hfill\hfill\hfill\hfill\hfill\hfill\hfill\hfill\hfill\hfill\hfill\hfill\hfill\hfill\hfill\hfill\hfill\hfill\hfill\hfill\hfill\hfill\hfill\hfill\hfill\hfill\hfill\hfill\hfill $0.468$ & $\textbf{0.473}$ & \textcolor{red}{\bm{$\downarrow$}}  \hfill\hfill\hfill\hfill\hfill\hfill\hfill\hfill\hfill\hfill\hfill\hfill\hfill\hfill\hfill\hfill\hfill\hfill\hfill\hfill\hfill\hfill\hfill\hfill\hfill\hfill\hfill\hfill\hfill\hfill\hfill $0.468$                                                                   \\ \midrule
\multirow{2}{*}{\textbf{Exp. 3}}  & LogMap                                                       & $0.44$ & $0.44$ & $0.44$ & $0.44$                                               \\ 
                         & AML                                                       & $\textbf{0.472}$                           & \textcolor{red}{\bm{$\downarrow$}}  \hfill\hfill\hfill\hfill\hfill\hfill\hfill\hfill\hfill\hfill\hfill\hfill\hfill\hfill\hfill\hfill\hfill\hfill\hfill\hfill\hfill\hfill\hfill\hfill\hfill\hfill\hfill\hfill\hfill\hfill\hfill $0.46$ & $\textbf{0.472}$ & \textcolor{red}{\bm{$\downarrow$}}  \hfill\hfill\hfill\hfill\hfill\hfill\hfill\hfill\hfill\hfill\hfill\hfill\hfill\hfill\hfill\hfill\hfill\hfill\hfill\hfill\hfill\hfill\hfill\hfill\hfill\hfill\hfill\hfill\hfill\hfill\hfill $0.46$                                                                  \\ \midrule
\multirow{2}{*}{\textbf{Exp. 4}}  & LogMap                                                       & $0.472$ & $0.472$ & $0.472$ & $0.472$                                             \\ 
                         & AML                                                      & $\textbf{0.2}$ & \textcolor{red}{\bm{$\downarrow$}}  \hfill\hfill\hfill\hfill\hfill\hfill\hfill\hfill\hfill\hfill\hfill\hfill\hfill\hfill\hfill\hfill\hfill\hfill\hfill\hfill\hfill\hfill\hfill\hfill\hfill\hfill\hfill\hfill\hfill\hfill\hfill $0.163$ & $\textbf{0.2}$ & \textcolor{red}{\bm{$\downarrow$}}  \hfill\hfill\hfill\hfill\hfill\hfill\hfill\hfill\hfill\hfill\hfill\hfill\hfill\hfill\hfill\hfill\hfill\hfill\hfill\hfill\hfill\hfill\hfill\hfill\hfill\hfill\hfill\hfill\hfill\hfill\hfill $0.163$                                                                   \\ \midrule
\multirow{2}{*}{\textbf{Exp. 5}}  & LogMap                                                       & $0.021$ & $0.021$ & $0.021$ & $0.021$                                              \\ 
                         & AML                                    & $0.021$ & $0.021$ & $0.021$ & $0.021$                                                                   \\ \midrule
\multirow{2}{*}{\textbf{Exp. 6}}  & LogMap                                                       & $0.3$ & $0.3$ & $0.3$ & $0.3$                                             \\ 
                         & AML                                                      & $0.274$ & $0.274$ & $0.274$ & $0.274$                                                                   \\ \midrule
\multirow{2}{*}{\textbf{Exp. 7}}  & LogMap                                                       & $0.554$ & $0.554$ & $0.554$ & $0.554$                                             \\ 
                         & AML                                                      & $\textbf{0.095}$ & $\textbf{0.095}$ & \textcolor{SeaGreen3}{\bm{$\downarrow$}}  \hfill\hfill\hfill\hfill\hfill\hfill\hfill\hfill\hfill\hfill\hfill\hfill\hfill\hfill\hfill\hfill\hfill\hfill\hfill\hfill\hfill\hfill\hfill\hfill\hfill\hfill\hfill\hfill\hfill\hfill\hfill $0.094$ & \textcolor{SeaGreen3}{\bm{$\downarrow$}}  \hfill\hfill\hfill\hfill\hfill\hfill\hfill\hfill\hfill\hfill\hfill\hfill\hfill\hfill\hfill\hfill\hfill\hfill\hfill\hfill\hfill\hfill\hfill\hfill\hfill\hfill\hfill\hfill\hfill\hfill\hfill $0.094$                                                                   \\ \midrule
\multirow{2}{*}{\textbf{Exp. 8}}  & LogMap                                                       & $0.974$ & $0.974$ & $0.974$ & $0.974$                                             \\ 
                         & AML                                                       & $0.964$ & $0.964$ & $0.964$ & $0.964$                                                                   \\ \midrule
\multirow{2}{*}{\textbf{Exp. 9}}                   & LogMap                                                       & $0.9$ & $0.9$ & $0.9$ & $0.9$                                              \\ 
                         & AML                                                       & $0.954$                          & $0.954$ & $0.954$ & $0.954$                                                                  \\ \midrule
\multirow{2}{*}{\textbf{Exp. 10}} & LogMap                                                       & $0.91$ & $0.91$ & $0.91$ & $0.91$                                               \\ 
                         & AML                                                       & $0.944$ & $0.944$ & $0.944$ & $0.944$                                                                  \\ \bottomrule
\end{tabular}
\begin{tablenotes}[para,flushleft]
        {\small
            \item[\boldmath{$1$}] in Table~\ref{tab1}. \item[\boldmath{$2$}] in Table~\ref{tab2}. \item[\boldmath{$3$}] in Table~\ref{tab3}. \item[\boldmath{$4$}] in Table~\ref{tab4}.
         }
    \end{tablenotes}
\end{threeparttable}
\end{adjustbox}

\bigskip
\bigskip

\centering
\caption{Comparison between \textit{F-measure} Results in Different Tables.\label{tab33}}
\begin{adjustbox}{scale=0.9,center}
\begin{threeparttable}
\begin{tabular}{l|lV{3.5}>{\raggedleft}p{0.14\linewidth}||r|r|r}
\toprule
\multicolumn{1}{c|}{\textbf{Exp.}}                     & \textbf{Tool}   & \multicolumn{1}{c||}{ {\cellcolor{CadetBlue1}} \textbf{\begin{tabular}[c]{@{}c@{}}\boldmath{$F\textnormal{-}measure$}$\,$\tnote{$1$}\end{tabular}}} & \multicolumn{1}{c|}{ {\cellcolor{CadetBlue1}} \textbf{\begin{tabular}[c]{@{}c@{}}\boldmath{$F\textnormal{-}measure$}$\,$\tnote{$2$}$\,\,\,$\textcolor{red}{$\downarrow \uparrow$}\end{tabular}}} & \multicolumn{1}{c|}{ {\cellcolor{CadetBlue1}} \textbf{\begin{tabular}[c]{@{}c@{}}\boldmath{$F\textnormal{-}measure$}$\,$\tnote{$3$}$\,\,\,$\textcolor{SeaGreen3}{$\downarrow \uparrow$}\end{tabular}}} & \multicolumn{1}{c}{ {\cellcolor{CadetBlue1}} \textbf{\begin{tabular}[c]{@{}c@{}}\boldmath{$F\textnormal{-}measure$}$\,$\tnote{$4$}$\,\,\,$\textcolor{blue}{$\downarrow \uparrow$}\end{tabular}}} \\ \midrule[1.35pt]
\multirow{2}{*}{\textbf{Exp. 1}}  & LogMap                                                   & $0.881$                             & \textcolor{red}{\bm{$\uparrow$}}  \hfill\hfill\hfill\hfill\hfill\hfill\hfill\hfill\hfill\hfill\hfill\hfill\hfill\hfill\hfill\hfill\hfill\hfill\hfill\hfill\hfill\hfill\hfill\hfill\hfill\hfill\hfill\hfill\hfill\hfill\hfill $\textbf{0.882}$ & $0.881$ & \textcolor{red}{\bm{$\uparrow$}}  \hfill\hfill\hfill\hfill\hfill\hfill\hfill\hfill\hfill\hfill\hfill\hfill\hfill\hfill\hfill\hfill\hfill\hfill\hfill\hfill\hfill\hfill\hfill\hfill\hfill\hfill\hfill\hfill\hfill\hfill\hfill $\textbf{0.882}$                                              \\ 
                         & AML & $0.256$                        & \textcolor{red}{\bm{$\uparrow$}}  \hfill\hfill\hfill\hfill\hfill\hfill\hfill\hfill\hfill\hfill\hfill\hfill\hfill\hfill\hfill\hfill\hfill\hfill\hfill\hfill\hfill\hfill\hfill\hfill\hfill\hfill\hfill\hfill\hfill\hfill\hfill $0.342$ & \textcolor{SeaGreen3}{\bm{$\uparrow$}}  \hfill\hfill\hfill\hfill\hfill\hfill\hfill\hfill\hfill\hfill\hfill\hfill\hfill\hfill\hfill\hfill\hfill\hfill\hfill\hfill\hfill\hfill\hfill\hfill\hfill\hfill\hfill\hfill\hfill\hfill\hfill $0.281$ & \textcolor{blue}{\bm{$\uparrow$}}  \hfill\hfill\hfill\hfill\hfill\hfill\hfill\hfill\hfill\hfill\hfill\hfill\hfill\hfill\hfill\hfill\hfill\hfill\hfill\hfill\hfill\hfill\hfill\hfill\hfill\hfill\hfill\hfill\hfill\hfill\hfill $\textbf{0.364}$                                                              \\ \midrule
\multirow{2}{*}{\textbf{Exp. 2}}  & LogMap                                                       & $0.619$ & $0.619$ & $0.619$ & $0.619$                                              \\ 
                         & AML                & $0.627$                           & \textcolor{red}{\bm{$\uparrow$}}  \hfill\hfill\hfill\hfill\hfill\hfill\hfill\hfill\hfill\hfill\hfill\hfill\hfill\hfill\hfill\hfill\hfill\hfill\hfill\hfill\hfill\hfill\hfill\hfill\hfill\hfill\hfill\hfill\hfill\hfill\hfill $\textbf{0.629}$ & $0.627$ & \textcolor{red}{\bm{$\uparrow$}}  \hfill\hfill\hfill\hfill\hfill\hfill\hfill\hfill\hfill\hfill\hfill\hfill\hfill\hfill\hfill\hfill\hfill\hfill\hfill\hfill\hfill\hfill\hfill\hfill\hfill\hfill\hfill\hfill\hfill\hfill\hfill $\textbf{0.629}$                                                                   \\ \midrule
\multirow{2}{*}{\textbf{Exp. 3}}  & LogMap                                                       & $0.6$ & $0.6$ & $0.6$ & $0.6$                                               \\ 
                         & AML                                                       & $0.62$                           & \textcolor{red}{\bm{$\uparrow$}}  \hfill\hfill\hfill\hfill\hfill\hfill\hfill\hfill\hfill\hfill\hfill\hfill\hfill\hfill\hfill\hfill\hfill\hfill\hfill\hfill\hfill\hfill\hfill\hfill\hfill\hfill\hfill\hfill\hfill\hfill\hfill $\textbf{0.621}$ & $0.62$ & \textcolor{red}{\bm{$\uparrow$}}  \hfill\hfill\hfill\hfill\hfill\hfill\hfill\hfill\hfill\hfill\hfill\hfill\hfill\hfill\hfill\hfill\hfill\hfill\hfill\hfill\hfill\hfill\hfill\hfill\hfill\hfill\hfill\hfill\hfill\hfill\hfill $\textbf{0.621}$                                                                  \\ \midrule
\multirow{2}{*}{\textbf{Exp. 4}}  & LogMap                                                       & $0.448$ & $0.448$ & $0.448$ & $0.448$                                             \\ 
                         & AML                                                      & $0.056$ & \textcolor{red}{\bm{$\uparrow$}}  \hfill\hfill\hfill\hfill\hfill\hfill\hfill\hfill\hfill\hfill\hfill\hfill\hfill\hfill\hfill\hfill\hfill\hfill\hfill\hfill\hfill\hfill\hfill\hfill\hfill\hfill\hfill\hfill\hfill\hfill\hfill $0.222$ & \textcolor{SeaGreen3}{\bm{$\uparrow$}}  \hfill\hfill\hfill\hfill\hfill\hfill\hfill\hfill\hfill\hfill\hfill\hfill\hfill\hfill\hfill\hfill\hfill\hfill\hfill\hfill\hfill\hfill\hfill\hfill\hfill\hfill\hfill\hfill\hfill\hfill\hfill $\textbf{0.241}$ & \textcolor{blue}{\bm{$\uparrow$}}  \hfill\hfill\hfill\hfill\hfill\hfill\hfill\hfill\hfill\hfill\hfill\hfill\hfill\hfill\hfill\hfill\hfill\hfill\hfill\hfill\hfill\hfill\hfill\hfill\hfill\hfill\hfill\hfill\hfill\hfill\hfill $0.24$                                                                   \\ \midrule
\multirow{2}{*}{\textbf{Exp. 5}}  & LogMap                                                       & $0.042$ & $0.042$ & $0.042$ & $0.042$                                              \\ 
                         & AML                                    & $0.042$ & $0.042$ & $0.042$ & $0.042$                                                                   \\ \midrule
\multirow{2}{*}{\textbf{Exp. 6}}  & LogMap                                                       & $0.466$ & $0.466$ & $0.466$ & $0.466$                                             \\ 
                         & AML                                                      & $0.215$ & \textcolor{red}{\bm{$\uparrow$}}  \hfill\hfill\hfill\hfill\hfill\hfill\hfill\hfill\hfill\hfill\hfill\hfill\hfill\hfill\hfill\hfill\hfill\hfill\hfill\hfill\hfill\hfill\hfill\hfill\hfill\hfill\hfill\hfill\hfill\hfill\hfill $0.394$ & \textcolor{SeaGreen3}{\bm{$\uparrow$}}  \hfill\hfill\hfill\hfill\hfill\hfill\hfill\hfill\hfill\hfill\hfill\hfill\hfill\hfill\hfill\hfill\hfill\hfill\hfill\hfill\hfill\hfill\hfill\hfill\hfill\hfill\hfill\hfill\hfill\hfill\hfill $0.4$ & \textcolor{blue}{\bm{$\uparrow$}}  \hfill\hfill\hfill\hfill\hfill\hfill\hfill\hfill\hfill\hfill\hfill\hfill\hfill\hfill\hfill\hfill\hfill\hfill\hfill\hfill\hfill\hfill\hfill\hfill\hfill\hfill\hfill\hfill\hfill\hfill\hfill $\textbf{0.427}$                                                                   \\ \midrule
\multirow{2}{*}{\textbf{Exp. 7}}  & LogMap                                                       & $0.712$ & $0.712$ & $0.712$ & $0.712$                                             \\ 
                         & AML                                                      & $0.023$ & \textcolor{red}{\bm{$\uparrow$}}  \hfill\hfill\hfill\hfill\hfill\hfill\hfill\hfill\hfill\hfill\hfill\hfill\hfill\hfill\hfill\hfill\hfill\hfill\hfill\hfill\hfill\hfill\hfill\hfill\hfill\hfill\hfill\hfill\hfill\hfill\hfill $0.144$ & \textcolor{SeaGreen3}{\bm{$\uparrow$}}  \hfill\hfill\hfill\hfill\hfill\hfill\hfill\hfill\hfill\hfill\hfill\hfill\hfill\hfill\hfill\hfill\hfill\hfill\hfill\hfill\hfill\hfill\hfill\hfill\hfill\hfill\hfill\hfill\hfill\hfill\hfill $\textbf{0.171}$ & \textcolor{SeaGreen3}{\bm{$\uparrow$}}  \hfill\hfill\hfill\hfill\hfill\hfill\hfill\hfill\hfill\hfill\hfill\hfill\hfill\hfill\hfill\hfill\hfill\hfill\hfill\hfill\hfill\hfill\hfill\hfill\hfill\hfill\hfill\hfill\hfill\hfill\hfill $\textbf{0.171}$                                                                   \\ \midrule
\multirow{2}{*}{\textbf{Exp. 8}}  & LogMap                                                       & $0.987$ & $0.987$ & $0.987$ & $0.987$                                             \\ 
                         & AML                                                       & $0.976$ & \textcolor{red}{\bm{$\uparrow$}}  \hfill\hfill\hfill\hfill\hfill\hfill\hfill\hfill\hfill\hfill\hfill\hfill\hfill\hfill\hfill\hfill\hfill\hfill\hfill\hfill\hfill\hfill\hfill\hfill\hfill\hfill\hfill\hfill\hfill\hfill\hfill $\textbf{0.979}$ & $0.976$ & \textcolor{red}{\bm{$\uparrow$}}  \hfill\hfill\hfill\hfill\hfill\hfill\hfill\hfill\hfill\hfill\hfill\hfill\hfill\hfill\hfill\hfill\hfill\hfill\hfill\hfill\hfill\hfill\hfill\hfill\hfill\hfill\hfill\hfill\hfill\hfill\hfill $\textbf{0.979}$                                                                   \\ \midrule
\multirow{2}{*}{\textbf{Exp. 9}}                   & LogMap                                                       & $0.949$ & $0.949$ & $0.949$ & $0.949$                                              \\ 
                         & AML                                                       & $0.963$                          & \textcolor{red}{\bm{$\uparrow$}}  \hfill\hfill\hfill\hfill\hfill\hfill\hfill\hfill\hfill\hfill\hfill\hfill\hfill\hfill\hfill\hfill\hfill\hfill\hfill\hfill\hfill\hfill\hfill\hfill\hfill\hfill\hfill\hfill\hfill\hfill\hfill $\textbf{0.966}$ & $0.963$ & \textcolor{red}{\bm{$\uparrow$}}  \hfill\hfill\hfill\hfill\hfill\hfill\hfill\hfill\hfill\hfill\hfill\hfill\hfill\hfill\hfill\hfill\hfill\hfill\hfill\hfill\hfill\hfill\hfill\hfill\hfill\hfill\hfill\hfill\hfill\hfill\hfill $\textbf{0.966}$                                                                  \\ \midrule
\multirow{2}{*}{\textbf{Exp. 10}} & LogMap                                                       & $0.952$ & $0.952$ & $0.952$ & $0.952$                                               \\ 
                         & AML                                                       & $0.968$ & $0.968$ & $0.968$ & $0.968$                                                                  \\ \bottomrule
\end{tabular}
\begin{tablenotes}[para,flushleft]
        {\small
            \item[\boldmath{$1$}] in Table~\ref{tab1}. \item[\boldmath{$2$}] in Table~\ref{tab2}. \item[\boldmath{$3$}] in Table~\ref{tab3}. \item[\boldmath{$4$}] in Table~\ref{tab4}.
         }
    \end{tablenotes}
\end{threeparttable}
\end{adjustbox}
\end{table}

\begin{table}[p]
\centering
\caption{Comparison between \textit{Overall} Results in Different Tables.\label{tab44}}
\begin{adjustbox}{scale=0.9,center}
\begin{threeparttable}
\begin{tabular}{l|lV{3.5}>{\raggedleft}p{0.11\linewidth}||r|r|r}
\toprule
\multicolumn{1}{c|}{\textbf{Exp.}}                     & \textbf{Tool}   & \multicolumn{1}{c||}{ {\cellcolor{OliveDrab1}} \textbf{\begin{tabular}[c]{@{}c@{}}\boldmath{$Overall$}$\,$\tnote{$1$}\end{tabular}}} & \multicolumn{1}{c|}{ {\cellcolor{OliveDrab1}} \textbf{\begin{tabular}[c]{@{}c@{}}\boldmath{$Overall$}$\,$\tnote{$2$}$\,\,\,$\textcolor{red}{$\downarrow \uparrow$}\end{tabular}}} & \multicolumn{1}{c|}{ {\cellcolor{OliveDrab1}} \textbf{\begin{tabular}[c]{@{}c@{}}\boldmath{$Overall$}$\,$\tnote{$3$}$\,\,\,$\textcolor{SeaGreen3}{$\downarrow \uparrow$}\end{tabular}}} & \multicolumn{1}{c}{ {\cellcolor{OliveDrab1}} \textbf{\begin{tabular}[c]{@{}c@{}}\boldmath{$Overall$}$\,$\tnote{$4$}$\,\,\,$\textcolor{blue}{$\downarrow \uparrow$}\end{tabular}}} \\ \midrule[1.35pt]
\multirow{2}{*}{\textbf{Exp. 1}}  & LogMap                                                   & $0.786$                             & \textcolor{red}{\bm{$\uparrow$}} \hfill\hfill\hfill\hfill\hfill\hfill\hfill\hfill\hfill\hfill\hfill\hfill\hfill\hfill\hfill\hfill\hfill\hfill\hfill\hfill\hfill\hfill\hfill\hfill\hfill\hfill\hfill\hfill\hfill\hfill $\textbf{0.79}$ & $0.786$ & \textcolor{red}{\bm{$\uparrow$}} \hfill\hfill\hfill\hfill\hfill\hfill\hfill\hfill\hfill\hfill\hfill\hfill\hfill\hfill\hfill\hfill\hfill\hfill\hfill\hfill\hfill\hfill\hfill\hfill\hfill\hfill\hfill\hfill\hfill\hfill $\textbf{0.79}$                                              \\ 
                         & AML & $-\,0.49$                        & \textcolor{red}{\bm{$\uparrow$}} \hfill\hfill\hfill\hfill\hfill\hfill\hfill\hfill\hfill\hfill\hfill\hfill\hfill\hfill\hfill\hfill\hfill\hfill\hfill\hfill\hfill\hfill\hfill\hfill\hfill\hfill\hfill\hfill\hfill\hfill $0.052$ & \textcolor{SeaGreen3}{\bm{$\uparrow$}}  \hfill\hfill\hfill\hfill\hfill\hfill\hfill\hfill\hfill\hfill\hfill\hfill\hfill\hfill\hfill\hfill\hfill\hfill\hfill\hfill\hfill\hfill\hfill\hfill\hfill\hfill\hfill\hfill\hfill\hfill $-\,0.28$ & \textcolor{blue}{\bm{$\uparrow$}} \hfill\hfill\hfill\hfill\hfill\hfill\hfill\hfill\hfill\hfill\hfill\hfill\hfill\hfill\hfill\hfill\hfill\hfill\hfill\hfill\hfill\hfill\hfill\hfill\hfill\hfill\hfill\hfill\hfill\hfill $\bm{0.149}$                                                              \\ \midrule
\multirow{2}{*}{\textbf{Exp. 2}}  & LogMap                                                       & $0.448$ & $0.448$ & $0.448$ & $0.448$                                              \\ 
                         & AML                & $0.437$                           & \textcolor{red}{\bm{$\uparrow$}} \hfill\hfill\hfill\hfill\hfill\hfill\hfill\hfill\hfill\hfill\hfill\hfill\hfill\hfill\hfill\hfill\hfill\hfill\hfill\hfill\hfill\hfill\hfill\hfill\hfill\hfill\hfill\hfill\hfill\hfill $\textbf{0.448}$ & $0.437$ & \textcolor{red}{\bm{$\uparrow$}} \hfill\hfill\hfill\hfill\hfill\hfill\hfill\hfill\hfill\hfill\hfill\hfill\hfill\hfill\hfill\hfill\hfill\hfill\hfill\hfill\hfill\hfill\hfill\hfill\hfill\hfill\hfill\hfill\hfill\hfill $\textbf{0.448}$                                                                   \\ \midrule
\multirow{2}{*}{\textbf{Exp. 3}}  & LogMap                                                       & $0.433$ & $0.433$ & $0.433$ & $0.433$                                               \\ 
                         & AML                                                       & $0.421$                           & \textcolor{red}{\bm{$\uparrow$}} \hfill\hfill\hfill\hfill\hfill\hfill\hfill\hfill\hfill\hfill\hfill\hfill\hfill\hfill\hfill\hfill\hfill\hfill\hfill\hfill\hfill\hfill\hfill\hfill\hfill\hfill\hfill\hfill\hfill\hfill $\textbf{0.437}$ & $0.421$ & \textcolor{red}{\bm{$\uparrow$}} \hfill\hfill\hfill\hfill\hfill\hfill\hfill\hfill\hfill\hfill\hfill\hfill\hfill\hfill\hfill\hfill\hfill\hfill\hfill\hfill\hfill\hfill\hfill\hfill\hfill\hfill\hfill\hfill\hfill\hfill $\textbf{0.437}$                                                                  \\ \midrule
\multirow{2}{*}{\textbf{Exp. 4}}  & LogMap                                                       & $-\,0.163$ & $-\,0.163$ & $-\,0.163$ & $-\,0.163$                                             \\ 
                         & AML                                                      & $-\,5.61$ & \textcolor{red}{\bm{$\uparrow$}} \hfill\hfill\hfill\hfill\hfill\hfill\hfill\hfill\hfill\hfill\hfill\hfill\hfill\hfill\hfill\hfill\hfill\hfill\hfill\hfill\hfill\hfill\hfill\hfill\hfill\hfill\hfill\hfill\hfill\hfill $-\,0.145$ & \textcolor{SeaGreen3}{\bm{$\uparrow$}}  \hfill\hfill\hfill\hfill\hfill\hfill\hfill\hfill\hfill\hfill\hfill\hfill\hfill\hfill\hfill\hfill\hfill\hfill\hfill\hfill\hfill\hfill\hfill\hfill\hfill\hfill\hfill\hfill\hfill\hfill $-\,0.254$ & \textcolor{blue}{\bm{$\uparrow$}} \hfill\hfill\hfill\hfill\hfill\hfill\hfill\hfill\hfill\hfill\hfill\hfill\hfill\hfill\hfill\hfill\hfill\hfill\hfill\hfill\hfill\hfill\hfill\hfill\hfill\hfill\hfill\hfill\hfill\hfill $\bm{-\,0.03}$                                                                   \\ \midrule
\multirow{2}{*}{\textbf{Exp. 5}}  & LogMap                                                       & $0.021$ & $0.021$ & $0.021$ & $0.021$                                              \\ 
                         & AML                                    & $0.021$ & $0.021$ & $0.021$ & $0.021$                                                                   \\ \midrule
\multirow{2}{*}{\textbf{Exp. 6}}  & LogMap                                                       & $0.303$ & $0.303$ & $0.303$ & $0.303$                                             \\ 
                         & AML                                                      & $-1.0$ & \textcolor{red}{\bm{$\uparrow$}} \hfill\hfill\hfill\hfill\hfill\hfill\hfill\hfill\hfill\hfill\hfill\hfill\hfill\hfill\hfill\hfill\hfill\hfill\hfill\hfill\hfill\hfill\hfill\hfill\hfill\hfill\hfill\hfill\hfill\hfill $0.156$ & \textcolor{SeaGreen3}{\bm{$\uparrow$}}  \hfill\hfill\hfill\hfill\hfill\hfill\hfill\hfill\hfill\hfill\hfill\hfill\hfill\hfill\hfill\hfill\hfill\hfill\hfill\hfill\hfill\hfill\hfill\hfill\hfill\hfill\hfill\hfill\hfill\hfill $0.196$ & \textcolor{blue}{\bm{$\uparrow$}} \hfill\hfill\hfill\hfill\hfill\hfill\hfill\hfill\hfill\hfill\hfill\hfill\hfill\hfill\hfill\hfill\hfill\hfill\hfill\hfill\hfill\hfill\hfill\hfill\hfill\hfill\hfill\hfill\hfill\hfill $\textbf{0.264}$                                                                   \\ \midrule
\multirow{2}{*}{\textbf{Exp. 7}}  & LogMap                                                       & $0.552$ & $0.552$ & $0.552$ & $0.552$                                             \\ 
                         & AML                                                      & $-\,6.871$ & \textcolor{red}{\bm{$\uparrow$}} \hfill\hfill\hfill\hfill\hfill\hfill\hfill\hfill\hfill\hfill\hfill\hfill\hfill\hfill\hfill\hfill\hfill\hfill\hfill\hfill\hfill\hfill\hfill\hfill\hfill\hfill\hfill\hfill\hfill\hfill $-\,0.136$ & \textcolor{SeaGreen3}{\bm{$\uparrow$}}  \hfill\hfill\hfill\hfill\hfill\hfill\hfill\hfill\hfill\hfill\hfill\hfill\hfill\hfill\hfill\hfill\hfill\hfill\hfill\hfill\hfill\hfill\hfill\hfill\hfill\hfill\hfill\hfill\hfill\hfill $\textbf{0.092}$ & \textcolor{SeaGreen3}{\bm{$\uparrow$}}  \hfill\hfill\hfill\hfill\hfill\hfill\hfill\hfill\hfill\hfill\hfill\hfill\hfill\hfill\hfill\hfill\hfill\hfill\hfill\hfill\hfill\hfill\hfill\hfill\hfill\hfill\hfill\hfill\hfill\hfill $\bm{0.092}$                                                                   \\ \midrule
\multirow{2}{*}{\textbf{Exp. 8}}  & LogMap                                                       & $0.974$ & $0.974$ & $0.974$ & $0.974$                                             \\ 
                         & AML                                                       & $0.953$ & \textcolor{red}{\bm{$\uparrow$}}  \hfill\hfill\hfill\hfill\hfill\hfill\hfill\hfill\hfill\hfill\hfill\hfill\hfill\hfill\hfill\hfill\hfill\hfill\hfill\hfill\hfill\hfill\hfill\hfill\hfill\hfill\hfill\hfill\hfill\hfill $\textbf{0.958}$ & $0.953$ & \textcolor{red}{\bm{$\uparrow$}}  \hfill\hfill\hfill\hfill\hfill\hfill\hfill\hfill\hfill\hfill\hfill\hfill\hfill\hfill\hfill\hfill\hfill\hfill\hfill\hfill\hfill\hfill\hfill\hfill\hfill\hfill\hfill\hfill\hfill\hfill $\textbf{0.958}$                                                                   \\ \midrule
\multirow{2}{*}{\textbf{Exp. 9}}                   & LogMap                                                       & $0.9$ & $0.9$ & $0.9$ & $0.9$                                              \\ 
                         & AML                                                       & $0.928$                          & \textcolor{red}{\bm{$\uparrow$}}  \hfill\hfill\hfill\hfill\hfill\hfill\hfill\hfill\hfill\hfill\hfill\hfill\hfill\hfill\hfill\hfill\hfill\hfill\hfill\hfill\hfill\hfill\hfill\hfill\hfill\hfill\hfill\hfill\hfill\hfill $\textbf{0.933}$ & $0.928$ & \textcolor{red}{\bm{$\uparrow$}}  \hfill\hfill\hfill\hfill\hfill\hfill\hfill\hfill\hfill\hfill\hfill\hfill\hfill\hfill\hfill\hfill\hfill\hfill\hfill\hfill\hfill\hfill\hfill\hfill\hfill\hfill\hfill\hfill\hfill\hfill $\textbf{0.933}$                                                                  \\ \midrule
\multirow{2}{*}{\textbf{Exp. 10}} & LogMap                                                       & $0.91$ & $0.91$ & $0.91$ & $0.91$                                               \\ 
                         & AML                                                       & $0.939$ & $0.939$ & $0.939$ & $0.939$                                                                  \\ \bottomrule
\end{tabular}
\begin{tablenotes}[para,flushleft]
        {\small \item[\boldmath{$1$}] in Table~\ref{tab1}. \item[\boldmath{$2$}] in Table~\ref{tab2}. \item[\boldmath{$3$}] in Table~\ref{tab3}. \item[\boldmath{$4$}] in Table~\ref{tab4}.
         }
    \end{tablenotes}
\end{threeparttable}
\end{adjustbox}

\bigskip
\bigskip

\centering
\caption{Comparison between \textit{Ambiguity} Results in Different Tables.\label{tab55}}
\begin{adjustbox}{scale=0.9,center}
\begin{threeparttable}
\begin{tabular}{l|lV{3.5}>{\raggedleft}p{0.145\linewidth}||>{\columncolor[gray]{0.83}}r|r|>{\columncolor[gray]{0.83}}r}
\toprule
\multicolumn{1}{c|}{\textbf{Exp.}}                     & \textbf{Tool}   & \multicolumn{1}{c||}{ {\cellcolor{Salmon1}} \textbf{\begin{tabular}[c]{@{}c@{}}\boldmath{$Ambiguity$}$\,$\tnote{$1$}\end{tabular}}} & \multicolumn{1}{c|}{ {\cellcolor{Salmon1}} \textbf{\begin{tabular}[c]{@{}c@{}}\boldmath{$Ambiguity$}$\,$\tnote{$2$}$\;\;$\textcolor{red}{$\downarrow \uparrow$}\end{tabular}}} & \multicolumn{1}{c|}{ {\cellcolor{Salmon1}} \textbf{\begin{tabular}[c]{@{}c@{}}\boldmath{$Ambiguity$}$\,$\tnote{$3$}$\;\;$\textcolor{SeaGreen3}{$\downarrow \uparrow$}\end{tabular}}} & \multicolumn{1}{c}{ {\cellcolor{Salmon1}} \textbf{\begin{tabular}[c]{@{}c@{}}\boldmath{$Ambiguity$}$\,$\tnote{$4$}$\;\;$\textcolor{blue}{$\downarrow \uparrow$}\end{tabular}}} \\ \midrule[1.35pt]
\multirow{2}{*}{\textbf{Exp. 1}}  & LogMap                                                   & $0.94\,\%$                             & \cellcolor[gray]{0.83} & $0.94\,\%$ & \cellcolor[gray]{0.83}                                              \\ 
                         & AML & $68.88\,\%$                        & \multirow{-2}{*}{\textcolor{red}{\bm{$\downarrow$}} \qquad\;\;\;\;\;\;\quad$\textbf{0\%}$} & \textcolor{SeaGreen3}{\bm{$\downarrow$}} \hfill\hfill\hfill\hfill\hfill\hfill\hfill\hfill\hfill\hfill\hfill\hfill\hfill\hfill\hfill\hfill\hfill\hfill\hfill\hfill\hfill\hfill\hfill\hfill\hfill\hfill\hfill\hfill\hfill\hfill $66\,\%$ & \multirow{-2}{*}{\textcolor{red}{\bm{$\downarrow$}} \qquad\;\;\;\;\;\;\quad$\textbf{0\%}$}                                                              \\ \midrule
\multirow{2}{*}{\textbf{Exp. 2}}  & LogMap                                                       & $0\,\%$ & \cellcolor[gray]{0.83} & $0\,\%$ & \cellcolor[gray]{0.83}                                              \\ 
                         & AML                & $6.6\,\%$                           & \multirow{-2}{*}{\textcolor{red}{\bm{$\downarrow$}} \qquad\;\;\;\;\;\;\quad$\textbf{0\%}$}  & $6.6\,\%$ & \multirow{-2}{*}{\textcolor{red}{\bm{$\downarrow$}} \qquad\;\;\;\;\;\;\quad$\textbf{0\%}$}                                                                    \\ \midrule
\multirow{2}{*}{\textbf{Exp. 3}}  & LogMap                                                       & $0\,\%$ & \cellcolor[gray]{0.83} & $0\,\%$ & \cellcolor[gray]{0.83}                                               \\ 
                         & AML                                                       & $11\,\%$                           & \multirow{-2}{*}{\textcolor{red}{\bm{$\downarrow$}} \qquad\;\;\;\;\;\;\quad$\textbf{0\%}$} & $11\,\%$ & \multirow{-2}{*}{\textcolor{red}{\bm{$\downarrow$}} \qquad\;\;\;\;\;\;\quad$\textbf{0\%}$}                                                                  \\ \midrule
\multirow{2}{*}{\textbf{Exp. 4}}  & LogMap                                                       & $0\,\%$ & \cellcolor[gray]{0.83} & $0\,\%$ & \cellcolor[gray]{0.83}                                             \\ 
                         & AML                                                      & $95.77\,\%$ & \multirow{-2}{*}{\textcolor{red}{\bm{$\downarrow$}} \qquad\;\;\;\;\;\;\quad$\textbf{0\%}$} & \textcolor{SeaGreen3}{\bm{$\downarrow$}}  \hfill\hfill\hfill\hfill\hfill\hfill\hfill\hfill\hfill\hfill\hfill\hfill\hfill\hfill\hfill\hfill\hfill\hfill\hfill\hfill\hfill\hfill\hfill\hfill\hfill\hfill\hfill\hfill\hfill\hfill $63.88\,\%$ &    \multirow{-2}{*}{\textcolor{red}{\bm{$\downarrow$}} \qquad\;\;\;\;\;\;\quad$\textbf{0\%}$}                                                                \\ \midrule
\multirow{2}{*}{\textbf{Exp. 5}}  & LogMap                                                       & $0\,\%$ & \cellcolor[gray]{0.83} & $0\,\%$ & \cellcolor[gray]{0.83}                                              \\ 
                         & AML                                    & $0\,\%$ & \multirow{-2}{*}{\textcolor{red}{\bm{$\downarrow$}} \qquad\;\;\;\;\;\;\quad$\textbf{0\%}$} & $0\,\%$ & \multirow{-2}{*}{\textcolor{red}{\bm{$\downarrow$}} \qquad\;\;\;\;\;\;\quad$\textbf{0\%}$}                                                                   \\ \midrule
\multirow{2}{*}{\textbf{Exp. 6}}  & LogMap                                                       & $0\,\%$ & \cellcolor[gray]{0.83} & $0\,\%$ & \cellcolor[gray]{0.83}                                             \\ 
                         & AML                                                      & $81.64\,\%$ & \multirow{-2}{*}{\textcolor{red}{\bm{$\downarrow$}} \qquad\;\;\;\;\;\;\quad$\textbf{0\%}$} & \textcolor{SeaGreen3}{\bm{$\downarrow$}}  \hfill\hfill\hfill\hfill\hfill\hfill\hfill\hfill\hfill\hfill\hfill\hfill\hfill\hfill\hfill\hfill\hfill\hfill\hfill\hfill\hfill\hfill\hfill\hfill\hfill\hfill\hfill\hfill\hfill\hfill $22.22\,\%$ &    \multirow{-2}{*}{\textcolor{red}{\bm{$\downarrow$}} \qquad\;\;\;\;\;\;\quad$\textbf{0\%}$}                                                                \\ \midrule
\multirow{2}{*}{\textbf{Exp. 7}}  & LogMap                                                       & $0\,\%$ & \cellcolor[gray]{0.83} & $0\,\%$ & \cellcolor[gray]{0.83}                                             \\ 
                         & AML                                                      & $98.58\,\%$ & \multirow{-2}{*}{\textcolor{red}{\bm{$\downarrow$}} \qquad\;\;\;\;\;\;\quad$\textbf{0\%}$} & \textcolor{SeaGreen3}{\bm{$\downarrow$}}  \hfill\hfill\hfill\hfill\hfill\hfill\hfill\hfill\hfill\hfill\hfill\hfill\hfill\hfill\hfill\hfill\hfill\hfill\hfill\hfill\hfill\hfill\hfill\hfill\hfill\hfill\hfill\hfill\hfill\hfill $0\,\%$ & \multirow{-2}{*}{\textcolor{red}{\bm{$\downarrow$}} \qquad\;\;\;\;\;\;\quad$\textbf{0\%}$}                                                                  \\ \midrule
\multirow{2}{*}{\textbf{Exp. 8}}  & LogMap                                                       & $0\,\%$ & \cellcolor[gray]{0.83} & $0\,\%$ & \cellcolor[gray]{0.83}                                             \\ 
                         & AML                                                       & $1\,\%$ & \multirow{-2}{*}{\textcolor{red}{\bm{$\downarrow$}} \qquad\;\;\;\;\;\;\quad$\textbf{0\%}$} & $1\,\%$ & \multirow{-2}{*}{\textcolor{red}{\bm{$\downarrow$}} \qquad\;\;\;\;\;\;\quad$\textbf{0\%}$}                                                                   \\ \midrule
\multirow{2}{*}{\textbf{Exp. 9}}                   & LogMap                                                       & $0\,\%$ & \cellcolor[gray]{0.83} & $0\,\%$ & \cellcolor[gray]{0.83}                                              \\ 
                         & AML                                                       & $1\,\%$                          & \multirow{-2}{*}{\textcolor{red}{\bm{$\downarrow$}} \qquad\;\;\;\;\;\;\quad$\textbf{0\%}$} & $1\,\%$ & \multirow{-2}{*}{\textcolor{red}{\bm{$\downarrow$}} \qquad\;\;\;\;\;\;\quad$\textbf{0\%}$}                                                                  \\ \midrule
\multirow{2}{*}{\textbf{Exp. 10}} & LogMap                                                       & $0\,\%$ & \cellcolor[gray]{0.83} & $0\,\%$ & \cellcolor[gray]{0.83}                                               \\ 
                         & AML                                                       & $0\,\%$ & \multirow{-2}{*}{\textcolor{red}{\bm{$\downarrow$}} \qquad\;\;\;\;\;\;\quad$\textbf{0\%}$} & $0\,\%$ & \multirow{-2}{*}{\textcolor{red}{\bm{$\downarrow$}} \qquad\;\;\;\;\;\;\quad$\textbf{0\%}$}                                                                  \\ \bottomrule
\end{tabular}
\begin{tablenotes}[para,flushleft]
        {\small
            \item[\boldmath{$1$}] in Table~\ref{tab1}. \item[\boldmath{$2$}] in Table~\ref{tab2}. \item[\boldmath{$3$}] in Table~\ref{tab3}. \item[\boldmath{$4$}] in Table~\ref{tab4}.
         }
    \end{tablenotes}
\end{threeparttable}
\end{adjustbox}
\end{table}

When we compare Tables~\ref{tab1} and  \ref{tab2}, we notice that there is a very negligible decrease in \textit{recall} scores in Experiments 1, 2, 3, and 4 for AML results (see Table~\ref{tab22}) because of the alignment \textit{disambiguation} process. However, there is a noticeable slight improvement in \textit{precision}, \textit{F-measure}, and \textit{overall} scores in all experiments (see Tables~\ref{tab11}, \ref{tab33}, and \ref{tab44}). There is also a slight improvement in \textit{precision}, \textit{F-measure}, and \textit{overall} scores in Experiment 1 for LogMap (see Tables~\ref{tab11}, \ref{tab33} and \ref{tab44}). We deduce that the sets of ambiguous correspondences do surely convey some false correspondences (due to their obvious uncertainty). So, after disambiguating the output alignments, all the alignment evaluation scores improve, and thus the global uncertainty of alignments decreases.

As shown in Tables~\ref{tab1} and~\ref{tab3}, AML generates a higher number of ambiguous correspondences ($Amb$) than does LogMap, especially in Experiments~1, 4, 6, and 7 (see Table~\ref{tab55}). Therefore, AML alignments contain more uncertainty cases than LogMap alignments. In Subsection~\ref{ex_amb}, we will show some ambiguity examples. When we compare Table~\ref{tab1} and Table~\ref{tab3}, we notice that there is a very negligible decrease in \textit{recall} scores in Experiments 1 and 7 for AML results (see Table~\ref{tab22}) because of the alignment \textit{trimming} process. However, there is a noticeable improvement in \textit{precision}, \textit{F-measure}, \textit{overall}, and \textit{ambiguity degree} scores in Experiments 1, 4, 6, and 7 (see Tables~\ref{tab11}, \ref{tab33}, \ref{tab44} and \ref{tab55}). We deduce that the removed (trimmed) correspondences do surely convey some false correspondences (due to their low confidence values). So after trimming the output alignments, all the alignment evaluation scores improve, and thus the global uncertainty of alignments decreases.

When we compare Table~\ref{tab1} and Table~\ref{tab4}, we notice that there is a very negligible decrease in \textit{recall} scores in Experiments 1, 2, 3, 4, and 7 for AML results (see Table~\ref{tab22}) because of the alignment \textit{trimming} and \textit{disambiguation} processes. However, there is a noticeable improvement in \textit{precision}, \textit{F-measure}, and \textit{overall} scores in all experiments (see Tables~\ref{tab11}, \ref{tab33}, and \ref{tab44}). We deduce that the deleted correspondences—after \textit{trimming} and \textit{disambiguation}--do surely convey false correspondences (due to their ambiguity and low confidence). Therefore, after trimming and disambiguating the output alignments, all the evaluation scores improve, and thus the global uncertainty of these alignments is reduced.

Among the four result tables (Tables~\ref{tab1}, \ref{tab2}, \ref{tab3},~and \ref{tab4}), Table~\ref{tab4} represents the best results. Therefore, Table~\ref{tab4} can be considered as the final results of our case study. By trimming and disambiguating the output alignments in Table~\ref{tab4}, we exclude the maximum of untrustworthy correspondences, and we minimize the number of false positives in the alignments as much as possible. Despite that, experiments in Table~\ref{tab4} still point out a significant number of missing correspondences (with a significant number of false negatives) and a more contained number of false positives. Hence, the resulting ontology alignments are still uncertain and thus not entirely reliable for the task of ontology integration.

By and large, based on the set of experiments performed, LogMap outperforms its competitor, AML. However, it is true that the experience of a real-world case study has amply demonstrated the uncertainty that the high number of returned false positives and false negatives quantitatively reflect. In the next section, we will explain the reasons for the weak performance of LogMap and AML in some experiments, in particular Experiments 5 and 7 (see Subsection~\ref{causes}).

\section{Lessons Learned from the Case Study\label{section7}}

\subsection{Uncertainty Causes in Ontology Matching}

In this subsection, we explain the most important uncertainty causes that we have encountered in our experiments.

\subsubsection{Uncertainty Caused by Different Ontology Granularity\label{causes}}

Current ontology matching tools face many difficulties when one ontology is more detailed or more general than the other. For example, an ontology $\mathcal{O}_1$ contains a concept that has the name of a specific country, while an ontology $\mathcal{O}_2$ contains concepts that have the names of sub-countries constituting the country of $\mathcal{O}_1$.

Table~\ref{gran} shows some examples from Experiment~1. In the first example of Table~\ref{gran}, LogMap and AML identified that "\texttt{Sudan}" in $\mathcal{O}_1$ is equivalent to "\texttt{Sudan}" in $\mathcal{O}_2$, which is false in this case. \texttt{Sudan} in $\mathcal{O}_2$ is actually the North Sudan, and \texttt{Sudan} in $\mathcal{O}_1$ is the former Sudan that is composed of both the northern and southern parts. The entities "\texttt{Sudan}" and "\texttt{South Sudan}" from $\mathcal{O}_2$ should rather be sub-entities of "\texttt{Sudan}" from $\mathcal{O}_1$ (using a \textit{subsumption} relation, not an \textit{equivalence} relation). Ontology matching tools cannot identify such \textit{subsumption} correspondences between entities.

\setcounter{table}{0}

\begin{table}[!htb]
\centering
\caption{Different Ontology Granularity (Subsumption Matching).}\label{gran}
\begin{adjustbox}{scale=0.9,center}
\begin{tabular}{l|l}
\toprule
\multicolumn{1}{c|}{\textbf{Ontology}~\boldmath{$\mathcal{O}_1$}}                  & \multicolumn{1}{c}{\textbf{Ontology}~\boldmath{$\mathcal{O}_2$}}                                        \\ \midrule[1.35pt]

$\bullet$ \texttt{Sudan}                                                          & \begin{tabular}[c]{@{}l@{}}$\bullet$ \texttt{Sudan}\\ $\bullet$ \texttt{South\_Sudan}\end{tabular}                          \\ \midrule
\begin{tabular}[c]{@{}l@{}}$\bullet$ \texttt{Gaza\_Strip}\\ $\bullet$ \texttt{West\_Bank}\end{tabular} & $\bullet$ \texttt{State\_of\_Palestine}                                                                   \\ \midrule
$\bullet$ \texttt{Netherlands\_Antilles}                                                             &  \begin{tabular}[c]{@{}l@{}} $\bullet$ \texttt{Sint\_Maarten\_(Dutch\_part)}\\ $\bullet$ \texttt{Bonaire,\_Sint Eustatius\_and\_Saba}\end{tabular}                                                                                    \\ \midrule
\begin{tabular}[c]{@{}l@{}} $\bullet$ \texttt{Jersey}\\ $\bullet$ \texttt{Guernsey}\end{tabular}                                                             &  $\bullet$ \texttt{Channel\_Islands}  \\ \midrule                     $\bullet$ \texttt{Agriculture}                                                  & \begin{tabular}[c]{@{}l@{}}$\bullet$ \texttt{Economy\_Agriculture}\\ $\bullet$ \texttt{Employment\_Agriculture}\end{tabular}                                                              \\ \bottomrule

\end{tabular}
\end{adjustbox}
\end{table}

Notice that $\mathcal{O}_1$ and $\mathcal{O}_2$ have a different granularity in each of these examples (in Table~\ref{gran}). That is, entities of $\mathcal{O}_1$ are not always more detailed than entities of $\mathcal{O}_2$; and entities of $\mathcal{O}_1$ are not always more general than entities of $\mathcal{O}_2$. Thus, we cannot say that $\mathcal{O}_1$ is more detailed (or more general) than $\mathcal{O}_2$. As a consequence, we cannot automatically replace every ambiguous \textit{equivalence} correspondence in the alignment with a \textit{subsumption} correspondence (as does the second approach of alignment disambiguation (see Subsection~\ref{approche2})).

We can add another example from Experiment~3, where $\mathcal{O}_1$ has datatype properties (or attributes) in the form of "year-year" intervals (\textit{e.g.}, "\texttt{97-1999}"), and $\mathcal{O}_2$ has attributes as separate years (\textit{e.g.}, "\texttt{1997}", "\texttt{1998}", "\texttt{1999}").

There are many more complicated cases, such as complex correspondences. To identify complex correspondences, we need to perform a \textit{complex matching} process. In complex matching, a complex correspondence can be composed of an entity from ontology $\mathcal{O}_1$ and a union of entities from ontology $\mathcal{O}_2$, or vice versa. Complex correspondences are extremely hard to identify. Table~\ref{complex} shows some examples of complex correspondences extracted from Experiment~1.

\begin{table}[!htb]
\captionsetup{justification=centering}
\centering
\begin{minipage}[]{0.3\linewidth}
\caption{Different Ontology Granularity (Complex Matching).\label{complex}}
\end{minipage}
\begin{adjustbox}{scale=0.9,center}
\begin{threeparttable}
\begin{tabular}{l|l}
\toprule
\multicolumn{1}{c|}{\textbf{Ontology}~\boldmath{$\mathcal{O}_1$}}                  & \multicolumn{1}{c}{\textbf{Ontology}~\boldmath{$\mathcal{O}_2$}}                                        \\ \midrule[1.35pt]

\begin{tabular}[c]{@{}l@{}}$\bullet$ \textcolor{red}{\texttt{Near\_East}}\\ $\bullet$ \textcolor{blue}{\texttt{Asia$\_$(EX.$\_$Near$\_$East)}}\tnote{$\star$}\end{tabular}                                                   & \begin{tabular}[c]{@{}l@{}}$\bullet$ \textcolor{red}{\texttt{WesternAsia}}\\ $\bullet$ \textcolor{blue}{\texttt{CentralAsia}} \\ $\bullet$ \textcolor{blue}{\texttt{EasternAsia}}\\ $\bullet$ \textcolor{blue}{\texttt{SouthernAsia}}\\ $\bullet$ \textcolor{blue}{\texttt{South-easternAsia}} \end{tabular} \\ \midrule

\begin{tabular}[c]{@{}l@{}}$\bullet$ \textcolor{red}{\texttt{Northern$\_$Africa}}\\ $\bullet$ \textcolor{blue}{\texttt{Sub$-$Saharan$\_$Africa}}\end{tabular}                                                          & \begin{tabular}[c]{@{}l@{}}$\bullet$ \textcolor{red}{\texttt{NorthernAfrica}}\\ $\bullet$ \textcolor{blue}{\texttt{MiddleAfrica}}\\ $\bullet$ \textcolor{blue}{\texttt{EasternAfrica}}\\ $\bullet$ \textcolor{blue}{\texttt{WesternAfrica}}\\ $\bullet$ \textcolor{blue}{\texttt{SouthernAfrica}} \end{tabular} \\ \midrule

\begin{tabular}[c]{@{}l@{}}$\bullet$ \textcolor{red}{\texttt{Northern$\_$America}}\\ $\bullet$ \textcolor{blue}{\texttt{Latin$\_$Amer$.\_$and$\_$Carib}}\end{tabular} &
\begin{tabular}[c]{@{}l@{}}$\bullet$ \textcolor{red}{\texttt{NorthernAmerica}}\\  $\bullet$ \textcolor{blue}{\texttt{CentralAmerica}} \\ $\bullet$ \textcolor{blue}{\texttt{SouthAmerica}}\\ $\bullet$ \textcolor{blue}{\texttt{Caribbean}}\\\end{tabular}                   \\ \bottomrule

\end{tabular}
\begin{tablenotes}[para,flushleft]
        {\small \item[$\star$] EX. = Except
         }
    \end{tablenotes}
\end{threeparttable}
\end{adjustbox}
\end{table}

The entity "\texttt{Near\_East}" from $\mathcal{O}_1$ should be matched to the entity "\texttt{WesternAsia}" from $\mathcal{O}_2$. While the entity "\texttt{Asia\_(EX.\_Near\_East)}" should be matched to the union of the entities "\texttt{CentralAsia}", "\texttt{EasternAsia}", "\texttt{SouthernAsia}", and "\texttt{South-easternAsia}". This is a complex \textit{compound} correspondence. Otherwise, this complex correspondence can be split into four \textit{subsumption} correspondences. None of these correspondences were detected by LogMap or AML.

The entities "\texttt{Northern\_Africa}" and "\texttt{NorthernAfrica}" were correctly matched by both LogMap and AML. However, it was impossible for them to detect that "\texttt{Sub-Saharan\_Africa}" is the union of "\texttt{MiddleAfrica}", "\texttt{EasternAfrica}", "\texttt{WesternAfrica}", and "\texttt{SouthernAfrica}". Complex matching is very difficult to perform with automated ontology matching systems. Only a human being can figure out such correspondences.

Similarly, "\texttt{Northern\_America}" and "\texttt{NorthernAmerica}" were successfully matched by both LogMap and AML. However, it is not evident to match "\texttt{Latin\_Amer.\_and\_Carib}" with the union of "\texttt{Central\_America}", "\texttt{South\_America}", and "\texttt{Caribbean}" (in fact, Latin America is composed of Central America + South America).

In the next example, we will encounter a more complicated matching case where the same entities are simultaneously involved in two types of correspondences. Tables~\ref{complexxxxxx} and~\ref{complexxxxx} show an example from Experiment~3. In this experiment, LogMap and AML have successfully matched "\texttt{China,\_Hong\_Kong\_SAR}" from $\mathcal{O}_1$ to "\texttt{Hong\_Kong\_SAR,\_China}" from $\mathcal{O}_2$, and "\texttt{China,\_Macao\_SAR}" from $\mathcal{O}_1$ to "\texttt{Macao\_SAR,\_China}" from $\mathcal{O}_2$. However, they both falsely matched "\texttt{China}" from $\mathcal{O}_1$ to "\texttt{China}" from $\mathcal{O}_2$. The correct correspondence should rather match "\texttt{China,\_mainland}" from $\mathcal{O}_1$ to "\texttt{China}" from $\mathcal{O}_2$. Indeed, the entity "\texttt{China}" in $\mathcal{O}_1$ actually means the whole China territory (including the mainland of China, Hong Kong, and Macao). Therefore, "\texttt{China}" from $\mathcal{O}_1$ should be matched to the union of "\texttt{China}", "\texttt{Hong\_Kong\_SAR,\_China}", and "\texttt{Macao\_SAR,\_China}" from $\mathcal{O}_2$ (or split into three \textit{subsumption} correspondences). Therefore, the entities "\texttt{China}", "\texttt{Hong\_Kong\_SAR,\_China}", and "\texttt{Macao\_SAR,\_China}" from $\mathcal{O}_2$ are simultaneously involved in \textit{equivalence} correspondences (see Table~\ref{complexxxxxx}) and \textit{subsumption} correspondences (see Table~\ref{complexxxxx}).

\begin{table}[!htb]
\centering
\begin{minipage}[t]{0.47\linewidth}\centering
\caption{Equivalence Matching Example.\label{complexxxxxx}}
\begin{adjustbox}{scale=0.9,center}
\begin{threeparttable}
\begin{tabular}{l|l}
\toprule
\multicolumn{1}{c|}{\textbf{Ontology}~\boldmath{$\mathcal{O}_1$}}                  & \multicolumn{1}{c}{\textbf{Ontology}~\boldmath{$\mathcal{O}_2$}}                                  \\\midrule

\begin{tabular}[c]{@{}l@{}}$\bullet$ \texttt{China}\\ $\bullet$ \textcolor{red}{\texttt{China,\_mainland}} \\ $\bullet$ \textcolor{blue}{\texttt{China,\_Hong\_Kong\_SAR}} \\ $\bullet$ \textcolor{cyan}{\texttt{China,\_Macao\_SAR}} \end{tabular} &

\begin{tabular}[c]{@{}l@{}}$\bullet$ \textcolor{red}{\texttt{China}}\\  $\bullet$ \textcolor{blue}{\texttt{Hong\_Kong\_SAR,\_China}} \\ $\bullet$ \textcolor{cyan}{\texttt{Macao\_SAR,\_China}}\end{tabular}                   \\ \bottomrule

\end{tabular}
\end{threeparttable}
\end{adjustbox}
\end{minipage}\hfill%
\begin{minipage}[t]{0.47\linewidth}\centering
\caption{Complex (or Subsumption) Matching Example.\label{complexxxxx}}
\begin{adjustbox}{scale=0.9,center}
\begin{threeparttable}
\begin{tabular}{l|l}
\toprule
\multicolumn{1}{c|}{\textbf{Ontology}~\boldmath{$\mathcal{O}_1$}}                  & \multicolumn{1}{c}{\textbf{Ontology}~\boldmath{$\mathcal{O}_2$}}                                  \\\midrule

\begin{tabular}[c]{@{}l@{}}$\bullet$ \textcolor{magenta}{\texttt{China}}\\ $\bullet$ \texttt{China,\_mainland} \\ $\bullet$ \textcolor{black}{\texttt{China,\_Hong\_Kong\_SAR}} \\ $\bullet$ \textcolor{black}{\texttt{China,\_Macao\_SAR}}  \end{tabular} &

\begin{tabular}[c]{@{}l@{}}$\bullet$ \textcolor{magenta}{\texttt{China}}\\  $\bullet$ \textcolor{magenta}{\texttt{Hong\_Kong\_SAR,\_China}} \\ $\bullet$ \textcolor{magenta}{\texttt{Macao\_SAR,\_China}}\end{tabular}                   \\ \bottomrule

\end{tabular}
\end{threeparttable}
\end{adjustbox}
\end{minipage}
\end{table}

We conclude that ontologies belonging to the same domain and having different granularities certainly introduce many cases of uncertainty.

\subsubsection{Uncertainty Caused by Different Entity Expressiveness}

Ontology matching tools are also very sensitive to the expressiveness of entities. Consider the following example extracted from Experiment~4:
\begin{table}[H]
\centering
\begin{adjustbox}{scale=0.9,center}
\begin{tabular}{l|l}
\toprule
\multicolumn{1}{c|}{\textbf{Ontology}~\boldmath{$\mathcal{O}_1$}}                  & \multicolumn{1}{c}{\textbf{$\qquad$Ontology~\boldmath{$\mathcal{O}_2$}$\qquad$}}                                        \\ \midrule

$\bullet$ \texttt{Regions.Subregions.Countries}                                                  & $\,\,\,\,$$\bullet$ \texttt{Country\_Name}                 \\ \bottomrule

\end{tabular}
\end{adjustbox}
\end{table}

In Experiment~4, ontology $\mathcal{O}_1$ contains a class that is named "\texttt{Regions.Subregions.Countries}". However, the instances of that class are only regions and sub-regions; it has no countries as instances. This class should be instead named "\textit{Regions.Subregions}". On the contrary, ontology $\mathcal{O}_2$ contains a class that is named "\texttt{Country\_Name}." However, instances of that class are countries as well as regions and sub-regions. This class should be rather named "\textit{Regions.Subregions.Countries}"; we notice that there is inherent uncertainty in the original datasets. The names and terms of the classes are originally not expressive enough and do not perfectly represent the intended meaning. This feature makes the ontology-matching task even more difficult. Here is another example from Experiment~6:

\begin{table}[H]
\centering
\begin{adjustbox}{scale=0.9,center}
\begin{tabular}{l|l}
\toprule
\multicolumn{1}{c|}{\textbf{$\quad$Ontology~\boldmath{$\mathcal{O}_1$}$\quad$}}                  & \multicolumn{1}{c}{\textbf{Ontology}~\boldmath{$\mathcal{O}_2$}}                                        \\ \midrule

\begin{tabular}[c]{@{}l@{}}$\bullet$ \texttt{Short\_Name}\\ $\bullet$ \texttt{Table\_Name} \\ $\bullet$ \texttt{Long\_Name}\end{tabular}                                                   & $\bullet$ \texttt{Country\_or\_Area}                 \\ \bottomrule

\end{tabular}
\end{adjustbox}
\end{table}

In Experiment~6, ontology $\mathcal{O}_1$ has three datatype properties (or attributes): ~"\texttt{Short\_Name}", "\texttt{Table\_Name}", and "\texttt{Long\_Name}". For example, for the country "VIR", the short name is "Virgin\_Islands", the table name is "Virgin\_Islands\_(U.S.)", and the long name is "Virgin\_Islands\_of\_the\_United\_States". These attributes are not significant enough; they should rather be named "\textit{Short\_Country\_Name}", "\textit{Table\_Country\_Name}", and "\textit{Long\_Country\_Name}". On the other hand, ontology $\mathcal{O}_2$ has a datatype property (attribute) called "\texttt{Country\_or\_Area}"; however, there are only instances of countries and no instances of areas. This attribute should rather be named "\textit{Country}" or "\textit{Country\_Name}" (not "\texttt{Country\_or\_Area}"). Obviously, LogMap and AML could not match the properties "\texttt{Short\_Name}", "\texttt{Table\_Name}", and "\texttt{Long\_Name}" with the property "\texttt{Country\_or\_Area}" since there is no word in common. There is an inherent uncertainty in these properties because their terms do not reflect their true and full meaning. We deduce that the choice of the entity names (in the original dataset) should be done carefully by domain experts. However, in real-world datasets, this is not always the case. Let us study one last example from Experiment~7:

\begin{table}[H]
\centering
\begin{adjustbox}{scale=0.9,center}
\begin{tabular}{l}
\toprule
\multicolumn{1}{c}{\textbf{Ontology}~\boldmath{$\mathcal{O}_1$}}                                        \\ \midrule

\begin{tabular}[c]{@{}l@{}}$\bullet$ \texttt{cshapes/1474} $\quad$ $\bullet$ \texttt{cshapes/1476} $\quad$ $\bullet$ \texttt{cshapes/1478}\\ $\bullet$ \texttt{cshapes/1475} $\quad$ $\bullet$ \texttt{cshapes/1477} $\quad$ $\bullet$ \textit{etc.}\end{tabular}                \\ \bottomrule

\end{tabular}
\end{adjustbox}
\end{table}

In Experiment~7, AML performs badly because some instances of $\mathcal{O}_1$ are composed of both letters and numbers at the same time. Numbers disturb the AML processing. For example, instances such as "\texttt{cshapes/1474}", "\texttt{cshapes/1475}", "\texttt{cshapes/1577}", and "\texttt{cshapes/1578}" from $\mathcal{O}_1$ are matched to many random country instances from $\mathcal{O}_2$, which lowers the precision of the AML alignment. These instances should not be matched to any instance from $\mathcal{O}_2$ because they are instance IDs that do not have any meaning; they are just contiguous unique codes identifying the table rows.

In ontologies, every entity has a meaning. Entity names can be composed of words or alphanumeric codes. Still, when an entity name is an alphanumeric code, it is highly recommended that this entity also has a \textit{label} (or multiple labels) mentioning the real full name or synonym(s) of the entity. Ontology matching tools often use the entity \textit{name} information and the entity \textit{labels} information to find correspondences between entities. However, real-world datasets do not contain such \textit{label} information. Therefore, some errors in the matching process may occur.

Overall, entity terms should be written as straightforward and clear as possible so that existing ontology matching tools can produce good results and avoid additional uncertainty.

\subsubsection{Uncertainty Caused by Different Conceptualization Choices}

Recall that there are four main types of ontological entities: concepts (or classes), object properties (or relations), datatype properties (or attributes), and individuals (or instances). Current ontology matching tools do not match different types of entities (e.g., classes to instances, classes to relations, or relations to instances). They are only capable of matching classes to classes, object properties to object properties, datatype properties to datatype properties, and instances to instances. However, there is often a difference in the structuring (or organization) of entities in two similar ontologies. Here is an example from Experiment~5:

\begin{table}[H]
\centering
\begin{adjustbox}{scale=0.9,center}
\begin{tabular}{l|c|c}
\toprule
Exp.~5 & \multicolumn{1}{c|}{\textbf{Ontology}~\boldmath{$\mathcal{O}_1$}}                  & \multicolumn{1}{c}{\textbf{Ontology}~\boldmath{$\mathcal{O}_2$}}                                        \\ \midrule

 \textbf{Attributes} & Countries                                                 & Years                \\ \addlinespace 
\textbf{Instances} & Years & Countries\\

\bottomrule

\end{tabular}
\end{adjustbox}
\end{table}

In ontology $\mathcal{O}_1$, countries are put in columns (as datatype properties or attributes), and years are put in rows (as instances); In contrast, in ontology $\mathcal{O}_2$, countries are put in rows (as instances) and years are put in columns (as datatype properties). The developers of these two datasets have made two different conceptual choices. The results of LogMap and AML are awful in Experiment~5 because there was nothing to match. The only ontology language that allows this kind of matching is the OWL-FULL language, which is not yet supported by the ontology community. Only OWL-Lite and OWL-DL languages are currently supported by the existing ontology building and matching tools. This is still a significant limitation that hinders interoperability and increases uncertainty in ontology integration tasks.

\subsubsection{Uncertainty Caused by the Domain Complexity}

The inherent domain complexity can also be a source of uncertainty. In this case, even normal users cannot be sure about ontology correspondences. We actually need a domain expert to identify the correct correspondences.

\begin{table}[H]
\centering
\caption{Domain Complexity (Different Ontology Granularity).\label{complexx}}
\begin{adjustbox}{scale=0.9,center}
\begin{threeparttable}
\begin{tabular}{l|l}
\toprule
\multicolumn{1}{c|}{\textbf{Ontology}~\boldmath{$\mathcal{O}_1$}}                  & \multicolumn{1}{c}{\textbf{Ontology}~\boldmath{$\mathcal{O}_2$}}                                  \\\midrule

\begin{tabular}[c]{@{}l@{}}$\bullet$ \texttt{Small\_island\_developing\_states}\\ $\bullet$ \textcolor{red}{\texttt{Caribbean}} \end{tabular} &

\begin{tabular}[c]{@{}l@{}}$\bullet$ \textcolor{black}{\texttt{Small\_states}}\\  $\bullet$ \textcolor{black}{\texttt{Pacific\_island\_small\_states}} \\ $\bullet$ \textcolor{red}{\texttt{Caribbean\_small\_states}} \\ $\bullet$ \textcolor{black}{\texttt{Other\_small\_states}}\end{tabular}                   \\ \bottomrule

\end{tabular}
\end{threeparttable}
\end{adjustbox}
\end{table}

In the following examples, we need an expert in the domain of geography (or the domain of geopolitics) to decide about the correspondences. Table~\ref{complexx} shows an example from Experiment~3. In this example, we are not sure whether the entity "\texttt{Small\_island\_developing\_states}" from $\mathcal{O}_1$ should be matched to the entity "\texttt{Small\_states}", the entity "\texttt{Pacific\_island\_small\_states}", or the entity "\texttt{Other\_small\_states}" from $\mathcal{O}_2$, or it should be matched to all of them at the same time. And we do not know if "\texttt{Caribbean\_small\_states}" can also be included. Tables~\ref{complexxx} and~\ref{complexxxx} show an example from Experiment~5. In this case, we are not sure whether the entity "\texttt{China}" from $\mathcal{O}_1$ should be matched to the entity "\texttt{China}" or to the entities "\texttt{China}", "\texttt{Hong\_Kong\_SAR,\_China}" and "\texttt{Macao\_SAR,\_China}" from $\mathcal{O}_2$. An expert should check all these possible correspondences and choose the correct ones.

\begin{table}[!htb]
\centering
\begin{minipage}[t]{0.4\linewidth}
\centering
\caption{Domain Complexity (Equivalence).\label{complexxx}}
\begin{adjustbox}{scale=0.9,center}
\begin{threeparttable}
\begin{tabular}{l|l}
\toprule
\multicolumn{1}{c|}{\textbf{Ontology}~\boldmath{$\mathcal{O}_1$}}                  & \multicolumn{1}{c}{\textbf{Ontology}~\boldmath{$\mathcal{O}_2$}}                                  \\\midrule

\begin{tabular}[c]{@{}l@{}}$\bullet$ \textcolor{red}{\texttt{China}} \end{tabular} &

\begin{tabular}[c]{@{}l@{}}$\bullet$ \textcolor{red}{\texttt{China}}\\  $\bullet$ \textcolor{black}{\texttt{Hong\_Kong\_SAR,\_China}} \\ $\bullet$ \textcolor{black}{\texttt{Macao\_SAR,\_China}}\end{tabular}                   \\ \bottomrule

\end{tabular}
\end{threeparttable}
\end{adjustbox}
\end{minipage}\qquad
\begin{minipage}[t]{0.4\linewidth}\centering
\centering
\caption{Domain Complexity (Subsumption).\label{complexxxx}}
\begin{adjustbox}{scale=0.9,center}
\begin{threeparttable}
\begin{tabular}{l|l}
\toprule
\multicolumn{1}{c|}{\textbf{Ontology}~\boldmath{$\mathcal{O}_1$}}                  & \multicolumn{1}{c}{\textbf{Ontology}~\boldmath{$\mathcal{O}_2$}}                                  \\\midrule

\begin{tabular}[c]{@{}l@{}}$\bullet$ \textcolor{blue}{\texttt{China}} \end{tabular} &

\begin{tabular}[c]{@{}l@{}}$\bullet$ \textcolor{blue}{\texttt{China}}\\  $\bullet$ \textcolor{blue}{\texttt{Hong\_Kong\_SAR,\_China}} \\ $\bullet$ \textcolor{blue}{\texttt{Macao\_SAR,\_China}}\end{tabular}                   \\ \bottomrule

\end{tabular}
\end{threeparttable}
\end{adjustbox}
\end{minipage}
\end{table}

\subsection{Examples of Uncertainty from the Resulting Alignments\label{ex_amb}}

The ontology matching tools used can be easily misled, as the input ontologies in these experiments contain different appellations for the same countries. In the following, we will show some examples of uncertainty cases (including false and missing correspondences) extracted from our resulting output alignments.

\begin{example}[Ambiguity Case]
\end{example}
As said in Subsection~\ref{ambiguity}, ambiguous \textit{equivalence} correspondences are a source of uncertainty in ontology alignments. Each set of ambiguous correspondences often brings with it some false correspondences. Let us take an example from Experiment~1:

\begin{table}[H]
\centering
\begin{adjustbox}{scale=0.9,center}
\begin{tabular}{l|l}
\toprule
\multicolumn{1}{c|}{\textbf{Ontology}~\boldmath{$\mathcal{O}_1$}}                  & \multicolumn{1}{c}{\textbf{Ontology}~\boldmath{$\mathcal{O}_2$}}                                        \\ \midrule

\begin{tabular}[c]{@{}l@{}}$\bullet$ \textcolor{red}{\texttt{Dem$.\_$Rep$.\_$Congo}}\\ $\bullet$ \textcolor{blue}{\texttt{Repub$.\_$of$\_$the$\_$Congo}}\end{tabular}                                                   & \begin{tabular}[c]{@{}l@{}} $\bullet$ \textcolor{red}{\texttt{Democratic$\_$Republic$\_$of$\_$the$\_$Congo}}\\ $\bullet$ \textcolor{blue}{\texttt{Congo}} \end{tabular}                  \\ \bottomrule

\end{tabular}
\end{adjustbox}
\end{table}

In Experiment~1, LogMap identified two ambiguous correspondences, as follows:

\begin{center}
\begin{minipage}{0.79\linewidth}
\begin{itemize}
    \item $\mathcal{O}_1$:$\,$\texttt{Repub.\_of\_the\_Congo} \,$\equiv$\, $\mathcal{O}_2$:$\,$\texttt{Democratic\_Republic\_of\_the\_Congo} $\,$ [$\,0.76\,$]
    \item $\mathcal{O}_1$:$\,$\texttt{Repub.\_of\_the\_Congo} ~$\equiv$~ $\mathcal{O}_2$:$\,$\texttt{Congo} $\quad$ [$\,0.8\,$]
\end{itemize}
\end{minipage}
\end{center}

The first correspondence that matches "\texttt{Repub.\_of\_the\_Congo}" with "\texttt{Democratic\_Republic\_of\_the\_Congo}" has a confidence value of $0.76$. The second correspondence that matches "\texttt{Repub.\_of\_the\_Congo}" with "\texttt{Congo}" has a confidence value of $0.8$. The first correspondence is incorrect; the second one is correct. In this case, LogMap succeeded in assigning the highest confidence value to the correct correspondence and the lower confidence value to the incorrect one, but this is not always the case. This was the only false positive generated by LogMap in Experiment~1.

\begin{example}[Ambiguity Case]
\end{example}
Let us take the same ambiguity example from Experiment~8:

\begin{table}[H]
\centering
\begin{adjustbox}{scale=0.9,center}
\begin{tabular}{l|l}
\toprule
\multicolumn{1}{c|}{\textbf{Ontology}~\boldmath{$\mathcal{O}_1$}}                  & \multicolumn{1}{c}{\textbf{Ontology}~\boldmath{$\mathcal{O}_2$}}                                        \\ \midrule

\begin{tabular}[c]{@{}l@{}}$\bullet$ \textcolor{red}{\texttt{Democratic$\_$Republic$\_$of$\_$Congo}}\\ $\bullet$ \textcolor{blue}{\texttt{Republic$\_$of$\_$Congo}}\end{tabular}                                                   & \begin{tabular}[c]{@{}l@{}} $\bullet$ \textcolor{red}{\texttt{Congo$\_$(Brazzaville)}}\\ $\bullet$ \textcolor{blue}{\texttt{Congo$\_$(Kinshasa)}} \end{tabular}                  \\ \bottomrule

\end{tabular}
\end{adjustbox}
\end{table}

In Experiment~8, AML identified two ambiguous correspondences, as follows:

\begin{center}
\begin{minipage}{0.79\linewidth}
\begin{itemize}
    \item $\mathcal{O}_1$:$\,$\texttt{Republic\_of\_Congo} ~$\equiv$~ $\mathcal{O}_2$:$\,$\texttt{Congo\_(Brazzaville)} $\quad$ [$\,0.8\,$]
    
    \item $\mathcal{O}_1$:$\,$\texttt{Republic\_of\_Congo} ~$\equiv$~ $\mathcal{O}_2$:$\,$\texttt{Congo\_(Kinshasa)} $\quad$ [$\,0.8\,$]
\end{itemize}
\end{minipage}
\end{center}

The first correspondence that matches "\texttt{Republic\_of\_Congo}" with "\texttt{Congo\_(Brazzaville)}" has a confidence value of $0.8$. The second correspondence that matches "\texttt{Republic\_of\_Congo}" with "\texttt{Congo\_(Kinshasa)}" has a confidence value of $0.8$. The first correspondence is incorrect; the second one is correct. In this case, both correspondences have the same confidence value, which makes it even trickier to choose automatically.

\begin{example}[Ambiguity Case]
\end{example}
Here is another ambiguity example from Experiment~2:

\begin{table}[H]
\centering
\begin{adjustbox}{scale=0.9,center}
\begin{tabular}{l|l}
\toprule
\multicolumn{1}{c|}{\textbf{Ontology}~\boldmath{$\mathcal{O}_1$}}                  & \multicolumn{1}{c}{\textbf{Ontology}~\boldmath{$\mathcal{O}_2$}}                                        \\ \midrule

\begin{tabular}[c]{@{}l@{}}$\bullet$ \textcolor{red}{\texttt{Sudan}}\\ $\bullet$ \textcolor{blue}{\texttt{South$\_$Sudan}}\\ $\bullet$ \texttt{South$\_$(former)} \end{tabular}                                                   & \begin{tabular}[c]{@{}l@{}} $\bullet$ \textcolor{red}{\texttt{Sudan}}\\ $\bullet$ \textcolor{blue}{\texttt{South$\_$Sudan}} \end{tabular}                  \\ \bottomrule

\end{tabular}
\end{adjustbox}
\end{table}

In Experiment~2, AML identified three correspondences including two ambiguous ones, as follows:

\begin{center}
\begin{minipage}{0.79\linewidth}
\begin{itemize}
    \item $\mathcal{O}_1$:$\,$\texttt{Sudan} ~$\equiv$~ $\mathcal{O}_2$:$\,$\texttt{Sudan} $\quad$ [$\,1.0\,$]
    
    \item $\mathcal{O}_1$:$\,$\texttt{South\_Sudan} ~$\equiv$~ $\mathcal{O}_2$:$\,$\texttt{South\_Sudan} $\quad$ [$\,1.0\,$]
    
    \item $\mathcal{O}_1$:$\,$\texttt{Sudan\_(former)} ~$\equiv$~ $\mathcal{O}_2$:$\,$\texttt{Sudan} $\quad$ [$\,1.0\,$]
\end{itemize}
\end{minipage}
\end{center}

The first and third correspondences are ambiguous because they have an entity in common ($\mathcal{O}_2$:$\,$\texttt{Sudan}). They have the same confidence measure, so we cannot automatically choose one from them. The first and second correspondences are correctly matched; however, the third one is incorrect. Sudan and South Sudan previously formed a single country called Sudan. Therefore, "\texttt{Sudan\_(former)}" from $\mathcal{O}_1$ should be a super-entity of both "\texttt{Sudan}" and "\texttt{South\_Sudan}" from $\mathcal{O}_2$ (i.e., both "\texttt{Sudan}" and "\texttt{South\_Sudan}" from $\mathcal{O}_2$ should be sub-entities of "\texttt{Sudan\_(former)}" from $\mathcal{O}_1$). The right matching would identify the two following \textit{subsumption} correspondences:

\begin{center}
\begin{minipage}{0.79\linewidth}
\begin{itemize}
    \item $\mathcal{O}_1$:$\,$\texttt{Sudan\_(former)} \,$\sqsupseteq$ ~$\mathcal{O}_2$:$\,$\texttt{Sudan} $\quad$
    
    \item $\mathcal{O}_1$:$\,$\texttt{Sudan\_(former)} \,$\sqsupseteq$~ $\mathcal{O}_2$:$\,$\texttt{South\_Sudan} $\quad$
\end{itemize}
\end{minipage}
\end{center}

Current ontology matching tools can only identify \textit{equivalence} correspondences. Only a human being can understand and identify \textit{subsumption} correspondences. In many cases, the ambiguous \textit{equivalence} correspondences convey subsumption. That is why it is preferred that a human expert check these ambiguous correspondences. Not all ambiguity cases can be resolved automatically.

\begin{example}[Semantic Matching Case]
\end{example}

As stated before, neither LogMap nor AML could detect the correspondence between the entity "\texttt{Near\_East}" from $\mathcal{O}_1$ and the entity "\texttt{WesternAsia}" from $\mathcal{O}_2$ in Experiment~1. Similarly, in Experiment~7, neither LogMap nor AML could detect the correspondence between the entity "\texttt{Swaziland}" from $\mathcal{O}_1$ and the entity "\texttt{Eswatini}" from $\mathcal{O}_2$. The same goes for the entity "\texttt{Palestine}" from $\mathcal{O}_1$ and the entity "\texttt{West\_Bank\_and\_Gaza}" from $\mathcal{O}_2$ in Experiment~3.

\begin{center}
\begin{minipage}{0.79\linewidth}
\begin{itemize}
    \item $\mathcal{O}_1$:$\,$\texttt{Near$\_$East} ~$\equiv$~ $\mathcal{O}_2$:$\,$\texttt{WesternAsia}

    \item $\mathcal{O}_1$:$\,$\texttt{Swaziland} ~$\equiv$~ $\mathcal{O}_2$:$\,$\texttt{Eswatini}
    
    \item $\mathcal{O}_1$:$\,$\texttt{Palestine} ~$\equiv$~ $\mathcal{O}_2$:$\,$\texttt{West\_Bank\_and\_Gaza}
\end{itemize}
\end{minipage}
\end{center}

In these examples, the entity from $\mathcal{O}_1$ and the entity from $\mathcal{O}_2$ have no word or token in common, but they are synonymous. As a result, \textit{lexical} and \textit{string-based} matchers are not able to identify such correspondences. In this case, a \textit{semantic} matcher is needed. However, the main limitation of current ontology matching tools is the \textit{semantic} matching task, because \textit{semantic} matching is the most challenging type of matching. To find similarities, a semantic matcher often requires external resources, such as dictionaries and thesauruses in different domains.
Here are some additional straightforward cases that LogMap and AML did not pick up on:

\begin{center}
\begin{minipage}{0.79\linewidth}
\begin{itemize}
    \item $\mathcal{O}_1$:$\,$\texttt{Kyrgyzstan} ~$\equiv$~ $\mathcal{O}_2$:$\,$\texttt{Kyrgyz$\_$Republic}
    
    \item $\mathcal{O}_1$:$\,$\texttt{Czechia} ~$\equiv$~ $\mathcal{O}_2$:$\,$\texttt{Czech$\_$Republic}
    
    \item $\mathcal{O}_1$:$\,$\texttt{Slovakia} ~$\equiv$~ $\mathcal{O}_2$:$\,$\texttt{Slovak$\_$Republic}
    
\end{itemize}
\end{minipage}
\end{center}

These correspondences were not detected because the used external dictionaries do not necessarily contain some particular words in their database (especially domain-oriented words). For example, \href{https://wordnet.princeton.edu/}{WordNet}, which is a large lexical database of English, does not recognize the word \texttt{Czechia} or the word \texttt{Kyrgyz} (although it recognizes the word \texttt{Czech} and the word \texttt{Kyrgyzstan}). Besides, the words from $\mathcal{O}_1$ are proper nouns. Thus, word preprocessing techniques used by the matching tools (e.g., \textit{tokenization} and \textit{lemmatization}) do not even return a common token between these entities.

\begin{example}[Abbreviations and Acronyms Case]
\end{example}

In Experiment~6, neither LogMap nor AML could identify the following correspondences:

\begin{center}
\begin{minipage}{0.79\linewidth}
\begin{itemize}
    \item $\mathcal{O}_1$:$\,$\texttt{Hong\_Kong\_Special\_Administrative\_Region\_of\_the\_People's\_Republic\_of\_China}\\ $\equiv$~ $\mathcal{O}_2$:$\,$\texttt{China,\_Hong\_Kong\_SAR}

    \item $\mathcal{O}_1$:$\,$\texttt{United\_Kingdom\_of\_Great\_Britain\_and\_Northern\_Ireland}\\$\equiv$~ $\mathcal{O}_2$:$\,$\texttt{United\_Kingdom}

    \item $\mathcal{O}_1$:$\,$\texttt{The$\_$Former$\_$Yugoslav$\_$Republic$\_$of$\_$Macedonia} ~$\equiv$~ $\mathcal{O}_2$:$\,$\texttt{Macedonia}

    \item $\mathcal{O}_1$:$\,$\texttt{Plurinational$\_$State$\_$of$\_$Bolivia} ~$\equiv$~ $\mathcal{O}_2$:$\,$\texttt{Bolivia}
\end{itemize}
\end{minipage}
\end{center}

It is tricky for any ontology matching tool to detect the above correspondences because one of the two entities contains a lot of tokens, to the point that it will appear to the tool that it is not a correspondence. Otherwise, the matching tool can detect these correspondences, but they will have very low confidence values. This proves that not all low-confidence correspondences are necessarily incorrect. Abbreviations and acronyms make it even harder for the matching tools to detect such correspondences. For example, in Experiment~3, LogMap did not find the following correspondence; however, AML identified it and gave it a confidence value of $0.51$:

\begin{center}
\begin{minipage}{0.79\linewidth}
\begin{itemize}
    \item $\mathcal{O}_1$:$\,$\texttt{Lao\_People's\_Democratic\_Republic} ~$\equiv$~ $\mathcal{O}_2$:$\,$\texttt{Lao\_PDR} $\quad$ [$\,0.51\,$]
\end{itemize}
\end{minipage}
\end{center}

In all these cases, human intervention is necessary to remove the false correspondences and add the missing correct correspondences. No automated ontology matching tool can avoid uncertainty, even the most powerful ones.


\subsection{Lessons Learnt}

Despite the very good performance of ontology matching tools, the set of experiments that have been conducted shows how even a relatively simple use case can present several flaws that arguably lead to a situation of uncertainty once the integration process is automated. Besides, approaches used in the state of the art for managing uncertainty in ontology alignments are not effective. They cannot avoid deleting correct correspondences while filtering the alignment from ambiguous or low-confidence correspondences.

We notice that there are many uncertainty cases in ontology alignments, and it is difficult to treat all of them at once. However, it is very important to decrease uncertainty as much as possible. Uncertainty in ontology alignments can appear in several forms, such as matcher unreliability (or confidence unreliability), the high level of domain complexity, the ambiguity of correspondences, the incompleteness and/or incorrectness of correspondences, etc. Thus, there are several factors that can contribute to uncertainty in ontology matching. We cannot measure the uncertainty amount of a given alignment by simply using a single metric that holds all uncertainty aspects. Still, it would be interesting to combine all the available evaluation metrics into a single metric that globally reflects the holistic uncertainty of an alignment.

There are no specific evaluation metrics dedicated to measuring uncertainty in ontology alignments. The error rate associated with a matching process can be measured according to standard metrics that evaluate the quality of alignments. These metrics assess the completeness and correctness of an alignment compared to a reference. We have also found the \textit{ambiguity degree} metric in the literature, which is very relevant for uncertainty measurement. However, the acceptability of a given error rate at an application level may depend on several factors. In general terms, it is reasonable to distinguish between error-tolerant applications that may assume a certain amount of imprecision and inaccuracies (such as \textit{text annotation} applications and \textit{recommendation} applications) and critical applications that require high correctness and precision (such as \textit{query answering} applications or the case study previously presented). Future works should work on suggesting new metrics that better reflect the uncertainty in ontology alignments.

It is difficult to quantify uncertainty in ontology alignments, especially implicit uncertainty that cannot be measured. For example, how to know if the confidence values are reliable or not? In this case, one can suggest a sample checking process where the user manually evaluates some random correspondences from the alignment, and then a global score is returned based on the user responses. Similarly, how to know if the domain(s) of the input ontologies has (have) a high level of complexity? In this case, a score can be proposed by the user or the expert after examining the two ontologies (to be matched) or after examining the ambiguous correspondences of the output alignment.

We can also have an idea about the risk of ambiguity between the two input ontologies (to be matched). We can suggest a metric that reflects the percentage of common words (or tokens) among all the words (or tokens) composing the entities of both ontologies. It can be considered a metric that shows how similar the two ontologies are, since ontology similarity is an important uncertainty indicator. The more the ontologies are similar, the more they represent a risk of ambiguity and uncertainty in the matching process. These proposed metrics can help us estimate the degree of uncertainty of the returned alignment.

Overall, the uncertainty topic is not well treated in the ontology matching area. Uncertainty management in ontology alignments is still an open issue. New metrics for uncertainty evaluation and new uncertainty management solutions are really needed in the area of ontology matching (and the area of schema matching in general).

The holistic performance in terms of reliability of the fully automated integration process suggests semi-supervised matching approaches. The latter can intuitively facilitate manual operations by providing suggestions that need human validation before commitment to the system. Generally speaking, semi-supervised matching processes are not necessarily scalable. However, they can be considered effective in validating detected correspondences by rejecting the incorrect ones. On the other hand, semi-supervision is not that effective in managing missed correspondences. It implies the validator should check the whole alignment. Such an issue becomes more and more relevant, if not an actual bottleneck, when the amount of information to integrate increases.
A situation in which the target alignment presents ambiguities or is not completely clear constitutes a special case for whom automated matching can provide an acceptable approximation, especially on a large scale.

In conclusion, despite the notable technological advances at both a theoretical level and an application level, a reliable automated integration process cannot be currently enforced, regardless of the scale and the characteristics of the target environment.

\section{Conclusion\label{section8}}

This paper reviews the ontology matching area, focusing on the aspects of uncertainty and uncertainty management. It also deals with the implementation of a knowledge-building case study from which we make observations, discuss uncertainty situations, and extract the lessons learned from using automatic ontology matching systems.
The case study aims to provide an integrated semantic data space from heterogeneous raw datasets by leveraging automatic matching tools.
We have conducted experiments on real data by combining the virtual table method with existing tools for automatic ontology matching.
Indeed, we first use a tool that supports a user-friendly conversion of data into a semantic format. Then, we perform an automatic ontology matching process using well-known tools for ontology matching.

The results clearly show the significant uncertainty introduced by automatic ontology matching methods, even considering a relatively simple case study. In general terms, regardless of the adopted matching tool, experiments behold a high number of detected false positives and false negatives. It can be summarized as a fundamental unreliability of automatic matching techniques, probably with the exception of error-tolerant applications.
Realistically, we believe that a further consolidation of current matching techniques with an increased level of customization may result in a more effective application of semi-supervised integration methods.

Future work will aim to manage uncertainty in practice by inferring potential relationships between the characteristics of the raw data to be integrated and the resulting integrated data space. Besides, we are starting to explore setting up a holistic alignment-based approach for integrating multiple ontologies driven by a holistic ontology matching process.


\section*{Acknowledgments}
The dataset conversion tool~\cite{pileggi2020ontological} described in the paper has been implemented by Hayden Crain and has been funded by the Faculty of Engineering and IT at the University of Technology Sydney through a Summer Scholarship. 

\section*{Declaration of interests} The authors declare that they have no known competing financial interests or personal relationships that could have appeared to influence the work reported in this paper.


\end{document}